\PassOptionsToPackage{table,dvipsnames}{xcolor}
\documentclass[letterpaper]{article} 
\usepackage{aaai2026}  
\usepackage{times}  
\usepackage{booktabs}
\usepackage{helvet}  
\usepackage{courier}  
\usepackage[hyphens]{url}  

\usepackage{csquotes}

\usepackage{pifont}

\usepackage{booktabs}
\usepackage{amssymb}
\usepackage{enumitem}
\usepackage{multirow}
\usepackage{marvosym}

\usepackage{graphicx} 
\usepackage{natbib}  
\usepackage{caption} 
\usepackage{algorithm}
\usepackage{algorithmic}
\usepackage{makecell}
\usepackage{array}
\usepackage{amsmath}

\usepackage{booktabs}
\usepackage{tabularx}

\newcolumntype{Y}{>{\centering\arraybackslash}X}

\usepackage{newfloat}
\usepackage{listings}
\DeclareCaptionStyle{ruled}{labelfont=normalfont,labelsep=colon,strut=off} 
\floatstyle{ruled}
\newfloat{listing}{tb}{lst}{}
\floatname{listing}{Listing}
\title{VisualTrans: A Benchmark for Real-World Visual Transformation Reasoning}
\author{
    Yuheng Ji$^{1,2,*}$, Yipu Wang$^{3,1,*,\dagger}$, Yuyang Liu$^{1,2}$, Xiaoshuai Hao$^{4}$, Yue Liu$^{5}$ \\
    Yuting Zhao$^{1,2}$, Huaihai Lyu$^{1,2}$, Xiaolong Zheng$^{1,2,3,\text{\Letter}}$ \\
}

\affiliations{
    $^1$ Institute of Automation, Chinese Academy of Sciences\\
    $^2$ School of Artificial Intelligence, University of Chinese Academy of Sciences\\
    $^3$ School of Advanced Interdisciplinary Sciences, University of Chinese Academy of Sciences\\
    $^4$ Beijing Academy of Artificial Intelligence \hspace{1.5em} 
    $^5$ National University of Singapore\\
}

\usepackage{bibentry}
\begin{document}

\newcommand\blfootnote[1]{%
\begingroup
\renewcommand\thefootnote{}\footnote{#1}%
\addtocounter{footnote}{-1}%
\endgroup
}

\thispagestyle{empty}
\pagestyle{empty}
\twocolumn[{
\renewcommand\twocolumn[1][]{#1}
\vspace{-30pt}
\maketitle
\begin{center}
    \captionsetup{type=figure}
    \vspace{-25pt}
    \includegraphics[width=1.0\textwidth]{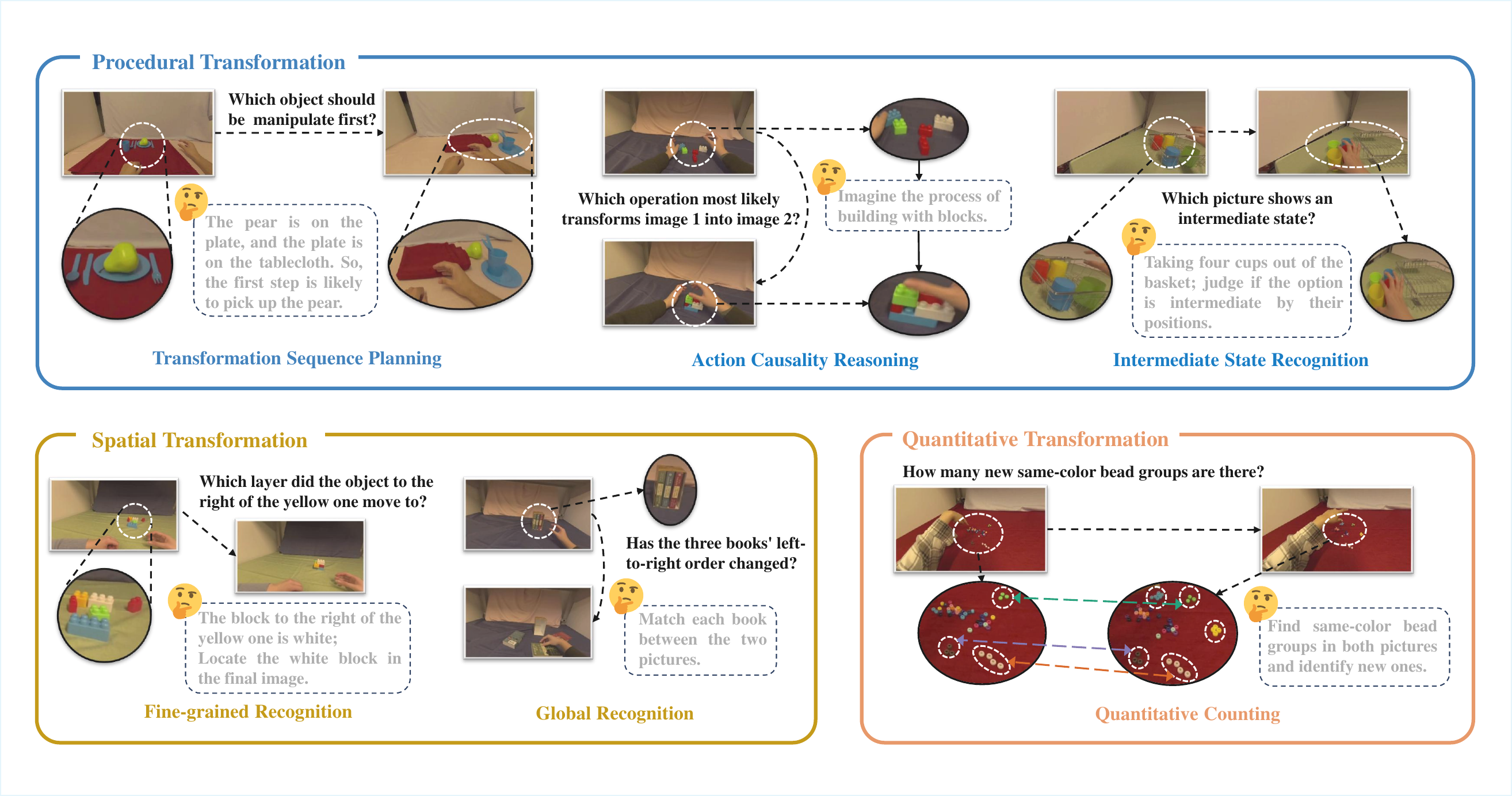}
    \vspace{-2em}
    \captionof{figure}{\textbf{Overview of VisualTrans.} We introduce VisualTrans, a real-world benchmark for Visual Transformation Reasoning (VTR), featuring 12 egocentric manipulation tasks and evaluating three core reasoning dimensions—spatial, procedural, and quantitative—through systematically constructed question-answer pairs, enabling diagnosis of model capabilities in dynamic scene understanding and causal inference.}
    \label{fig:teaser}
\end{center}

}]

\let\thefootnote\relax\footnotetext{$^{*}$ Equal contribution.}
\let\thefootnote\relax\footnotetext{$^{\dagger}$ Project leader.}
\let\thefootnote\relax\footnotetext{$^{\text{\Letter}}$ Corresponding author.}

\begin{abstract}
Visual transformation reasoning (VTR) is a vital cognitive capability that empowers intelligent agents to understand dynamic scenes, model causal relationships, and predict future states, and thereby guiding actions and laying the foundation for advanced intelligent systems.
However, current TVR benchmarks face significant limitations, including a sim-to-real gap, restricted task complexity, and incomplete coverage of reasoning dimensions, which diminish their practical utility for intelligent systems, especially in real-world scenarios.
To address these limitations, we introduce \textbf{\textit{VisualTrans}}, the first comprehensive benchmark specifically designed for VTR in real-world human-object interaction scenarios. 
VisualTrans encompasses 12 semantically diverse manipulation tasks and systematically evaluates three essential reasoning dimensions—\textit{\textbf{spatial}}, \textit{\textbf{procedural}}, and \textit{\textbf{quantitative}}—through 6 well-defined subtask types. The benchmark features 472 high-quality question-answer pairs in various formats, including multiple-choice, open-ended counting, and target enumeration.
We introduce a scalable data construction pipeline built upon first-person manipulation videos, which integrates task selection, image pair extraction, automated metadata annotation with large multimodal models, and structured question generation. Human verification ensures the final benchmark is both high-quality and interpretable.
Evaluations of various state-of-the-art vision-language models show strong performance in static spatial tasks. However, they reveal notable shortcomings in dynamic, multi-step reasoning scenarios, particularly in areas like intermediate state recognition and transformation sequence planning.
These findings highlight fundamental weaknesses in temporal modeling and causal reasoning, providing clear directions for future research aimed at developing more capable and generalizable VTR systems. The benchmark toolkit and its associated code have been made publicly available at https://github.com/WangYipu2002/VisualTrans.

\end{abstract}

\section{Introduction}

\begin{table*}[t]
\centering
\caption{Comparison of VTR benchmarks across key evaluation dimensions.}
\scalebox{1.0}{
\begin{tabular}{lccccc}
\toprule
\textbf{Benchmark} & \textbf{Real-world} & \textbf{Multi-object} & \textbf{Spatial} & \textbf{Procedural} & \textbf{Quantitative} \\
\midrule
CLEVR-Change~\cite{clevr_change}        & \textcolor{red}{\ding{55}} & \textcolor{red}{\ding{55}} & \textcolor{ForestGreen}{\ding{51}} & \textcolor{red}{\ding{55}} & \textcolor{red}{\ding{55}} \\
BIRD~\cite{BIRD}                        & \textcolor{red}{\ding{55}} & \textcolor{ForestGreen}{\ding{51}} & \textcolor{red}{\ding{55}} & \textcolor{ForestGreen}{\ding{51}} & \textcolor{red}{\ding{55}} \\
TRANCE~\cite{TRANCE}                    & \textcolor{red}{\ding{55}} & \textit{partial} & \textcolor{ForestGreen}{\ding{51}} & \textcolor{red}{\ding{55}} & \textcolor{ForestGreen}{\ding{51}} \\
Tranco~\cite{TRANCO}                    & \textcolor{ForestGreen}{\ding{51}} & \textit{partial} & \textcolor{red}{\ding{55}} & \textcolor{ForestGreen}{\ding{51}} & \textcolor{red}{\ding{55}} \\
SAT~\cite{SAT}                          & \textcolor{ForestGreen}{\ding{51}} & \textcolor{red}{\ding{55}} & \textcolor{ForestGreen}{\ding{51}} & \textcolor{red}{\ding{55}} & \textcolor{red}{\ding{55}} \\
\textbf{VisualTrans (Ours)}            & \textcolor{ForestGreen}{\ding{51}} & \textcolor{ForestGreen}{\ding{51}} & \textcolor{ForestGreen}{\ding{51}} & \textcolor{ForestGreen}{\ding{51}} & \textcolor{ForestGreen}{\ding{51}} \\
\bottomrule
\end{tabular}
}
\label{tab:benchmark_comparison}
\end{table*}

Perceiving and reasoning about dynamic changes of objects in a scene across different time points is a key capability of the human cognitive system~\cite{Piaget, goddu2024development}, commonly referred to as visual transformation reasoning (VTR). This faculty enables individuals to understand causal relationships among objects, anticipate future states, and thereby perform complex manipulation and decision-making tasks. With the rapid development of general Vision-Language Models (VLMs)~\cite{gemini25pro,hurst2024gpt,huang2025foundation}, enabling models to acquire similar VTR abilities has become an important step toward applications in virtual reality~\cite{Spatialrgpt,ViDDAR,egoprompt} and embodied intelligent systems~\cite{zhang2025embodied,robobrain,robobrain2.0}. Unlike traditional image recognition or visual question answering tasks~\cite{han2024latency, hu2024prompting, advlora}, VTR emphasizes on the modeling and understanding of dynamic states. It requires models not only to capture fine-grained changes across images at different time points, but also to possess causal reasoning and process modeling abilities, recovering the underlying manipulation logic and transformation trajectories from visual representations.

Although early works such as TRANCE~\cite{TRANCE} have attempted to construct evaluation benchmarks for VTR tasks, existing benchmarks still suffer from significant shortcomings in terms of task realism, transformation complexity, and coverage of reasoning dimensions, as shown in Tab.~\ref{tab:benchmark_comparison}. Specifically:
\textit{\textbf{(1) Significant sim-to-real gap:}} Most benchmarks (\emph{e.g.}, CLEVR-Change~\cite{clevr_change}, TRANCE~\cite{TRANCE}) mainly rely on synthetic images generated by 3D engines. Although they offer controllability, they lack the semantic diversity and visual complexity of the real world. For example, objects in CLEVR-Change are regular geometric shapes and lack real semantic attributes, making it difficult for models to generalize to real-world natural images or robotic manipulation scenarios.
\textit{\textbf{(2) Limited complexity of object transformations:}} Most benchmarks involve only single-object, single-step changes, lacking the complex dynamic operations commonly found in real-world tasks such as multi-object collaboration, object reconfiguration, or nested structure changes. For example, SAT~\cite{SAT} mainly focuses on static stacking of single structured objects; although BIRD~\cite{BIRD} considers multiple objects, its interaction processes and transformation sequences remain rather static and lack process modeling.
\textit{\textbf{(3) Lack of systematic coverage of multi-dimensional reasoning abilities:}} Existing benchmarks lack systematic design and cannot comprehensively evaluate models’ performance across multiple levels such as spatial understanding, procedural reasoning, and quantity perception. This limits their effectiveness as diagnostic tools for general VTR ability.

To overcome the above challenges, we propose \textit{\textbf{VisualTrans}}, a multi-dimensional VTR benchmark constructed from real-world human manipulation scenarios, as shown in Fig.~\ref{fig:teaser}. VisualTrans is based on egocentric video data and covers 12 types of real-world semantic daily tasks such as block stacking, desk tidying, dish stacking, etc. Specifically, we designed an automated data construction pipeline that integrates multimodal expert models~\cite{GroundingDino,gemini25pro} to perform object detection, metadata extraction, and question-answer (QA) generation, followed by human verification, ultimately constructing an evaluation set containing 472 high-quality QA samples. VisualTrans systematically assesses models’ understanding of visual transformations across 3 core reasoning dimensions: \textit{spatial transformation} (perception of spatial structure changes), \textit{quantitative transformation} (perception of quantity changes), and \textit{procedural transformation} (modeling of manipulation processes). It includes 6 fine-grained subtask types and supports 3 QA formats—multiple choice, open-ended counting, and target enumeration—enabling comprehensive evaluation of visual reasoning capabilities.

Compared with existing VTR benchmarks, VisualTrans offers three distinct advantages: \textit{\textbf{(1) Real-world fidelity:}} All samples are sourced from real human manipulation videos, significantly reducing the sim-to-real gap and better reflecting real-world complexity. \textit{\textbf{(2) Transformation complexity:}} VisualTrans includes multi-object, multi-step, and multi-type transformation processes, capturing the rich dynamics of everyday tasks that are often absent in prior benchmarks. \textit{\textbf{(3) Multi-dimensional reasoning coverage:}} three core dimensions: spatial understanding, procedural reasoning, and quantity perception. This enables a comprehensive diagnostic tool for assessing VLMs’ visual reasoning capabilities. We further evaluate several mainstream open-source and closed-source VLMs on VisualTrans and observe that current models still exhibit significant limitations in multi-step reasoning, temporal modeling, and object relationship understanding. In particular, challenges remain in intermediate state recognition and action sequence planning, which highlight critical areas for improvement in future intelligent systems. Our main contributions are as follows:

\begin{itemize}
    \item We introduce \textit{\textbf{VisualTrans}}, the first VTR benchmark built from real egocentric manipulation videos, covering 12 real-world task types, 3 core reasoning dimensions, 6 subtask categories, and 3 QA formats, with a total of 472 high-quality samples for systematic evaluation.
    \item We develop an efficient automated data construction pipeline that integrates multimodal expert models for object extraction, metadata generation, and QA construction, followed by human verification, enabling scalable and high-quality benchmark creation.
    \item We conduct comprehensive evaluations of multiple state-of-the-art VLMs, revealing common limitations in temporal reasoning, multi-step process modeling, and causal understanding—offering insights for future model design and training strategies.

\end{itemize}

\section{Related Work}
\textbf{Visual Transformation Tasks and Benchmarks.} Early research on visual transformation tasks was predominantly built upon simulated environments using the CLEVR~\cite{Clevr} engine. For instance, CLEVR-Change~\cite{clevr_change} formulates the task as image captioning that describes subtle attribute changes between paired images. TRANCE~\cite{TRANCE} introduces the notion of transformation reasoning for the first time, providing a benchmark with multi-step changes and diverse observation viewpoints. BIRD~\cite{BIRD} focuses on inferring event sequences throughout visual transformations. However, these benchmarks rely heavily on synthetic scenes and regular geometric objects, lacking the semantic richness and visual complexity of the real world. Tranco~\cite{TRANCO} addresses this limitation by building on the open-world COIN video dataset~\cite{Coin}, enabling models to reason about more realistic state changes in unconstrained environments. SAT~\cite{SAT} is a benchmark centered on motion perception and spatial reasoning, testing models’ abilities to infer how object movements affect spatial configurations—implicitly involving object transformation understanding. In contrast to these efforts, our proposed VisualTrans is the first benchmark constructed from real-world human interaction scenarios, specifically designed to assess VTR across multiple dimensions. It captures complex object dynamics and diverse reasoning challenges that better reflect realistic manipulation tasks.

\textbf{Visual Reasoning in Vision-Language Models.} Recent advances in VLMs have significantly improved their performance on visual reasoning tasks, encompassing spatial understanding~\cite{Robospatial,zhou2025code,reasonrft,zhou2025roborefer}, chart interpretation~\cite{chartassistant,shen2025chart}, instruction following~\cite{Vila,Ground-V}, safety moderation \cite{liu2025guardreasoner}, and long-chain reasoning based on visual input~\cite{li2025vocot,chenbring}. These models have demonstrated strong generalization capabilities in benchmark evaluations, particularly in fine-grained perception and multi-step reasoning. Alongside these developments, a growing body of research has begun to treat visual reasoning as an independent and essential capability, emphasizing compositional understanding, temporal modeling, and causal inference. Some approaches introduce operation chains to emulate human multi-stage cognitive processes when solving visual problems~\cite{CogCoM}. Others draw inspiration from hierarchical decision-making and planning theories in cognitive science to guide multi-hop reasoning~\cite{Visuothink}. Prompt-based strategies have also been explored to induce structured reasoning patterns in tasks such as object detection and relational modeling~\cite{DetToolChain}. While these efforts have yielded promising progress, many remain limited to synthetic datasets or short-horizon, static scenarios, lacking coverage of more complex reasoning challenges encountered in the real world. In contrast, our proposed benchmark, VisualTrans, focuses on VTR in real-world contexts, requiring models to understand the dynamic evolution of object states over time.

\section{VisualTrans Benchmark}

\subsection{Task Definition}
VisualTrans benchmark can be formally defined as the task of identifying transformation processes between paired visual inputs. Specifically, each sample consists of two images: an initial scene image \( I_1 \in \mathbb{R}^{H \times W \times C} \) and a transformed scene image \( I_2 \in \mathbb{R}^{H \times W \times C} \), where \( H \), \( W \), and \( C \) denote the height, width, and number of channels respectively. The objective is to infer a transformation-aware answer \( A \) based on the change from \( I_1 \) to \( I_2 \), guided by a task-specific question \( Q \).

This process can be formulated as a function:
\begin{equation}
T: (I_1, I_2, Q) \rightarrow A,
\end{equation}
where \( Q \) is a prompt describing the reasoning requirement (\emph{e.g.}, ``Which object was removed?'' or ``How many objects changed color?''), and \( A \) is the answer in one of the defined output formats. The questions in VisualTrans fall into three primary categories:
\begin{itemize}
    \item \textbf{Multiple-choice questions}: Select the most plausible answer from a set of candidates \( \{A, B, C, D\} \), and output the corresponding option label. The candidates may be textual descriptions or image options, depending on the task type.
    \item \textbf{Open-ended counting questions}: Predict the number of relevant transformations, yielding an integer response (\emph{e.g.}, \( A = 2 \)).
    \item \textbf{Target enumeration questions}: Enumerate all objects satisfying a transformation condition (\emph{e.g.}, ``red block'', ``green cup''), producing a set \( A = \{a_1, a_2, \ldots, a_n\} \).
\end{itemize}

By modeling the transformation function \( T \), VisualTrans aims to evaluate a model's capacity for structured visual reasoning over dynamic changes in scenes.

\subsection{Task Categorization}

To systematically evaluate the model’s multi-level capabilities in VTR, we categorize tasks into three reasoning types:

\textbf{Spatial Transformation.}  
This category focuses on recognizing changes in object positions, structures, and overall spatial arrangements between the initial and transformed scenes. Models are required to reason about spatial displacements—such as front-to-back, left-to-right, or top-to-bottom shifts—and reconstruct coherent object layouts. Such spatial reasoning is crucial in dynamic environments like robotic manipulation, where fine-grained spatial understanding governs planning and control. We further define two subtypes:

\begin{itemize}
    \item \textbf{Fine-grained change recognition}: Identify the new location of a previously fixed-position object.  
    \textit{E.g., Where is the object that was originally at the rightmost position now located?}
    
    \item \textbf{Global change recognition}: Determine whether the overall spatial configuration or relative positions among objects have changed.  
    \textit{E.g., Has the relative position between the two cups changed?}
\end{itemize}

\textbf{Procedural Transformation.}  
This category targets the modeling of transformation procedures from the initial state to the final outcome. The goal is to infer intermediate steps or underlying causal actions that could plausibly result in the observed transformation. This form of reasoning is especially pertinent to embodied agents and task planning systems. It comprises the following subtypes:

\begin{itemize}
    \item \textbf{Intermediate state recognition}: Identify a plausible intermediate configuration that could occur between the initial and final states.  
    \textit{E.g., Which candidate image represents a valid intermediate stage?}
    
    \item \textbf{Action causality reasoning}: Infer the likely action or manipulation responsible for the observed changes.  
    \textit{E.g., Which of the provided operations most likely caused the transformation?}
    
    \item \textbf{Transformation sequence planning}: Predict the most probable final action needed to complete the transformation process.  
    \textit{E.g., What is the final step required to reach the target state?}
\end{itemize}

\textbf{Quantitative Transformation.}  
This category evaluates the model’s ability to detect numerical changes in object presence across transformations. Tasks include identifying added or removed objects and estimating their quantities.  
\textit{E.g., How many new food items have appeared on the plate compared to the initial scene?}

\begin{figure}[t]
    \centering
    \includegraphics[width=\linewidth]{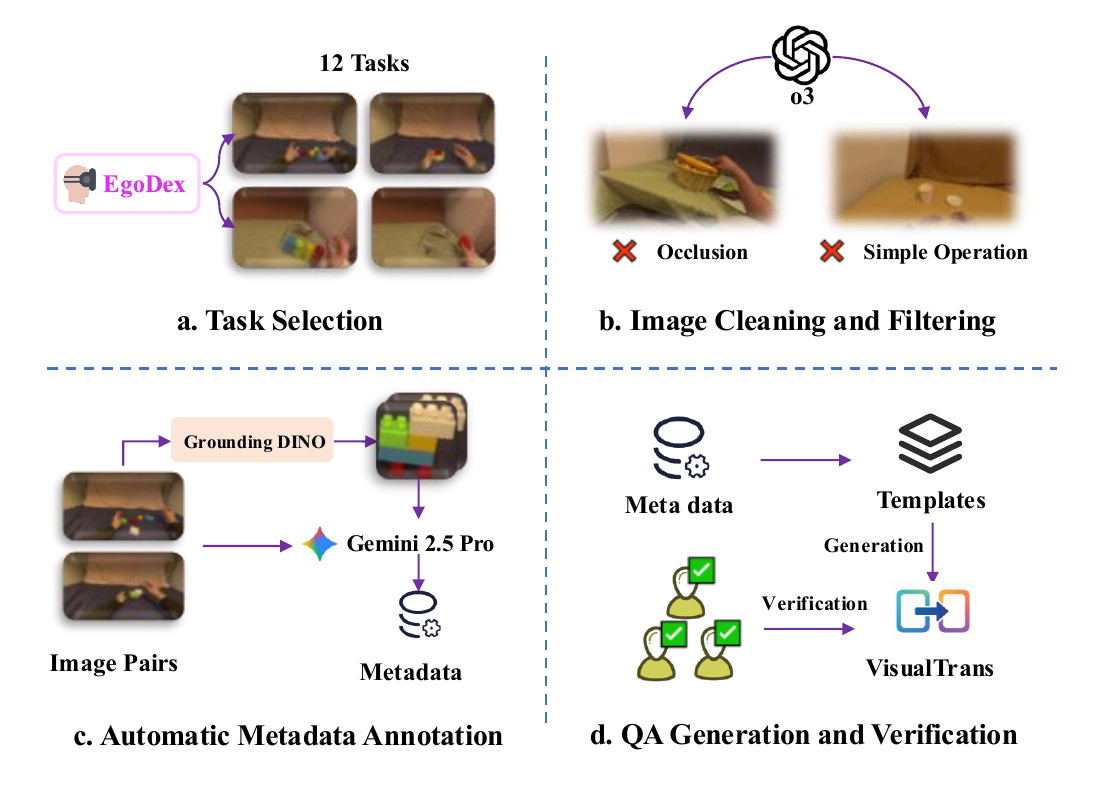}
    \caption{\textbf{Overview of Construction Pipeline.}}
    \label{fig:visualtrans-pipeline}
\end{figure}

\subsection{VisualTrans Construction Pipeline}

We constructed VisualTrans based on EgoDex~\cite{EgoDex}, which contains a large collection of first-person operation videos. The construction process involves task selection, image cleaning, sample filtering, metadata annotation, QA generation and verification, as illustrated in Fig.~\ref{fig:visualtrans-pipeline}.

\textbf{Task Selection.} The EgoDex dataset comprises 118 categories of real-world manipulation tasks. Among them, some tasks involve only single-object state changes (\emph{e.g.}, twisting a bottle cap or folding a sheet of paper) or highly localized operations (\emph{e.g.}, washing a plate). These tasks lack significant spatial displacement or structural reconfiguration, resulting in overly simple transformations that fall outside the scope of our target VTR definition. We manually selected 12 representative real-world task scenarios that exhibit key characteristics of visual transformation, including multi-object interaction, substantial spatial rearrangement, and complex procedural steps (\emph{e.g.}, building block assembly, desktop organization). For each selected instance, we extracted the initial and final frames from the corresponding video clips to construct image pairs, which serve as the fundamental data units.

\textbf{Image Cleaning and Sample Filtering.} Since image pairs are directly extracted from raw video frames, some suffer from issues such as motion blur or occlusion (particularly from operator arms obscuring key objects), which can negatively impact model evaluation. Additionally, certain samples lack sufficient procedural complexity or structural variation, limiting their value for reasoning-based tasks. We utilized the o3 model~\cite{gpto3-o4-mini} to automatically assess image quality, filtering out samples with severe occlusion, blur, or insufficient structural change. After filtering, approximately 85\% of high-quality samples were retained.

\textbf{Automated Metadata Annotation.} Fine-grained spatial metadata is critical for supporting compositional reasoning. Given that original images often contain cluttered backgrounds and distractors, direct structural parsing is nontrivial. To address this, we first employed Grounding DINO~\cite{GroundingDino} to detect task-relevant objects (\emph{e.g.}, block stacks, tableware clusters) and generate bounding boxes. These localized regions were then cropped and provided as auxiliary view inputs to enhance spatial grounding. Building on this, we used Gemini 2.5 Pro~\cite{gemini25pro} to automatically generate structured scene metadata, including task context type, operation completion status, object lists, absolute positions, pairwise spatial relations, stacking hierarchies, and object counts. Metadata schemas are dynamically adapted to task scenarios; not all fields are present in every instance.

\textbf{Question-Answer Generation and Verification.} For each fine-grained subtask type, we crafted QA templates with multiple placeholder slots \texttt{[MASK]}. These slots are programmatically populated using structured metadata to synthesize task-relevant questions. We randomly sampled 500 image pairs and generated QA instances spanning all three major reasoning dimensions. To ensure reliability, we manually reviewed all generated QA pairs. This involved correcting misaligned labels, removing ambiguous questions, and refining incorrect or underspecified answers. After quality control, a final set of 472 high-quality QA pairs was retained for benchmark release.

\subsection{Benchmark Statistics}

\begin{figure}[!t]
    \centering
    \includegraphics[width=1.0\linewidth]{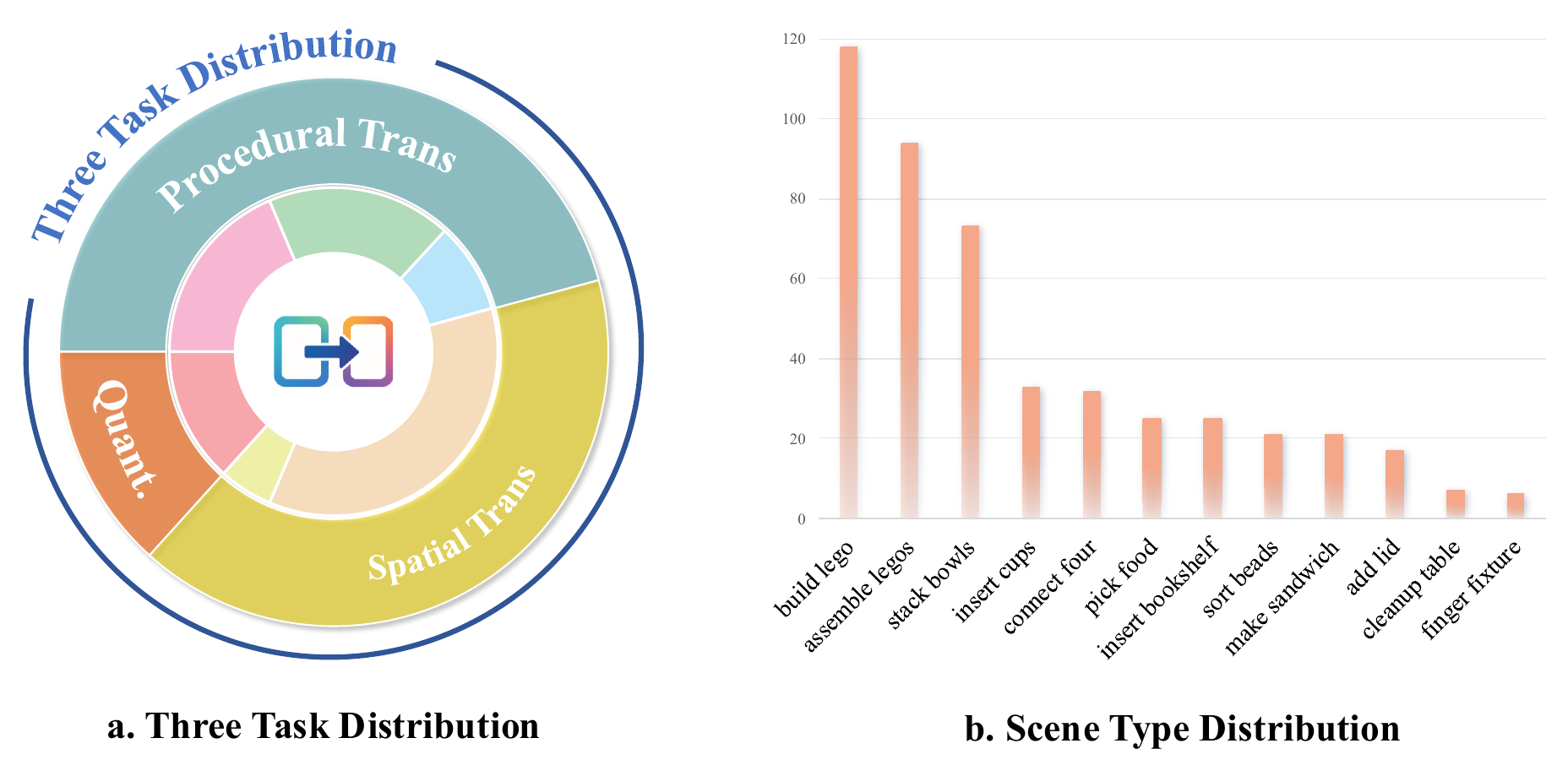}
    \caption{\textbf{Overview of VisualTrans.} (a) Task distribution across three reasoning types. (b) Scene type distribution from real-world manipulation.}
    \label{fig:data_statistics}
\end{figure}

Fig.~\ref{fig:data_statistics} presents an overview of the VisualTrans benchmark. It contains 472 high-quality QA samples spanning 12 real-world manipulation scenarios, please refer to Appendix for more details. VisualTrans is designed to systematically cover three core reasoning dimensions:
\begin{itemize}
    \item \textbf{Spatial Transformation}, including \textit{Fine-Grained Spatial} and \textit{Global Spatial} understanding.
    \item \textbf{Procedural Transformation}, involving \textit{Intermediate State Recognition}, \textit{Action Causality Reasoning}, and \textit{Transformation Sequence Planning}.
    \item \textbf{Quantitative Transformation}, requiring perception of object count changes.
\end{itemize}

The dataset also supports three QA formats: multiple choice, open-ended counting, and target enumeration, enabling diverse evaluation protocols across tasks.

\section{Experiments}

\subsection{Experimental Setup}

\textbf{Baseline Models.} We systematically evaluated 17 currently mainstream open-source and closed-source VLMs on VisualTrans. For closed-source models, we considered GPT-4o~\cite{GPT-4o}, GPT-4.1~\cite{gpt-4.1}, o3~\cite{gpto3-o4-mini}, Gemini-2.5-Pro~\cite{gemini25pro}, and Claude-3.7-Sonnet~\cite{Claude-3.7-Sonnet}. For open-source models, we evaluated InternVL3~\cite{Internvl3}, Qwen-2.5VL~\cite{Qwen2.5-VL}, Pixtral~\cite{agrawal2024pixtral}, Llama3.2-Vision~\cite{llama3.2-vision}, Llama-4-Maverick~\cite{Llama-4-maverick}, Phi-4-Multimodal~\cite{abouelenin2025phi}, Kimi-VL-A3B-Thinking~\cite{team2025kimi} and GLM-4.1V-9B-Thinking~\cite{hong2025glm}. All models were evaluated under a zero-shot setting.

\textbf{Evaluation Metric.}
We adopt a rule-based evaluation process. To assess reasoning ability, models are required to first generate a detailed thinking process followed by a final answer. Only the final answer is evaluated. Two matching strategies are used: \textit{Scalar-level Exact Match}, where a scalar output (\emph{e.g.}, a letter or integer) must exactly match the reference; and \textit{Set-level Exact Match}, where the predicted set must fully match the reference set in both elements and cardinality, with no omissions or redundancies. For multiple-choice and counting questions, we report accuracy based on \textit{Scalar-level Exact Match}. For target enumeration questions, we use accuracy based on \textit{Set-level Exact Match}.

\subsection{Quantitative Results}
We evaluate a range of proprietary and open-source VLMs on the VisualTrans benchmark. The results across spatial, procedural, and quantitative tasks are shown in Tab.~\ref{tab:visualtrans-results}. Based on the evaluation, we observe the following key patterns:

\begin{table*}[ht]
\centering
\renewcommand{\arraystretch}{1.05}
\begin{tabularx}{\linewidth}{l|YYY|YYYY|Y|Y}
\toprule
\multirow{2}{*}{\textbf{Model}} & \multicolumn{3}{c|}{\textbf{Spatial Transformation}} & \multicolumn{4}{c|}{\textbf{Procedural Transformation}} & \textbf{Quant.} & \textbf{Overall} \\
\cmidrule(lr){2-4} \cmidrule(lr){5-8}
& Fine & Global & Mean & Interm. & Causal & Plan & Mean & Count & Mean \\
\midrule
\rowcolor[HTML]{F2F2F2} \multicolumn{10}{l}{\textbf{Proprietary Models}} \\ \midrule

GPT-4o-2024-11-20 & 30.36 & 30.00 & 30.18 & \underline{31.82} & 36.05 & 45.24 & 37.70 & 57.14 & 36.22 \\
GPT-4.1 & \underline{51.79} & 38.00 & 44.90 & 27.27 & 29.07 & \underline{59.52} & 38.62 & \underline{65.08} & 44.47 \\
Claude-3.7-Sonnet & 44.05 & \underline{52.00} & 48.03 & 21.59 & \underline{55.81} & 57.14 & 44.85 & 47.62 & 44.47 \\
Gemini-2.5-Pro & 48.21 & 48.00 & \underline{48.11} & \textbf{54.55} & \underline{55.81} & \textbf{78.57} & \textbf{62.98} & 61.90 & \underline{54.93} \\
o3 & \textbf{58.33} & \textbf{84.00} & \textbf{71.17} & 23.86 & \textbf{62.79} & \textbf{78.57} & \underline{55.07} & \textbf{79.37} & \textbf{59.96} 
\\\midrule
\rowcolor[HTML]{F2F2F2} \multicolumn{10}{l}{\textbf{Open-Source Models}} \\ \midrule

Phi-4-Multimodal-Instruct & 19.05 & 20.00 & 19.53 & - & 32.56 & 14.29 & 23.43 & 23.81 & 18.31 \\
Llama-3.2-11b-Vision-Instruct & 12.50 & 24.00 & 18.25 & 19.32 & 26.74 & 23.81 & 23.29 & 14.29 & 18.51 \\
Qwen2.5-VL-7B & 17.86 & 24.00 & 20.93 & - & 26.74 & 26.19 & 26.47 & 36.51 & 19.92 \\
Pixtral-12B & 19.05 & 32.00 & 25.53 & 18.18 & 22.09 & 28.57 & 22.95 & 30.16 & 22.94 \\
Kimi-VL-A3B-Thinking & 20.24 & 40.00 & 30.12 & 20.45 & 32.56 & 30.95 & 27.99 & 20.63 & 25.35 \\
Pixtral-Large & 20.24 & 28.00 & 24.12 & 19.32 & 29.07 & 42.86 & 30.42 & 30.16 & 25.55 \\
Qwen2.5-VL-72B & 22.02 & 26.00 & 24.01 & \underline{26.14} & 30.23 & 33.33 & 29.90 & 46.03 & 28.57 \\
InternVL3-14B & 18.45 & \underline{42.00} & 30.23 & \underline{26.14} & 24.42 & \underline{45.24} & 31.93 & 42.86 & 28.57 \\
Qwen2.5-VL-32B & 23.21 & 14.00 & 18.61 & 23.86 & \textbf{36.05} & \underline{45.24} & \underline{35.05} & 44.44 & 29.18 \\
Llama-4-Maverick & \textbf{31.55} & \textbf{44.00} & \textbf{37.78} & 19.32 & 26.74 & \textbf{50.00} & 32.02 & 39.68 & 32.39 \\
GLM-4.1V-9B-Thinking & \underline{29.76} & 32.00 & 30.88 & \textbf{28.41} & 29.07 & 42.86 & 33.45 & \underline{47.62} & \underline{33.00} \\
InternVL3-78B & 26.79 & \underline{42.00} & \underline{34.40} & \textbf{28.41} & \underline{33.72} & \textbf{50.00} & \textbf{37.38} & \textbf{52.38} & \textbf{35.01} \\

\bottomrule
\end{tabularx}
\caption{\textbf{Performance of State-of-the-art VLMs on the VisualTrans Benchmark.}  
\textbf{Fine} and \textbf{Global} evaluate \textit{fine-grained} and \textit{global spatial understanding};  
\textbf{Interm.}, \textbf{Causal}, and \textbf{Plan} correspond to \textit{intermediate state recognition}, \textit{action causality reasoning}, and \textit{transformation sequence planning};  
\textbf{Quant.} denotes \textit{quantitative counting}.  
"--" indicates unsupported input formats.  
The final column shows the overall mean accuracy.}
\label{tab:visualtrans-results}
\end{table*}

\textit{\textbf{Limited VLMs performance.}} While recent VLMs have demonstrated progress in visual reasoning, their overall performance on our real-world, multi-dimensional benchmark VisualTrans remains limited. Most models achieve only 30\%–50\% accuracy, struggling to effectively model spatial structures, quantify object changes, and reason about dynamic processes. These results highlight persistent challenges in complex state understanding and causal inference.

\textit{\textbf{Closed models perform better.}}
A clear and consistent performance gap is observed between closed-source and open-source VLMs. Closed models such as o3 (59.96\%) and Gemini-2.5-Pro (54.93\%) lead the benchmark by a large margin, with GPT-4o and Claude-3.7-Sonnet also performing reliably across all subtask categories. In contrast, the best open-source model, InternVL3-78B, reaches only 35.01\%, while others like Qwen2.5-VL, Pixtral, and LLaMA demonstrate weak performance across spatial, procedural, and quantitative tasks. These results indicate that open-source models still lag significantly behind their closed-source counterparts when faced with complex, real-world visual transformation challenges.

\textit{\textbf{Strong perception of quantitative changes.}} Among all subtasks, quantitative perception proves the most tractable. o3 achieves the highest accuracy of 79.37\% in this category, leveraging strong multimodal alignment capabilities. In general, models perform significantly better on quantitative transformations than on spatial or procedural ones, suggesting that current VLMs are more adept at detecting salient object count changes while struggling with more complex relational reasoning or dynamic process modeling.

\textit{\textbf{Limited procedural modeling ability.}} Performance on procedural reasoning tasks remains weak. In the \textit{intermediate state recognition} subtask, no model exceeds 55\% accuracy, highlighting challenges in temporal modeling and cross-state understanding. For \textit{transformation sequence planning}, o3 and Gemini-2.5-Pro reach 78.57\%, showing partial competence in temporal reasoning. In \textit{action causality reasoning}, o3, Claude-3.7-Sonnet, and Gemini-2.5-Pro surpass 50\%, while most open-source models stay near 30\%. These results underscore current limitations in modeling causal dependencies and long-horizon visual logic.

\textit{\textbf{Layered understanding of spatial changes.}} Spatial reasoning performance varies by granularity. Most models struggle with \textit{fine-grained} spatial changes such as stacking, overlapping, or support—Qwen2.5-VL-7B and InternVL3-14B achieve only 17.86\% and 18.45\% accuracy, respectively. In contrast, performance improves on \textit{global} spatial changes, indicating that current models are better at capturing coarse structural shifts than subtle local variations.

\subsection{Error Analyses}

\begin{figure*}[t!]
    \centering
    \includegraphics[width=1.0\linewidth]{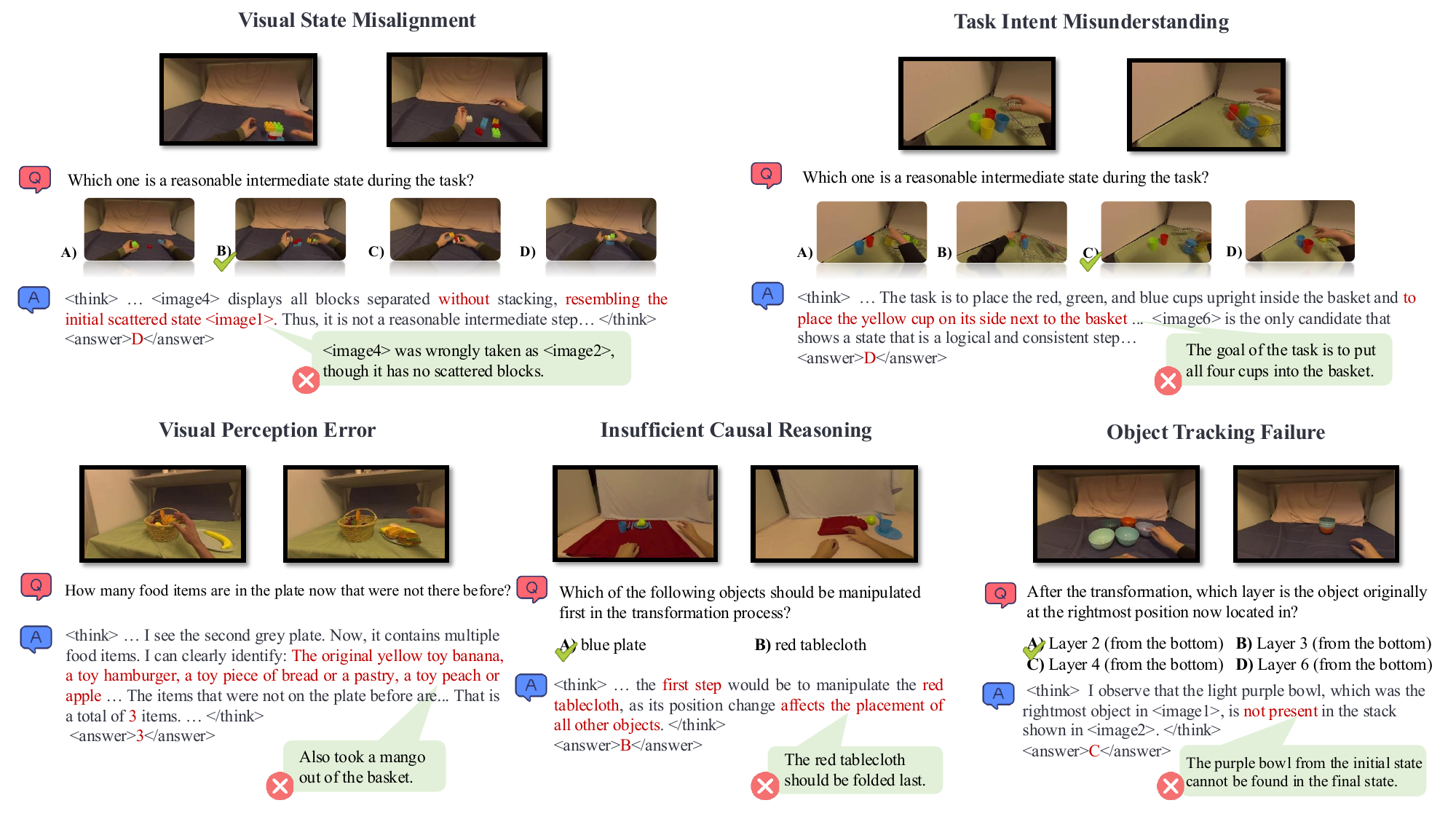}
    \caption{\textbf{Error Analyses}. This figure illustrates the five error types identified in VisualTrans. For each error type, a representative example is provided, with the specific nature of the error clearly indicated.}
    \label{fig:Error Analysis}
\end{figure*}

To analyze the limitations of current VLMs in VTR, we conducted a systematic error analysis and identified five recurring failure patterns, as shown in Fig.~\ref{fig:Error Analysis}. These highlight key challenges in multi-step reasoning, temporal modeling, and object relationship understanding, especially in tasks like intermediate state recognition and sequence planning.

\textit{\textbf{Visual State Misalignment.}} This error arises when models fail to accurately map image indices to their corresponding visual states, resulting in mismatches between referenced objects and actual image content. For instance, in intermediate state recognition tasks, a scattered block configuration (\emph{e.g.}, \textit{image4}) may be incorrectly interpreted as an earlier step (\emph{e.g.}, \textit{image2}), despite lacking visual consistency with the initial state. Such confusions often stem from models overlooking subtle structural cues, leading to incorrect assumptions about temporal order or missing intermediate steps altogether. These errors are not confined to visual transformation tasks—they also frequently appear in other multi-image scenarios (typically involving 5–6 frames), suggesting broader challenges in cross-image alignment, temporal modeling, and consistent multi-step representation.

\textit{\textbf{Task intent misunderstanding.}} Models fail to correctly interpret the task goals or state transformation trends implied in image sequences, resulting in generated operation steps that deviate from reasonable goal-oriented paths. In some cases, model predictions are affected by irrelevant interfering factors in images, such as background objects or non-task-related actions, which mislead their reasoning direction. This reflects the models’ shortcomings in temporal understanding and task intent modeling, as well as a lack of grasp and abstraction of key state evolution rules in visual transformation processes.

\textit{\textbf{Visual perception error.}} Models show deficiencies in perceiving objects and their spatial structures, including misrecognition of basic visual elements such as object categories, quantities, and positions, as well as limited understanding of spatial relationships between components in the scene. For instance, in quantitative change perception tasks, models may fail to perceive all food items on a plate, leading to counting errors; similarly, in block-building scenarios, they often miss the changed position of a specific block.

\textit{\textbf{Insufficient causal reasoning.}} Models lack the ability to model complex causal processes and struggle to effectively capture key fine-grained changes and their causal relationships during state transitions. Particularly in "intermediate state recognition" tasks, the reasoning process overlooks fine-grained causal cues in state evolution; in "transformation sequence planning" tasks, they also fail to establish multi-stage, complex causal chains, which limits in-depth understanding of continuous visual processes.

\textit{\textbf{Object tracking failure.}} Models struggles to consistently identify and track the same object across different image states, particularly when the scene layout changes—such as background shifts, the introduction of occluding objects, or the rearrangement of spatial relations. During the transformation process, appearance cues of the object may become weakened, making it difficult for the model to establish cross-frame identity associations, which leads to recognition failure. For example, a purple bowl can be correctly identified when the bowls are separated, but becomes unlocatable when they are stacked together.

In general, model failures across transformation tasks can be attributed to three key factors:
\textbf{\textit{(1) Deficient visual perception of complex structures.}}
Models often struggle to perceive intricate spatial relationships. Before transformation, object layouts can be complex; after transformation, occlusion, overlap, or support relations require detailed spatial and temporal understanding. Errors at this stage tend to propagate, affecting subsequent planning and action prediction. \textbf{\textit{(2) Limited causal reasoning and physical understanding.}}
Models perform well in tasks with simple causal logic—\emph{e.g.}, inferring that upper objects must be removed before accessing lower ones, or understanding that screws must be removed before detaching parts. However, they often fail to reason over complex, multi-step transformations, especially in open-source models, which lack fine-grained causal modeling. \textbf{\textit{(3) Weak temporal grounding and dynamic process modeling.}}
Failures in task intent understanding and object tracking hinder dynamic reasoning. Models frequently misinterpret goals or transformation trends, and struggle to maintain object identity across frames—particularly under layout changes—leading to degraded downstream decisions.

\section{Conclusion}
We present VisualTrans, the first comprehensive benchmark for evaluating VTR in real-world human-object interaction scenarios. It covers three core reasoning dimensions—spatial, procedural, and quantitative—through six subtask types derived from 12 semantically diverse manipulation tasks. Our scalable construction pipeline leverages first-person videos, multimodal model-assisted annotation, and human verification, resulting in 472 high-quality, interpretable QA pairs. Empirical results on state-of-the-art VLMs show that while current models perform well on static spatial tasks, they struggle with dynamic and causal reasoning subtasks such as intermediate state recognition and transformation sequence planning. These findings reveal fundamental limitations in temporal and transformation modeling, highlighting key gaps in current VTR capabilities.


\section{Acknowledgments}
This work is supported by the Natural Science Foundation of China under Grant Nos. 72225011and 72434005.

\bibliography{aaai2026}

\begin{thebibliography}{58}
\providecommand{\natexlab}[1]{#1}

\bibitem[{Abouelenin et~al.(2025)Abouelenin, Ashfaq, Atkinson, Awadalla, Bach, Bao, Benhaim, Cai, Chaudhary, Chen et~al.}]{abouelenin2025phi}
Abouelenin, A.; Ashfaq, A.; Atkinson, A.; Awadalla, H.; Bach, N.; Bao, J.; Benhaim, A.; Cai, M.; Chaudhary, V.; Chen, C.; et~al. 2025.
\newblock Phi-4-mini technical report: Compact yet powerful multimodal language models via mixture-of-loras.
\newblock \emph{arXiv preprint arXiv:2503.01743}.

\bibitem[{Agrawal et~al.(2024)Agrawal, Antoniak, Hanna, Bout, Chaplot, Chudnovsky, Costa, De~Monicault, Garg, Gervet et~al.}]{agrawal2024pixtral}
Agrawal, P.; Antoniak, S.; Hanna, E.~B.; Bout, B.; Chaplot, D.; Chudnovsky, J.; Costa, D.; De~Monicault, B.; Garg, S.; Gervet, T.; et~al. 2024.
\newblock Pixtral 12B.
\newblock \emph{arXiv preprint arXiv:2410.07073}.

\bibitem[{{Anthropic}(2025)}]{Claude-3.7-Sonnet}
{Anthropic}. 2025.
\newblock Claude 3.7 Sonnet and Claude Code.
\newblock \url{https://www.anthropic.com/news/claude-3-7-sonnet}.
\newblock Accessed: 2025-02-25.

\bibitem[{Bai et~al.(2025)Bai, Chen, Liu, Wang, Ge, Song, Dang, Wang, Wang, Tang et~al.}]{Qwen2.5-VL}
Bai, S.; Chen, K.; Liu, X.; Wang, J.; Ge, W.; Song, S.; Dang, K.; Wang, P.; Wang, S.; Tang, J.; et~al. 2025.
\newblock Qwen2. 5-vl technical report.
\newblock \emph{arXiv preprint arXiv:2502.13923}.

\bibitem[{Chen et~al.(2025)Chen, Zhang, Zhu, Liu, Gao, Xiong, Li, and He}]{chenbring}
Chen, S.; Zhang, J.; Zhu, T.; Liu, W.; Gao, S.; Xiong, M.; Li, M.; and He, J. 2025.
\newblock Bring Reason to Vision: Understanding Perception and Reasoning through Model Merging.
\newblock In \emph{Forty-second International Conference on Machine Learning}.

\bibitem[{Cheng et~al.(2024)Cheng, Yin, Fu, Guo, Yang, Kautz, Wang, and Liu}]{Spatialrgpt}
Cheng, A.-C.; Yin, H.; Fu, Y.; Guo, Q.; Yang, R.; Kautz, J.; Wang, X.; and Liu, S. 2024.
\newblock Spatialrgpt: Grounded spatial reasoning in vision-language models.
\newblock \emph{Advances in Neural Information Processing Systems}, 37: 135062--135093.

\bibitem[{Goddu and Gopnik(2024)}]{goddu2024development}
Goddu, M.~K.; and Gopnik, A. 2024.
\newblock The development of human causal learning and reasoning.
\newblock \emph{Nature Reviews Psychology}, 3(5): 319--339.

\bibitem[{Gokhale et~al.(2019)Gokhale, Sampat, Fang, Yang, and Baral}]{BIRD}
Gokhale, T.; Sampat, S.; Fang, Z.; Yang, Y.; and Baral, C. 2019.
\newblock Cooking with blocks: A recipe for visual reasoning on image-pairs.
\newblock In \emph{Proceedings of the IEEE/CVF Conference on Computer Vision and Pattern Recognition Workshops}, 5--8.

\bibitem[{Google(2025)}]{gemini25pro}
Google. 2025.
\newblock Gemini 2.5 Pro Preview: even better coding performance.
\newblock \url{https://developers.googleblog.com/en/gemini-2-5-pro-io-improved-coding-performance/}.
\newblock Accessed: 2025-05-06.

\bibitem[{Han et~al.(2024)Han, Liu, Yuan, Pu, Wang, Song, and Huang}]{han2024latency}
Han, Y.; Liu, Z.; Yuan, Z.; Pu, Y.; Wang, C.; Song, S.; and Huang, G. 2024.
\newblock Latency-aware unified dynamic networks for efficient image recognition.
\newblock \emph{IEEE Transactions on Pattern Analysis and Machine Intelligence}, 46(12): 7760--7774.

\bibitem[{Hao et~al.(2025{\natexlab{a}})Hao, Zhang, Li, Cao, Hao, Cui, and Wang}]{hao2025tla}
Hao, P.; Zhang, C.; Li, D.; Cao, X.; Hao, X.; Cui, S.; and Wang, S. 2025{\natexlab{a}}.
\newblock Tla: Tactile-language-action model for contact-rich manipulation.
\newblock \emph{arXiv preprint arXiv:2503.08548}.

\bibitem[{Hao et~al.(2025{\natexlab{b}})Hao, Diao, Wei, Yang, Hao, Yin, Zhang, Li, Zhao, and Liu}]{HAO2025103018}
Hao, X.; Diao, Y.; Wei, M.; Yang, Y.; Hao, P.; Yin, R.; Zhang, H.; Li, W.; Zhao, S.; and Liu, Y. 2025{\natexlab{b}}.
\newblock MapFusion: A novel BEV feature fusion network for multi-modal map construction.
\newblock \emph{Information Fusion}, 119: 103018.

\bibitem[{Hong et~al.(2025)Hong, Yu, Gu, Wang, Gan, Tang, Cheng, Qi, Ji, Pan et~al.}]{hong2025glm}
Hong, W.; Yu, W.; Gu, X.; Wang, G.; Gan, G.; Tang, H.; Cheng, J.; Qi, J.; Ji, J.; Pan, L.; et~al. 2025.
\newblock GLM-4.1 V-Thinking: Towards Versatile Multimodal Reasoning with Scalable Reinforcement Learning.
\newblock \emph{arXiv preprint arXiv:2507.01006}.

\bibitem[{Hong et~al.(2021)Hong, Lan, Pang, Guo, and Cheng}]{TRANCE}
Hong, X.; Lan, Y.; Pang, L.; Guo, J.; and Cheng, X. 2021.
\newblock Transformation driven visual reasoning.
\newblock In \emph{Proceedings of the IEEE/CVF Conference on computer vision and pattern recognition}, 6903--6912.

\bibitem[{Hong et~al.(2023)Hong, Lan, Pang, Guo, and Cheng}]{TRANCO}
Hong, X.; Lan, Y.; Pang, L.; Guo, J.; and Cheng, X. 2023.
\newblock Visual reasoning: From state to transformation.
\newblock \emph{IEEE Transactions on Pattern Analysis and Machine Intelligence}, 45(9): 11352--11364.

\bibitem[{Hoque et~al.(2025)Hoque, Huang, Yoon, Sivapurapu, and Zhang}]{EgoDex}
Hoque, R.; Huang, P.; Yoon, D.~J.; Sivapurapu, M.; and Zhang, J. 2025.
\newblock EgoDex: Learning Dexterous Manipulation from Large-Scale Egocentric Video.
\newblock \emph{arXiv preprint arXiv:2505.11709}.

\bibitem[{Hu et~al.(2024)Hu, Yang, Jiang, and Bai}]{hu2024prompting}
Hu, Z.; Yang, P.; Jiang, Y.; and Bai, Z. 2024.
\newblock Prompting large language model with context and pre-answer for knowledge-based VQA.
\newblock \emph{Pattern Recognition}, 151: 110399.

\bibitem[{Huang et~al.(2025)Huang, Xu, Wang, Wang, Liang, Wang, Zhang, Wei, Zhang, Huang et~al.}]{huang2025foundation}
Huang, J.; Xu, Y.; Wang, Q.; Wang, Q.~C.; Liang, X.; Wang, F.; Zhang, Z.; Wei, W.; Zhang, B.; Huang, L.; et~al. 2025.
\newblock Foundation models and intelligent decision-making: Progress, challenges, and perspectives.
\newblock \emph{The Innovation}.

\bibitem[{Hurst et~al.(2024{\natexlab{a}})Hurst, Lerer, Goucher, Perelman, Ramesh, Clark, Ostrow, Welihinda, Hayes, Radford et~al.}]{hurst2024gpt}
Hurst, A.; Lerer, A.; Goucher, A.~P.; Perelman, A.; Ramesh, A.; Clark, A.; Ostrow, A.; Welihinda, A.; Hayes, A.; Radford, A.; et~al. 2024{\natexlab{a}}.
\newblock Gpt-4o system card.
\newblock \emph{arXiv preprint arXiv:2410.21276}.

\bibitem[{Hurst et~al.(2024{\natexlab{b}})Hurst, Lerer, Goucher, Perelman, Ramesh, Clark, Ostrow, Welihinda, Hayes, Radford et~al.}]{GPT-4o}
Hurst, A.; Lerer, A.; Goucher, A.~P.; Perelman, A.; Ramesh, A.; Clark, A.; Ostrow, A.; Welihinda, A.; Hayes, A.; Radford, A.; et~al. 2024{\natexlab{b}}.
\newblock Gpt-4o system card.
\newblock \emph{arXiv preprint arXiv:2410.21276}.

\bibitem[{Ji et~al.(2025{\natexlab{a}})Ji, Liu, Zhang, Zhang, Zhao, Hao, Zhou, Zhang, and Zheng}]{advlora}
Ji, Y.; Liu, Y.; Zhang, Z.; Zhang, Z.; Zhao, Y.; Hao, X.; Zhou, G.; Zhang, X.; and Zheng, X. 2025{\natexlab{a}}.
\newblock Enhancing adversarial robustness of vision-language models through low-rank adaptation.
\newblock In \emph{Proceedings of the 2025 International Conference on Multimedia Retrieval}, 550--559.

\bibitem[{Ji et~al.(2025{\natexlab{b}})Ji, Tan, Shi, Hao, Zhang, Zhang, Wang, Zhao, Mu, An et~al.}]{robobrain}
Ji, Y.; Tan, H.; Shi, J.; Hao, X.; Zhang, Y.; Zhang, H.; Wang, P.; Zhao, M.; Mu, Y.; An, P.; et~al. 2025{\natexlab{b}}.
\newblock Robobrain: A unified brain model for robotic manipulation from abstract to concrete.
\newblock In \emph{Proceedings of the Computer Vision and Pattern Recognition Conference}, 1724--1734.

\bibitem[{Johnson et~al.(2017)Johnson, Hariharan, Van Der~Maaten, Fei-Fei, Lawrence~Zitnick, and Girshick}]{Clevr}
Johnson, J.; Hariharan, B.; Van Der~Maaten, L.; Fei-Fei, L.; Lawrence~Zitnick, C.; and Girshick, R. 2017.
\newblock Clevr: A diagnostic dataset for compositional language and elementary visual reasoning.
\newblock In \emph{Proceedings of the IEEE conference on computer vision and pattern recognition}, 2901--2910.

\bibitem[{Li et~al.(2024)Li, Jin, Sun, Yu, Shi, Hao, Hao, Liu, Sun, Zhang et~al.}]{li2024foundation}
Li, D.; Jin, Y.; Sun, Y.; Yu, H.; Shi, J.; Hao, X.; Hao, P.; Liu, H.; Sun, F.; Zhang, J.; et~al. 2024.
\newblock What foundation models can bring for robot learning in manipulation: A survey.
\newblock \emph{arXiv preprint arXiv:2404.18201}.

\bibitem[{Li et~al.(2025)Li, Luo, Zhang, Qiu, Huang, and Wei}]{li2025vocot}
Li, Z.; Luo, R.; Zhang, J.; Qiu, M.; Huang, X.-J.; and Wei, Z. 2025.
\newblock VoCoT: Unleashing Visually Grounded Multi-Step Reasoning in Large Multi-Modal Models.
\newblock In \emph{Proceedings of the 2025 Conference of the Nations of the Americas Chapter of the Association for Computational Linguistics: Human Language Technologies (Volume 1: Long Papers)}, 3769--3798.

\bibitem[{Lin et~al.(2024)Lin, Yin, Ping, Molchanov, Shoeybi, and Han}]{Vila}
Lin, J.; Yin, H.; Ping, W.; Molchanov, P.; Shoeybi, M.; and Han, S. 2024.
\newblock Vila: On pre-training for visual language models.
\newblock In \emph{Proceedings of the IEEE/CVF conference on computer vision and pattern recognition}, 26689--26699.

\bibitem[{Liu et~al.(2024)Liu, Zeng, Ren, Li, Zhang, Yang, Jiang, Li, Yang, Su et~al.}]{GroundingDino}
Liu, S.; Zeng, Z.; Ren, T.; Li, F.; Zhang, H.; Yang, J.; Jiang, Q.; Li, C.; Yang, J.; Su, H.; et~al. 2024.
\newblock Grounding dino: Marrying dino with grounded pre-training for open-set object detection.
\newblock In \emph{European conference on computer vision}, 38--55. Springer.

\bibitem[{Liu et~al.(2025)Liu, Zhai, Du, Chen, Cao, Gao, Wang, Li, Wang, Fang et~al.}]{liu2025guardreasoner}
Liu, Y.; Zhai, S.; Du, M.; Chen, Y.; Cao, T.; Gao, H.; Wang, C.; Li, X.; Wang, K.; Fang, J.; et~al. 2025.
\newblock Guardreasoner-vl: Safeguarding vlms via reinforced reasoning.
\newblock \emph{arXiv preprint arXiv:2505.11049}.

\bibitem[{Lyu et~al.(2025)Lyu, Chen, Ji, and Xu}]{egoprompt}
Lyu, H.; Chen, C.; Ji, Y.; and Xu, C. 2025.
\newblock EgoPrompt: Prompt Pool Learning for Egocentric Action Recognition.
\newblock \emph{arXiv preprint arXiv:2508.03266}.

\bibitem[{Meng et~al.(2024)Meng, Shao, Lu, Gao, Zhang, Qiao, and Luo}]{chartassistant}
Meng, F.; Shao, W.; Lu, Q.; Gao, P.; Zhang, K.; Qiao, Y.; and Luo, P. 2024.
\newblock ChartAssistant: A Universal Chart Multimodal Language Model via Chart-to-Table Pre-training and Multitask Instruction Tuning.
\newblock In \emph{ACL (Findings)}.

\bibitem[{{Meta}(2025{\natexlab{a}})}]{Llama-4-maverick}
{Meta}. 2025{\natexlab{a}}.
\newblock The Llama 4 herd: The beginning of a new era of natively multimodal AI innovation.
\newblock \url{https://ai.meta.com/blog/llama-4-multimodal-intelligence/}.
\newblock Accessed: 2025-04-05.

\bibitem[{{Meta}(2025{\natexlab{b}})}]{llama3.2-vision}
{Meta}. 2025{\natexlab{b}}.
\newblock Llama3.2-Vision Model Card.
\newblock \url{https://github.com/meta-llama/llama-models/blob/main/models/llama3_2/MODEL_CARD_VISION.md}.
\newblock Accessed: 2024-09-25.

\bibitem[{{OpenAI}(2025{\natexlab{a}})}]{gpt-4.1}
{OpenAI}. 2025{\natexlab{a}}.
\newblock Introducing GPT-4.1 in the API.
\newblock \url{https://openai.com/index/gpt-4-1/}.
\newblock Accessed: 2025-04-14.

\bibitem[{{OpenAI}(2025{\natexlab{b}})}]{gpto3-o4-mini}
{OpenAI}. 2025{\natexlab{b}}.
\newblock OpenAI o3 and o4-mini System Card.
\newblock \url{https://openai.com/index/introducing-o3-and-o4-mini/}.
\newblock Accessed: 2025-04-16.

\bibitem[{Park et~al.(2019)}]{clevr_change}
Park, D.~H.; et~al. 2019.
\newblock Robust change captioning.
\newblock In \emph{Proceedings of the IEEE/CVF International Conference on Computer Vision}, 4624--4633.

\bibitem[{Piaget(2013)}]{Piaget}
Piaget, J. 2013.
\newblock \emph{The construction of reality in the child}.
\newblock Routledge.

\bibitem[{Qi et~al.(2025)Qi, Ding, Wang, Bai, Lv, Hong, Xu, Hou, Li, Dong, and Tang}]{CogCoM}
Qi, J.; Ding, M.; Wang, W.; Bai, Y.; Lv, Q.; Hong, W.; Xu, B.; Hou, L.; Li, J.; Dong, Y.; and Tang, J. 2025.
\newblock CogCoM: A Visual Language Model with Chain-of-Manipulations Reasoning.
\newblock In \emph{The Thirteenth International Conference on Learning Representations}.

\bibitem[{Qin et~al.()Qin, Shi, Yu, Wang, Zhou, Li, Yin, Liu, Sheng, Shao et~al.}]{worldsimbench}
Qin, Y.; Shi, Z.; Yu, J.; Wang, X.; Zhou, E.; Li, L.; Yin, Z.; Liu, X.; Sheng, L.; Shao, J.; et~al. ????
\newblock WorldSimBench: Towards Video Generation Models as World Simulators.
\newblock In \emph{Forty-second International Conference on Machine Learning}.

\bibitem[{Ray et~al.(2024)Ray, Duan, Tan, Bashkirova, Hendrix, Ehsani, Kembhavi, Plummer, Krishna, Zeng, and Saenko}]{SAT}
Ray, A.; Duan, J.; Tan, R.; Bashkirova, D.; Hendrix, R.; Ehsani, K.; Kembhavi, A.; Plummer, B.~A.; Krishna, R.; Zeng, K.-H.; and Saenko, K. 2024.
\newblock SAT: Spatial Aptitude Training for Multimodal Language Models.
\newblock \emph{CoRR}, abs/2412.07755.

\bibitem[{Shen et~al.(2025)}]{shen2025chart}
Shen, Q.; et~al. 2025.
\newblock From Chart to QA Pairs: A Context-Aware Generation Framework for Chart-Containing Documents.
\newblock In \emph{Proceedings of the 2025 2nd International Conference on Generative Artificial Intelligence and Information Security}, 101--107.

\bibitem[{Song et~al.(2025)Song, Blukis, Tremblay, Tyree, Su, and Birchfield}]{Robospatial}
Song, C.~H.; Blukis, V.; Tremblay, J.; Tyree, S.; Su, Y.; and Birchfield, S. 2025.
\newblock Robospatial: Teaching spatial understanding to 2d and 3d vision-language models for robotics.
\newblock In \emph{Proceedings of the Computer Vision and Pattern Recognition Conference}, 15768--15780.

\bibitem[{Tan et~al.(2025{\natexlab{a}})Tan, Hao, Chi, Lin, Lyu, Cao, Liang, Chen, Lyu, Peng et~al.}]{tan2025roboos}
Tan, H.; Hao, X.; Chi, C.; Lin, M.; Lyu, Y.; Cao, M.; Liang, D.; Chen, Z.; Lyu, M.; Peng, C.; et~al. 2025{\natexlab{a}}.
\newblock Roboos: A hierarchical embodied framework for cross-embodiment and multi-agent collaboration.
\newblock \emph{arXiv preprint arXiv:2505.03673}.

\bibitem[{Tan et~al.(2025{\natexlab{b}})Tan, Ji, Hao, Lin, Wang, Wang, and Zhang}]{reasonrft}
Tan, H.; Ji, Y.; Hao, X.; Lin, M.; Wang, P.; Wang, Z.; and Zhang, S. 2025{\natexlab{b}}.
\newblock Reason-rft: Reinforcement fine-tuning for visual reasoning.
\newblock \emph{arXiv preprint arXiv:2503.20752}.

\bibitem[{Tang et~al.(2019)Tang, Ding, Rao, Zheng, Zhang, Zhao, Lu, and Zhou}]{Coin}
Tang, Y.; Ding, D.; Rao, Y.; Zheng, Y.; Zhang, D.; Zhao, L.; Lu, J.; and Zhou, J. 2019.
\newblock Coin: A large-scale dataset for comprehensive instructional video analysis.
\newblock In \emph{Proceedings of the IEEE/CVF Conference on Computer Vision and Pattern Recognition}, 1207--1216.

\bibitem[{Tang et~al.(2025)Tang, Zhang, Hao, Wang, Wu, Wang, and Zhang}]{tang2025affordgrasp}
Tang, Y.; Zhang, S.; Hao, X.; Wang, P.; Wu, J.; Wang, Z.; and Zhang, S. 2025.
\newblock Affordgrasp: In-context affordance reasoning for open-vocabulary task-oriented grasping in clutter.
\newblock \emph{arXiv preprint arXiv:2503.00778}.

\bibitem[{Team et~al.(2025{\natexlab{a}})Team, Cao, Tan, Ji, Lin, Li, Cao, Wang, Zhou, Han et~al.}]{robobrain2.0}
Team, B.~R.; Cao, M.; Tan, H.; Ji, Y.; Lin, M.; Li, Z.; Cao, Z.; Wang, P.; Zhou, E.; Han, Y.; et~al. 2025{\natexlab{a}}.
\newblock RoboBrain 2.0 Technical Report.
\newblock \emph{arXiv preprint arXiv:2507.02029}.

\bibitem[{Team et~al.(2025{\natexlab{b}})Team, Du, Yin, Xing, Qu, Wang, Chen, Zhang, Du, Wei et~al.}]{team2025kimi}
Team, K.; Du, A.; Yin, B.; Xing, B.; Qu, B.; Wang, B.; Chen, C.; Zhang, C.; Du, C.; Wei, C.; et~al. 2025{\natexlab{b}}.
\newblock Kimi-vl technical report.
\newblock \emph{arXiv preprint arXiv:2504.07491}.

\bibitem[{Wang et~al.(2025)Wang, Wang, Cheng, Fei, Ding, Guo, Tao, and Qiu}]{Visuothink}
Wang, Y.; Wang, S.; Cheng, Q.; Fei, Z.; Ding, L.; Guo, Q.; Tao, D.; and Qiu, X. 2025.
\newblock {V}isuo{T}hink: Empowering {LVLM} Reasoning with Multimodal Tree Search.
\newblock In Che, W.; Nabende, J.; Shutova, E.; and Pilehvar, M.~T., eds., \emph{Proceedings of the 63rd Annual Meeting of the Association for Computational Linguistics (Volume 1: Long Papers)}, 21707--21719. Vienna, Austria: Association for Computational Linguistics.
\newblock ISBN 979-8-89176-251-0.

\bibitem[{Wu et~al.(2024)Wu, Wang, Tang, Wu, He, Ouyang, Torr, and Wu}]{DetToolChain}
Wu, Y.; Wang, Y.; Tang, S.; Wu, W.; He, T.; Ouyang, W.; Torr, P.; and Wu, J. 2024.
\newblock Dettoolchain: A new prompting paradigm to unleash detection ability of mllm.
\newblock In \emph{European Conference on Computer Vision}, 164--182. Springer.

\bibitem[{Xiu et~al.(2025)}]{ViDDAR}
Xiu, Y.; et~al. 2025.
\newblock ViDDAR: Vision language model-based task-detrimental content detection for augmented reality.
\newblock \emph{IEEE transactions on visualization and computer graphics}.

\bibitem[{Zhang et~al.(2025{\natexlab{a}})Zhang, Hao, Cao, Hao, Cui, and Wang}]{zhang2025vtla}
Zhang, C.; Hao, P.; Cao, X.; Hao, X.; Cui, S.; and Wang, S. 2025{\natexlab{a}}.
\newblock Vtla: Vision-tactile-language-action model with preference learning for insertion manipulation.
\newblock \emph{arXiv preprint arXiv:2505.09577}.

\bibitem[{Zhang et~al.(2025{\natexlab{b}})Zhang, Liu, Duan, Zheng, Yu, and Zhang}]{zhang2025embodied}
Zhang, C.; Liu, C.; Duan, S.; Zheng, X.; Yu, T.; and Zhang, J. 2025{\natexlab{b}}.
\newblock Embodied cognitive intelligence guided Moon sample collection.
\newblock \emph{The Innovation}.

\bibitem[{Zhang et~al.(2025{\natexlab{c}})Zhang, Hao, Xu, Zhang, Zhang, Wang, Zhang, Wang, Zhang, and Xu}]{zhang2025novel}
Zhang, L.; Hao, X.; Xu, Q.; Zhang, Q.; Zhang, X.; Wang, P.; Zhang, J.; Wang, Z.; Zhang, S.; and Xu, R.~M. 2025{\natexlab{c}}.
\newblock A novel memory representation via annotated semantic maps for vlm-based vision-and-language navigation.
\newblock \emph{arXiv preprint arXiv:2502.13451}.

\bibitem[{Zhao et~al.(2025)Zhao, Yuan, Xu, Hao, Zhang, Wu, Che, Liu, and Tang}]{zhao2025training}
Zhao, Y.; Yuan, J.; Xu, Z.; Hao, X.; Zhang, X.; Wu, K.; Che, Z.; Liu, C.~H.; and Tang, J. 2025.
\newblock Training-free Generation of Temporally Consistent Rewards from VLMs.
\newblock \emph{arXiv preprint arXiv:2507.04789}.

\bibitem[{Zhou et~al.(2025{\natexlab{a}})Zhou, An, Chi, Han, Rong, Zhang, Wang, Wang, Huang, Sheng et~al.}]{zhou2025roborefer}
Zhou, E.; An, J.; Chi, C.; Han, Y.; Rong, S.; Zhang, C.; Wang, P.; Wang, Z.; Huang, T.; Sheng, L.; et~al. 2025{\natexlab{a}}.
\newblock RoboRefer: Towards Spatial Referring with Reasoning in Vision-Language Models for Robotics.
\newblock \emph{arXiv preprint arXiv:2506.04308}.

\bibitem[{Zhou et~al.(2025{\natexlab{b}})Zhou, Su, Chi, Zhang, Wang, Huang, Sheng, and Wang}]{zhou2025code}
Zhou, E.; Su, Q.; Chi, C.; Zhang, Z.; Wang, Z.; Huang, T.; Sheng, L.; and Wang, H. 2025{\natexlab{b}}.
\newblock Code-as-monitor: Constraint-aware visual programming for reactive and proactive robotic failure detection.
\newblock In \emph{Proceedings of the Computer Vision and Pattern Recognition Conference}, 6919--6929.

\bibitem[{Zhu et~al.(2025)Zhu, Wang, Chen, Liu, Ye, Gu, Tian, Duan, Su, Shao et~al.}]{Internvl3}
Zhu, J.; Wang, W.; Chen, Z.; Liu, Z.; Ye, S.; Gu, L.; Tian, H.; Duan, Y.; Su, W.; Shao, J.; et~al. 2025.
\newblock Internvl3: Exploring advanced training and test-time recipes for open-source multimodal models.
\newblock \emph{arXiv preprint arXiv:2504.10479}.

\bibitem[{Zong et~al.(2025)Zong, Zhang, An, Li, Xu, Xu, Tu, Xing, and Dabeer}]{Ground-V}
Zong, Y.; Zhang, Q.; An, D.; Li, Z.; Xu, X.; Xu, L.; Tu, Z.; Xing, Y.; and Dabeer, O. 2025.
\newblock Ground-V: Teaching VLMs to Ground Complex Instructions in Pixels.
\newblock In \emph{Proceedings of the Computer Vision and Pattern Recognition Conference}, 24635--24645.

\end{thebibliography}


\clearpage
\newpage
\appendix
\section*{Appendix}
This supplementary material provides in-depth information about the construction, annotation, and evaluation procedures of the proposed \textbf{VisualTrans} benchmark. It complements the main paper by offering implementation details and additional resources to support reproducibility and further research. The document is organized as follows:

\begin{itemize}
    \item \textbf{Sec.~A} describes the raw video source used in VisualTrans and outlines its key properties.
    \item \textbf{Sec.~B} details the data curation pipeline, including task selection, image filtering, automatic metadata annotation, and question-answer (QA) pair generation.
    \item \textbf{Sec.~C} presents the evaluation setup, model inference configurations, and computational resources used throughout the benchmark.
    \item \textbf{Sec.~D} includes prompt templates used for data cleaning and metadata annotation, along with qualitative case studies from benchmark evaluation.
    \item \textbf{Sec.~E} summarizes future directions for expanding task diversity, incorporating multi-agent scenarios, and promoting data transparency through prompt sharing and case studies.

\end{itemize}

\begin{figure}[!ht]
    \centering
    \includegraphics[width=1.0\linewidth]{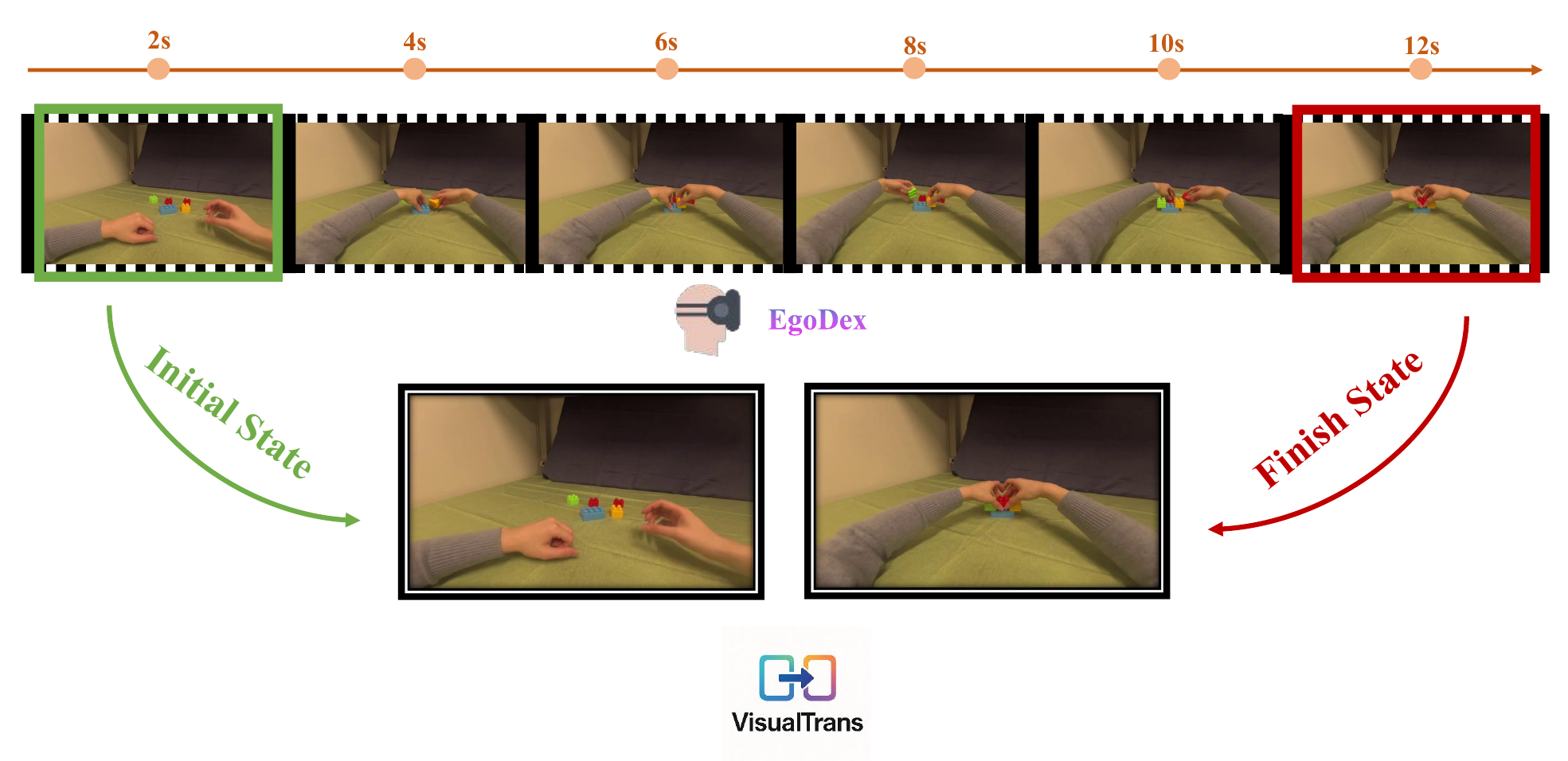}
    \caption{Egodex is a video collection featuring diverse manipulation scenarios. Each image pair in our benchmark corresponds to the first and last frame of a video, capturing the transformation process of a manipulation action.}
    \label{fig:Egodex}
\end{figure}

\section{A Details of Data Source}
\label{data_source}

This section introduces the raw video dataset leveraged in our benchmark and highlights its key properties.

EgoDex~\cite{EgoDex} is a large-scale egocentric dataset dedicated to capturing and analyzing human dexterous manipulation behaviors. VisualTrans is constructed by extracting representative image pairs from each manipulation episode in EgoDex, as illustrated in Fig.~\ref{fig:Egodex}. The dataset is released under a CC-BY-NC-ND license and encompasses approximately 2.0 TB of content, enabling a wide range of research applications in robotics and computer vision, including dexterous manipulation trajectory prediction and human-object interaction modeling.

EgoDex is collected using Apple Vision Pro devices running visionOS 2 and contains 829 hours of egocentric video, accompanied by rich multimodal annotations. The videos are recorded at a resolution of 1920$\times$1080 and a frame rate of 30 FPS, comprising approximately 90 million frames across 338{,}000 task episodes. The dataset covers 194 diverse tabletop manipulation tasks involving around 500 everyday household objects.

In addition to video, EgoDex provides synchronized 3D pose annotations, including the 3D positions and orientations of 25 joints for each hand and all upper-body joints. It also includes camera intrinsics and extrinsics, pose prediction confidence scores, and natural language annotations describing each manipulation episode. Task types are categorized into: (1) reversible tasks (e.g., \textit{charging} and \textit{uncharging}), (2) reset-free tasks that do not require reinitialization (e.g., throwing and catching a ball), and (3) reset-required tasks that involve restoring the initial scene configuration (e.g., decluttering a desk).

\section{B Details of Data Curation}

This section details the pipeline used to construct the VisualTrans benchmark, including task selection, image filtering, metadata annotation, and QA generation.

\subsection{Task Selection}

We begin by manually downloading and organizing the complete video resources provided in the EgoDex dataset, which encompasses 118 real-world manipulation task categories. An initial inspection reveals that a substantial portion of these tasks involves either minimal object state changes (e.g., twisting a bottle cap, folding paper) or low-dynamic, localized actions (e.g., wiping, cleaning). Such tasks typically lack meaningful spatial structure reconfiguration or multi-object interactions, which are essential for evaluating \textit{visual transformation} reasoning.

To ensure the benchmark captures rich visual reasoning signals, we apply the following selection criteria:

\begin{itemize}
\item \textbf{Multi-object involvement:} The manipulation involves at least two independently identifiable objects;
\item \textbf{Significant spatial displacement:} Objects undergo noticeable changes in position or configuration between the initial and final states;
\item \textbf{Process complexity and temporal structure:} The task comprises at least two discernible and sequential manipulation steps.
\end{itemize}

Based on these criteria, we manually curate 12 representative manipulation scenarios that exhibit salient visual transitions and spatial relationship changes:

\begin{itemize}
  \item \texttt{stack\_unstack\_bowls}
  \item \texttt{add\_remove\_lid} 
  \item \texttt{sort\_beads}
  \item \texttt{build\_unstack\_lego}
  \item \texttt{pick\_place\_food}
  \item \texttt{make\_sandwich}
  \item \texttt{setup\_cleanup\_table}
  \item \texttt{insert\_remove\_bookshelf}
  \item \texttt{insert\_remove\_cups\_from\_rack}
  \item \texttt{assemble\_disassemble\_legos}
  \item \texttt{play\_reset\_connect\_four}
  \item \texttt{screw\_unscrew\_fingers\_fixture}
\end{itemize}

For each selected task, we extract two keyframes: the \textit{initial frame} (before the manipulation begins) and the \textit{completion frame} (after the task concludes). These image pairs serve as the core data unit of our static visual reasoning benchmark. Additionally, we sample one or two \textit{intermediate frames} per video to support the construction of QA samples. These intermediate frames provide contextual cues for understanding transformation progress and are also used to generate semantically plausible distractors, thereby enhancing both the diversity and difficulty of the QA tasks.

\subsection{Image Cleaning and Filtering}

While directly extracting the initial and final frames from video sequences enables the construction of visual state contrast samples, it also introduces two primary challenges that negatively impact data quality and downstream reasoning performance:

\begin{itemize}
\item \textbf{Image blur and occlusion:} In many manipulation scenarios, key regions of interest are frequently occluded, often by the operator’s hands or arms, which obstruct target objects such as the manipulated item or the receiving container. In addition, camera motion or limitations in depth of field may result in image blur. These visual artifacts hinder comprehensive scene understanding and impair accurate recognition of object states.
\item \textbf{Insufficient manipulation complexity and low transformation saliency:} This issue is particularly prevalent in certain instances from tasks such as \texttt{pick\_place\_food} and \texttt{add\_remove\_lid}. For example, in the latter, the operator manipulates only one or two cups, merely placing or removing a lid. Similarly, the former involves relocating a single food item. Such cases entail minimal spatial reconfiguration and limited visual transitions, making them suboptimal for modeling multi-step reasoning or intermediate state understanding.
\end{itemize}

To mitigate these issues, we employ the o3~\cite{gpto3-o4-mini} model for automated image-pair assessment and filtering. Specifically, the model generates detailed scene descriptions and evaluates each sample’s adherence to the benchmark’s visual transformation criteria. These criteria include image clarity, visibility and distinctness of key objects, and the absence of severe occlusions. The model also estimates the number of actively manipulated objects in each scene to assess the complexity and relevance of the transformation. Samples that do not satisfy these thresholds—due to either quality degradation or limited interaction diversity—are excluded. As a result, the final benchmark retains approximately 85\% of high-quality samples.

\subsection{Automatic Metadata Annotation}

For each filtered sample, we automatically annotate the associated metadata. Tailored metadata types and corresponding prompt templates are designed for different manipulation scenarios. The full prompt contents appear in Sec.~D. The annotated metadata categories are described below.

\begin{table*}[ht]
\centering
\begin{tabular}{@{}ll@{}}
\toprule
\textbf{Task} & \textbf{Question Template} \\
\midrule
Intermediate State Recognition & \begin{tabular}[t]{@{}l@{}}
• \textit{Provide additional image3, image4, image5, and image6, which represents a} \\
\quad \textit{possible intermediate state during a manipulation task. Which one is a} \\
\quad \textit{reasonable intermediate state during the task?}
\end{tabular} \\
\midrule
Action Causality Reasoning & \begin{tabular}[t]{@{}l@{}}
• \textit{Identify which of the following operations is most likely to occur during the} \\
\quad \textit{transformation process.}
\end{tabular} \\
\midrule
Transformation Sequential Planning & \begin{tabular}[t]{@{}l@{}}
• \textit{What is the most likely operation in the \textbf{[first/last]} step to achieve this} \\
\quad \textit{transformation?} \\
• \textit{Which of the following objects should be manipulated \textbf{[first/last]} in the} \\
\quad \textit{transformation process?}
\end{tabular} \\
\midrule
Fine-Grained Change Recognition & \begin{tabular}[t]{@{}l@{}}
• \textit{After the transformation, where is the object originally at the \textbf{[absolute relation]}} \\
\quad \textit{position now located?} \\
• \textit{From the camera's viewpoint, after the transformation, which layer did the object} \\
\quad \textit{that was originally \textbf{[relative relation]} \textbf{[object2]} move to?} \\
• \textit{After the transformation, list all objects that are positioned \textbf{[above/below]} the} \\
\quad \textit{object that was originally \textbf{[absolute relation]}, whether or not they are in direct} \\
\quad \textit{contact.} \\
• \textit{From the camera's viewpoint, after the transformation, list all objects that are} \\
\quad \textit{positioned \textbf{[above/below]} the object that was originally \textbf{[relative relation]}} \\
\quad \textit{\textbf{[object2]}, whether or not they are in direct contact.}
\end{tabular} \\
\midrule
Global Change Recognition & \begin{tabular}[t]{@{}l@{}}
• \textit{Has the relative left-to-right position of the three books (left, middle, right)} \\
\quad \textit{changed?} \\
• \textit{Has the relative position (top-left, top-right, bottom-left, bottom-right) of the} \\
\quad \textit{four objects changed?} \\
• \textit{Has the clockwise/counterclockwise spatial order of the four objects changed?}
\end{tabular} \\
\midrule
Quantitative Transformation & \begin{tabular}[t]{@{}l@{}}
• \textit{How many food items are placed back into the basket?} \\
• \textit{How many food items are in the plate now that were not there before?} \\
• \textit{How many paper cups had their lids added or removed?} \\
• \textit{How many new groups consisting of beads with the same color have been} \\
\quad \textit{formed after the transformation?}
\end{tabular} \\
\bottomrule
\end{tabular}
\caption{Task Categories and Question Templates.}
\label{tab:task-templates}
\end{table*}

\begin{itemize}
    \item \textbf{Operation direction} indicates the directional nature of the manipulation, such as distinguishing between “put back” and “take out,” or “assemble” and “disassemble.” This is particularly useful for constructing reversible manipulation samples.
    
    \item \textbf{Background environment} captures the contextual setting of the image and is primarily used in intermediate state recognition tasks to ensure consistency between negative samples and the original scene background.
    
    \item \textbf{Object list} lists all manipulated objects, offering essential context for reasoning.
    
    \item \textbf{Initial absolute position} describes the spatial location of objects relative to the observer or the scene, using terms such as “leftmost” or “closest.”
    
    \item \textbf{Initial relative position} describes spatial relationships among objects, such as “the blue block is to the right of the red block.” These two types of position annotations serve as inputs for spatial reasoning tasks.
    
    \item \textbf{Final structure layout} captures the configuration of objects after task completion and supports a variety of tasks including spatial transformation, causal reasoning, and transformation sequence prediction.
    
    \item \textbf{Plate food state} records the conditions of food items at the beginning and end of a task, while \textbf{bead group count} indicates the number of bead clusters of the same color. Both support quantitative transformation.
    
    \item \textbf{Board state} encodes the positions and colors of game pieces in a board-based task and is used to support reasoning about implicit manipulations.
\end{itemize}

To improve the accuracy of structural annotations, we use Gemini 2.5 Pro~\cite{gemini25pro} to automatically label all metadata categories. For the final structure layout field, direct extraction from full-frame images is often unreliable due to background clutter and visual noise. To address this, we first apply Grounding DINO~\cite{GroundingDino} to detect task-relevant objects (such as block stacks or bowl sets) and extract their bounding boxes. These object regions are then cropped and provided as auxiliary views to Gemini 2.5 Pro, enabling more precise structural descriptions.

In particular, for the \texttt{build\_unstack\_lego} scenario, we observe that the model’s structural parsing remains unreliable. To ensure annotation quality, we manually annotate all metadata for this scene.

\subsection{QA Generation and Verification}

In the QA generation stage, we select representative scenes for each fine-grained subtask and design corresponding question templates. Several templates include masked placeholders (e.g., [MASK]), which are automatically populated using structured metadata to produce complete question texts. Table~\ref{tab:task-templates} illustrates examples of templates aligned with different VisualTrans subtasks.

For the majority of QA instances, the ground-truth answer is directly retrieved from the associated metadata. Distractor options are generated accordingly to construct multiple-choice formats. In a small number of cases where metadata coverage is incomplete, we manually annotate the correct answers to ensure reliability.

To assess the quality of automatically generated QA samples, we randomly select 500 pairs for downstream evaluation. Given that automatic annotation introduces an estimated labeling error rate of approximately 10 percent, and that certain scenes may exhibit ambiguous spatial configurations or action sequences, all sampled QA pairs undergo rigorous human verification.

To facilitate efficient and consistent review, we develop a custom web-based annotation interface. Annotators are instructed to answer each question based solely on the visual inputs, without access to any metadata or ground-truth annotations. They also flag problematic questions exhibiting issues such as missing correct options, incorrect ground truths, semantic ambiguity, or scenes with insufficient visual evidence to determine the answer. Identified QA issues are corrected or discarded, and the corresponding metadata is updated to maintain alignment with the verified content.

By iteratively refining the question templates and QA synthesis procedures, we substantially enhance the quality and clarity of the resulting QA pairs. Ultimately, we curate a set of 472 high-quality QA samples. These samples are distributed across three primary task categories: \textbf{procedural transformation} (45.8\%), \textbf{spatial transformation} (40.9\%), and \textbf{quantitative transformation} (13.3\%). More specifically, \textit{intermediate state recognition} (18.6\%), \textit{action causality reasoning} (18.2\%), and \textit{transformation sequence planning} (8.9\%) are grouped under procedural transformation, while \textit{fine-grained change recognition} (35.6\%) and \textit{global change recognition} (5.3\%) fall under spatial transformation.

\section{C Details of Evaluation}

This section outlines the evaluation protocol, including model inference configurations and the computational resources used in all experiments conducted on the VisualTrans benchmark.

\subsection{C.1 Computational Resources}

All data processing, benchmark construction, and evaluation experiments use a workstation equipped with Intel Xeon E5-2698 v4 CPUs and four NVIDIA Tesla V100-SXM2 GPUs, each with 32GB of memory.

\subsection{C.2 General Evaluation Setup}

For evaluation on the VisualTrans benchmark, we perform inference using the official APIs provided by each model. Default parameter configurations are adopted, including decoding strategy, temperature, top-p, and top-k settings. To ensure consistency and fairness across all models, we apply a unified prompt format (shown in Fig.\ref{fig:visual-trans-prompt21} and Fig.\ref{fig:visual-trans-prompt22}), followed by task-specific questions corresponding to each reasoning type.

\section{D Additional Visualizations}

This section presents the prompt templates used for data cleaning and metadata annotation, along with selected case studies to illustrate benchmark performance.

\subsection{D.1 Cleaning and Filtering Prompts}

We provide the complete set of prompts used for image-pair filtering with the o3 model~\cite{gpto3-o4-mini}. These prompts, illustrated in Fig.\ref{fig:BOOKEND_PROMPT} through Fig.\ref{fig:OTHER_PROMPT}, support automated filtering based on image clarity, object visibility, and transformation complexity.

\subsection{D.2 Metadata Annotation Prompts}

This subsection presents the full collection of prompts designed for automatic metadata annotation using the Gemini 2.5 model. All prompts are shown in Fig.\ref{fig:visual-trans-prompt1} through Fig.\ref{fig:visual-trans-prompt20}.

\subsection{D.3 Case Studies}

We include case studies for representative sub-tasks in the benchmark. Each case provides two questions and corresponding model responses. All examples are illustrated in Fig.\ref{fig:Visualization1} through Fig.\ref{fig:Visualization12}.

\section{E Future Work}

VisualTrans lays the groundwork for evaluating VTR in real-world human-object interaction settings, yet several directions remain open for future development.

\paragraph{Expanding Task Diversity.}
The current benchmark focuses on table-top manipulation scenarios. We plan to extend VisualTrans to include a broader range of everyday activities, such as tool use, deformable object manipulation~\cite{zhang2025vtla,hao2025tla,li2024foundation}, and outdoor tasks~\cite{worldsimbench,zhang2025novel,HAO2025103018}. This expansion will enable more comprehensive evaluation of visual reasoning under varied spatial and temporal dynamics.

\paragraph{Integrating Multi-Agent Interactions.}
All current samples involve single-agent behavior. In future iterations, we aim to incorporate multi-agent scenarios, which introduce additional complexity in terms of intention modeling, joint planning~\cite{tan2025roboos,zhao2025training}, and social affordances~\cite{tang2025affordgrasp}.

We release VisualTrans and its extensible pipeline to facilitate community-driven improvements and encourage collaborative expansion toward a more generalizable evaluation of visual reasoning models.

\renewcommand{\thefigure}{\arabic{figure}a}

\begin{figure*}[htbp]
    \centering
    \includegraphics[width=1.0\linewidth]{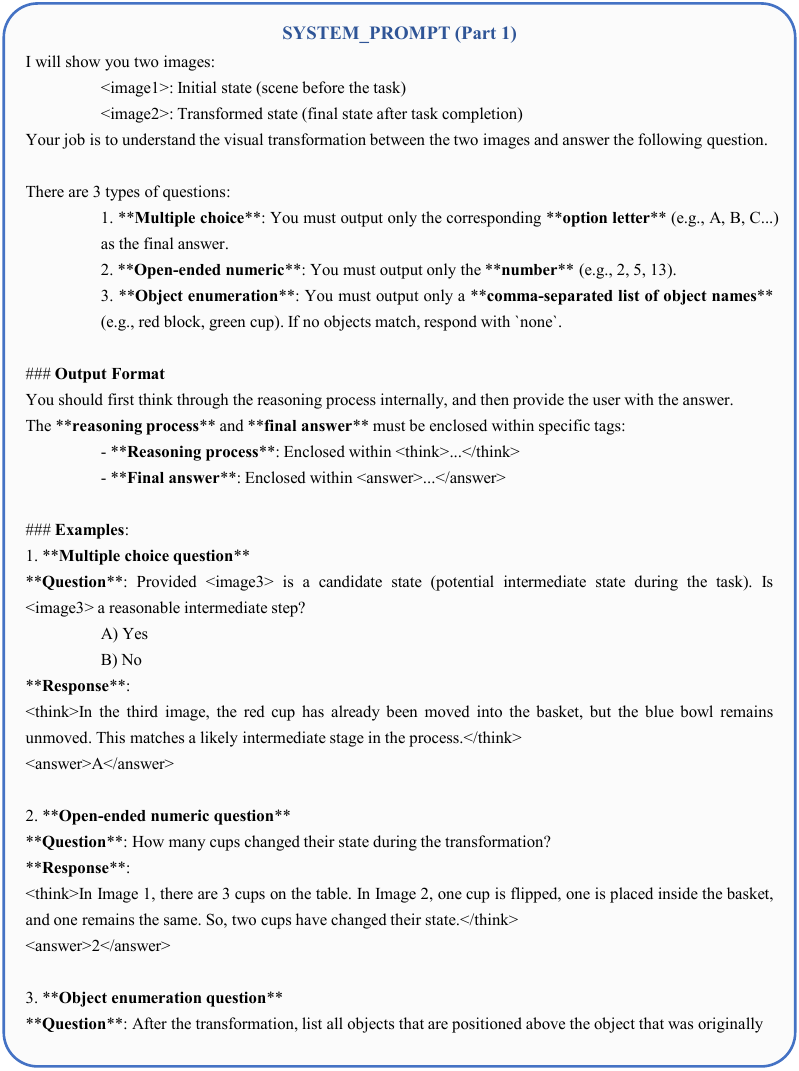}
    \caption{SYSTEM PROMPT (Part 1)}
    \label{fig:visual-trans-prompt21}
\end{figure*}

\renewcommand{\thefigure}{\arabic{figure}b}
\addtocounter{figure}{-1} 

\begin{figure*}[htbp]
    \centering
    \includegraphics[width=1.0\linewidth]{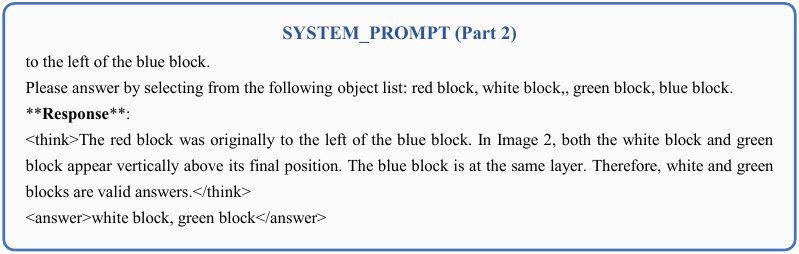}
    \caption{SYSTEM PROMPT (Part 2)}
    \label{fig:visual-trans-prompt22}
\end{figure*}

\renewcommand{\thefigure}{\arabic{figure}}

\begin{figure*}[htbp]
    \centering
    \includegraphics[width=1.0\linewidth]{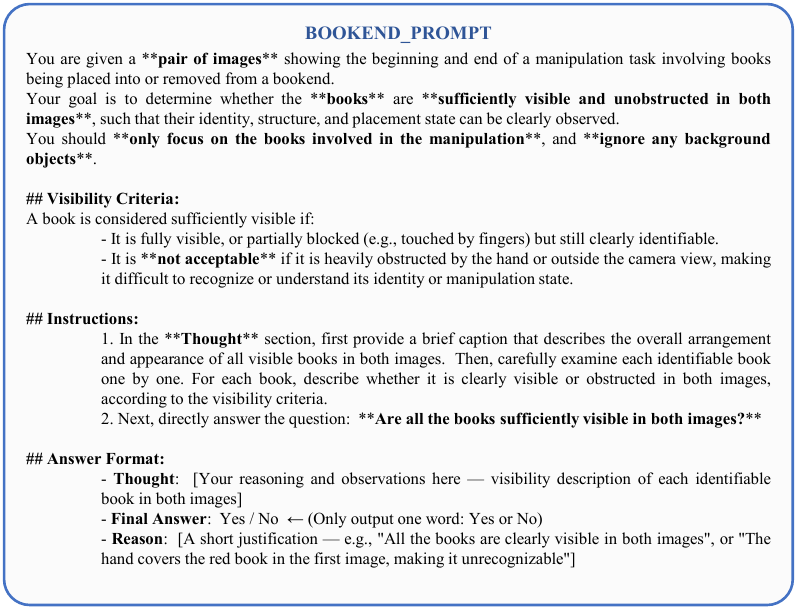}
    \caption{BOOKEND PROMPT}
    \label{fig:BOOKEND_PROMPT}
\end{figure*}

\begin{figure*}[htbp]
    \centering
    \includegraphics[width=1.0\linewidth]{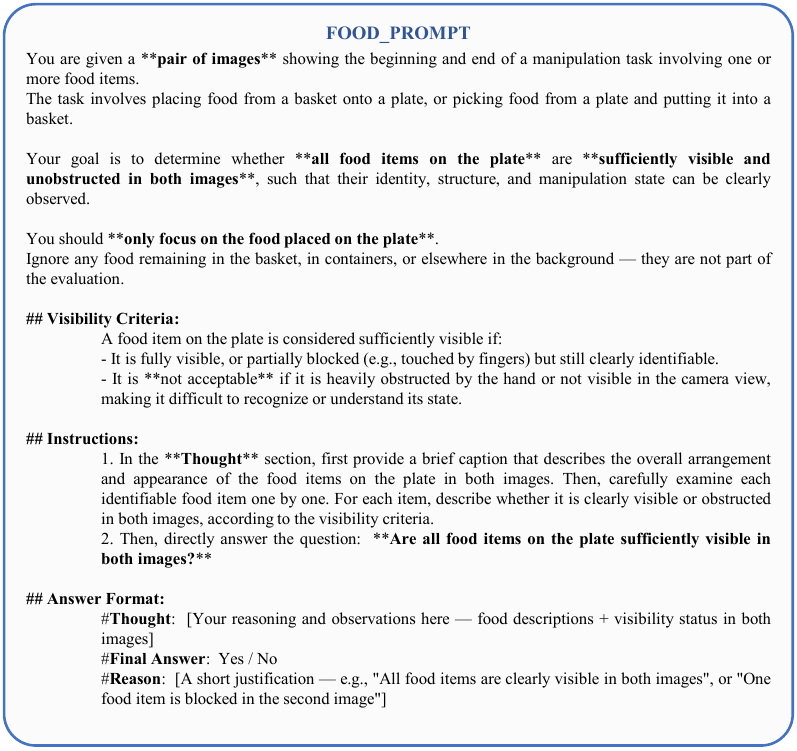}
    \caption{FOOD PROMPT}
    \label{fig:FOOD_PROMPT}
\end{figure*}

\begin{figure*}[htbp]
    \centering
    \includegraphics[width=1.0\linewidth]{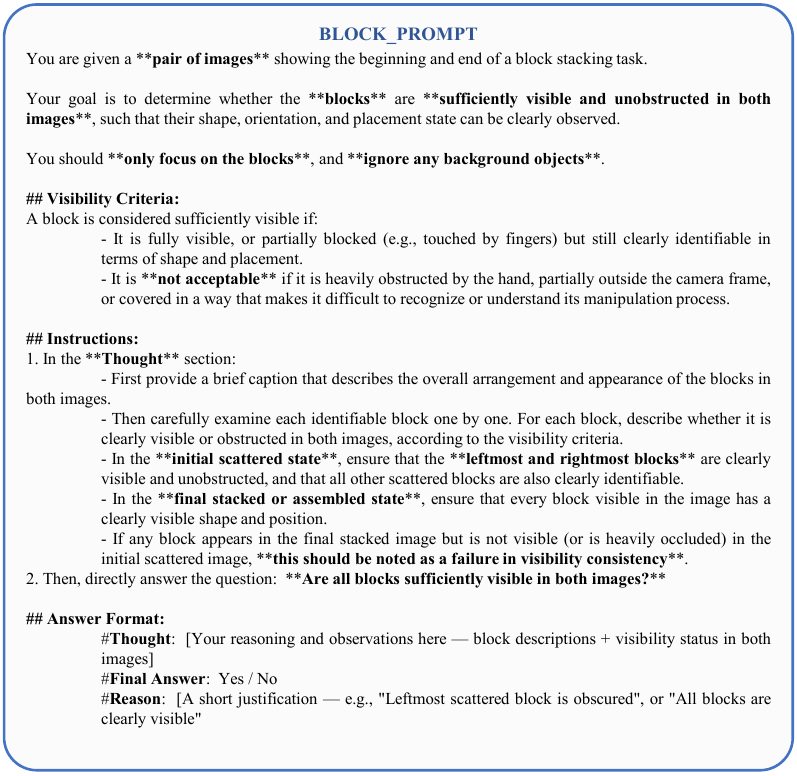}
    \caption{BLOCK PROMPT}
    \label{fig:BLOCK_PROMPT}
\end{figure*}

\begin{figure*}[htbp]
    \centering
    \includegraphics[width=1.0\linewidth]{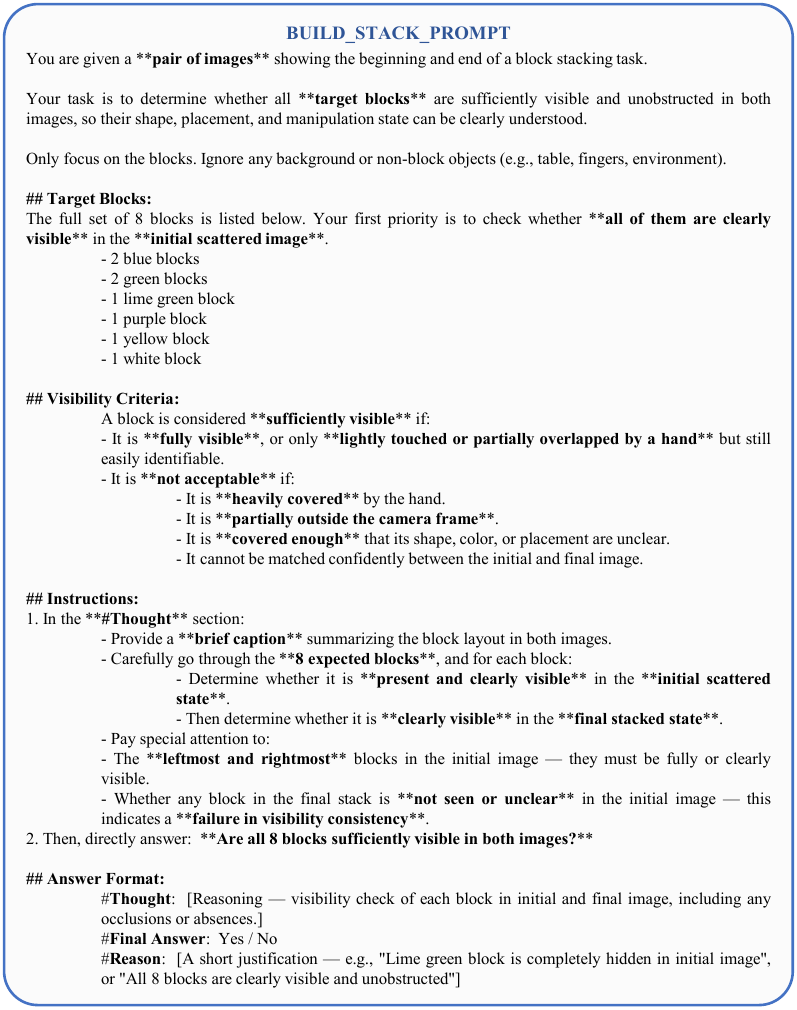}
    \caption{BUILD STACK PROMPT}
    \label{fig:BUILD_STACK_PROMPT}
\end{figure*}

\begin{figure*}[htbp]
    \centering
    \includegraphics[width=1.0\linewidth]{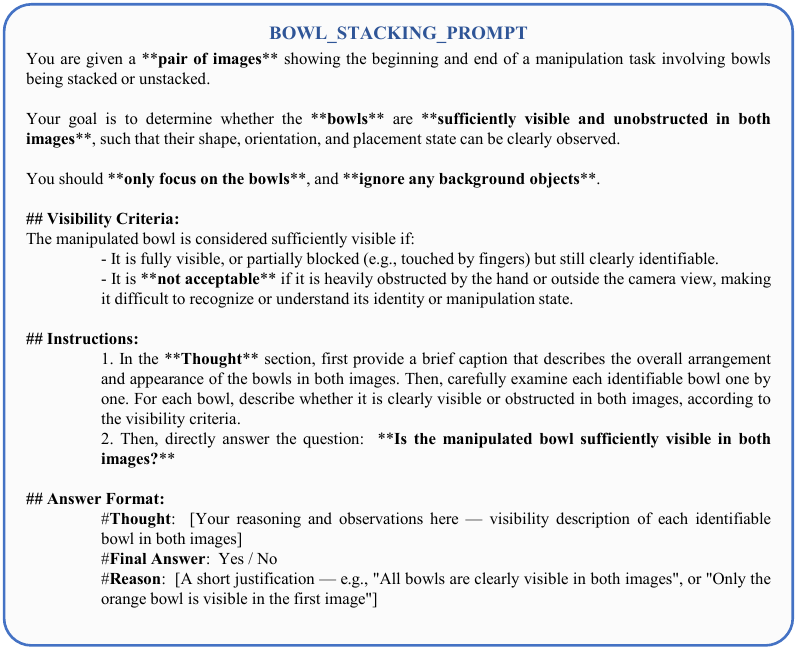}
    \caption{BOWL STACKING PROMPT}
    \label{fig:BOWL_STACKING_PROMPT}
\end{figure*}

\begin{figure*}[htbp]
    \centering
    \includegraphics[width=1.0\linewidth]{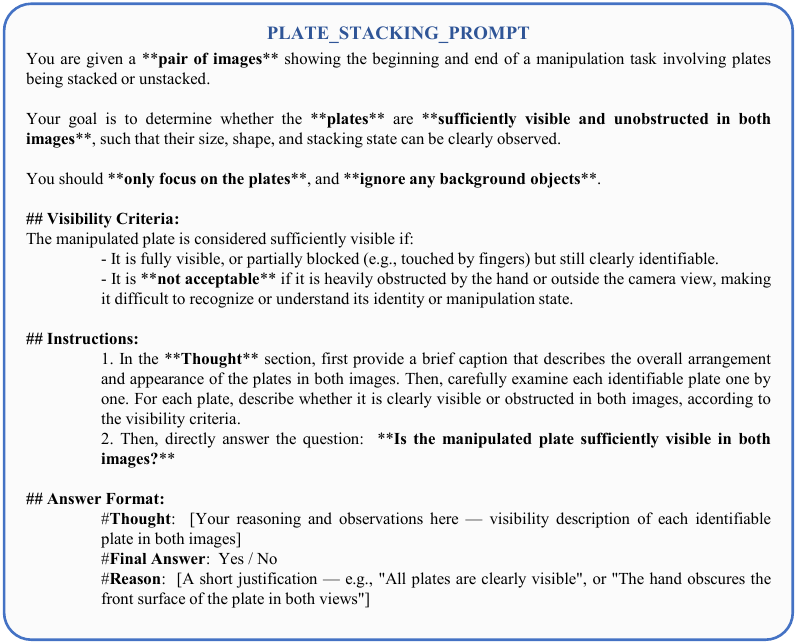}
    \caption{PLATE STACKING PROMPT}
    \label{fig:PLATE_STACKING_PROMPT}
\end{figure*}

\begin{figure*}[htbp]
    \centering
    \includegraphics[width=1.0\linewidth]{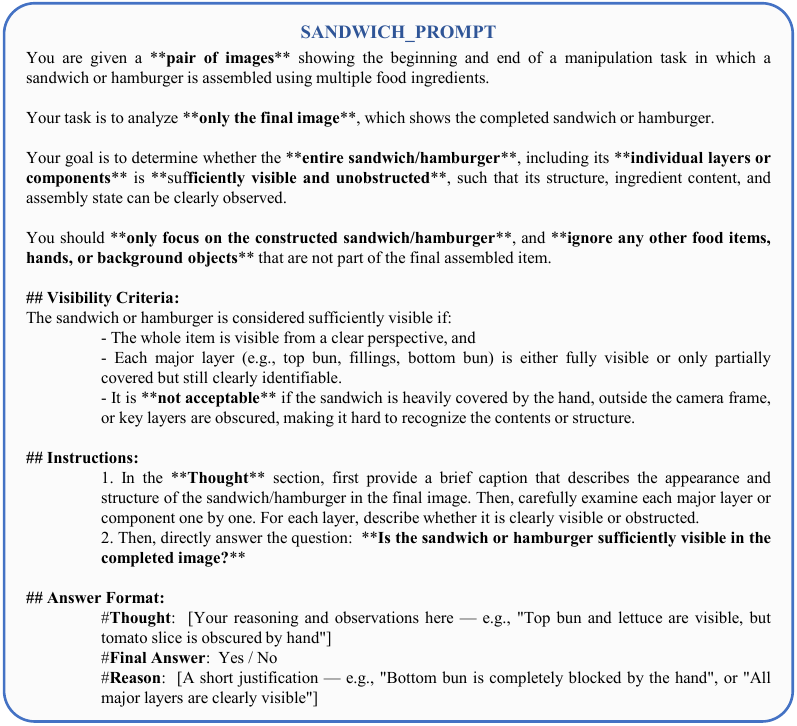}
    \caption{SANDWICH PROMPT}
    \label{fig:SANDWICH_PROMPT}
\end{figure*}

\begin{figure*}[htbp]
    \centering
    \includegraphics[width=1.0\linewidth]{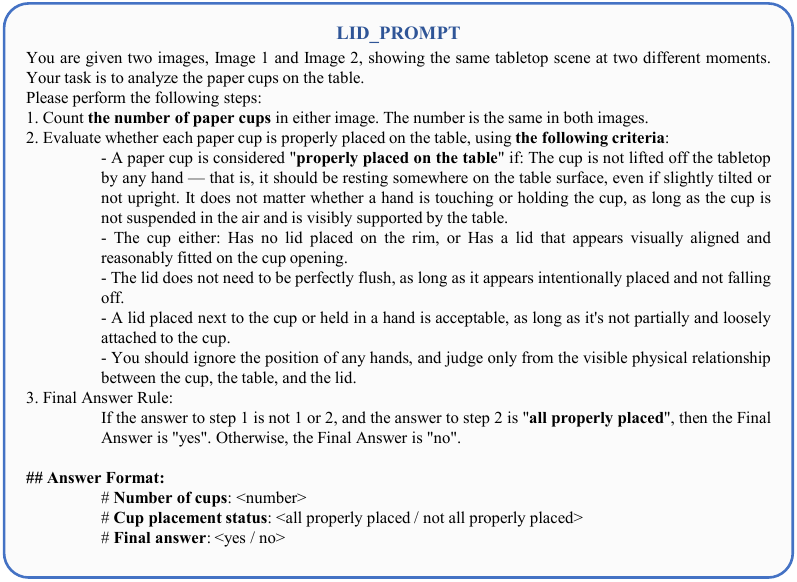}
    \caption{LID PROMPT}
    \label{fig:LID_PROMPT}
\end{figure*}

\begin{figure*}[htbp]
    \centering
    \includegraphics[width=1.0\linewidth]{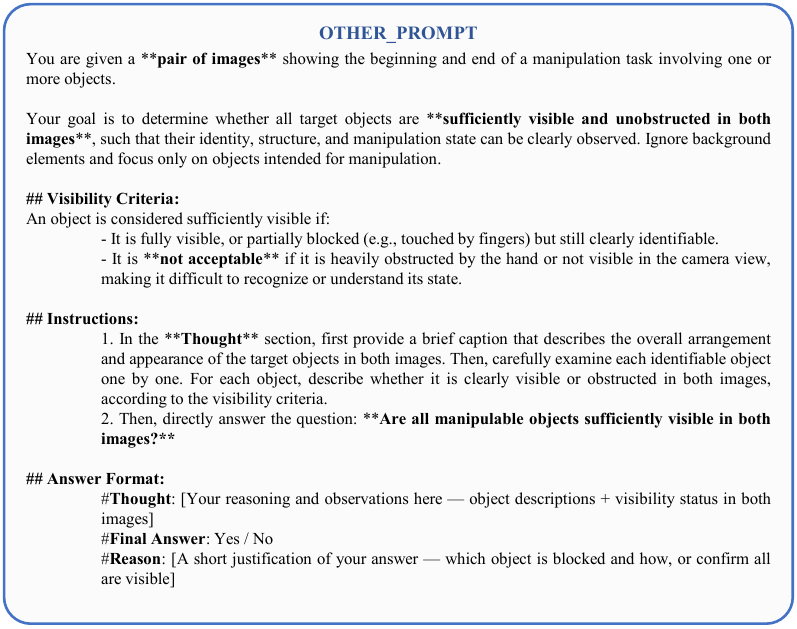}
    \caption{OTHER PROMPT}
    \label{fig:OTHER_PROMPT}
\end{figure*}

\renewcommand{\thefigure}{\arabic{figure}a}

\begin{figure*}[htbp]
    \centering
    \includegraphics[width=1.0\linewidth]{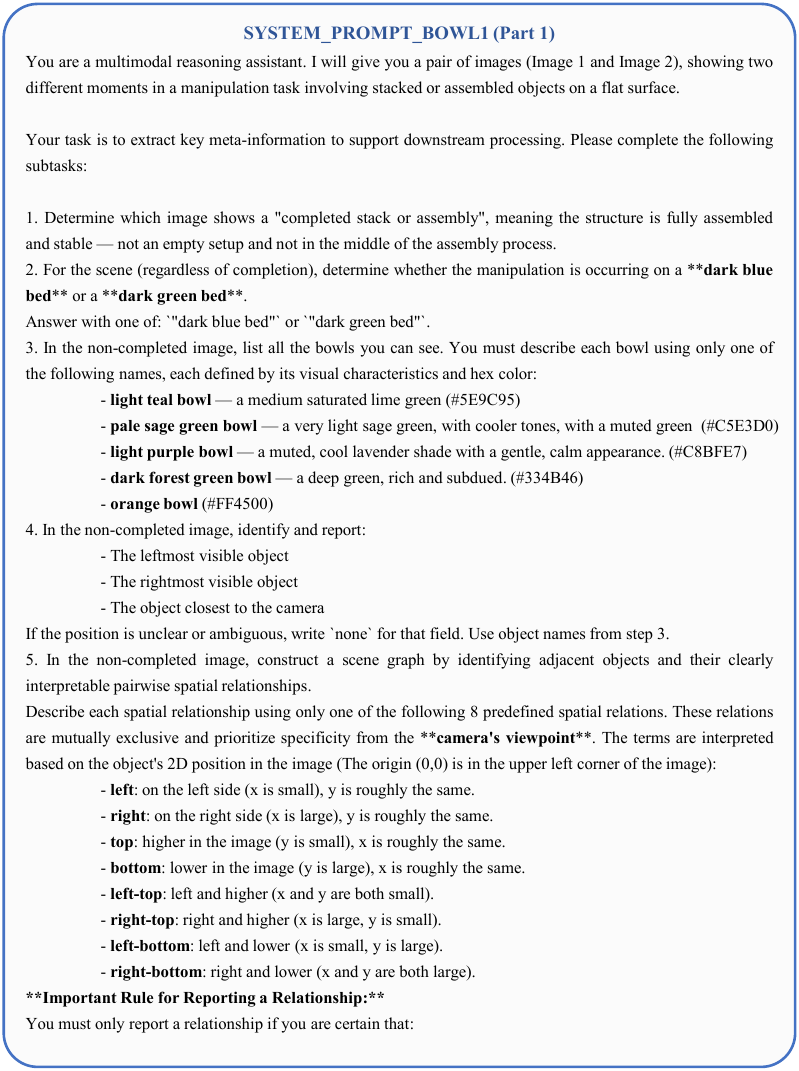}
    \caption{SYSTEM PROMPT BOWL1 (Part 1)}
    \label{fig:visual-trans-prompt1}
\end{figure*}

\renewcommand{\thefigure}{\arabic{figure}b}
\addtocounter{figure}{-1} 

\begin{figure*}[htbp]
    \centering
    \includegraphics[width=1.0\linewidth]{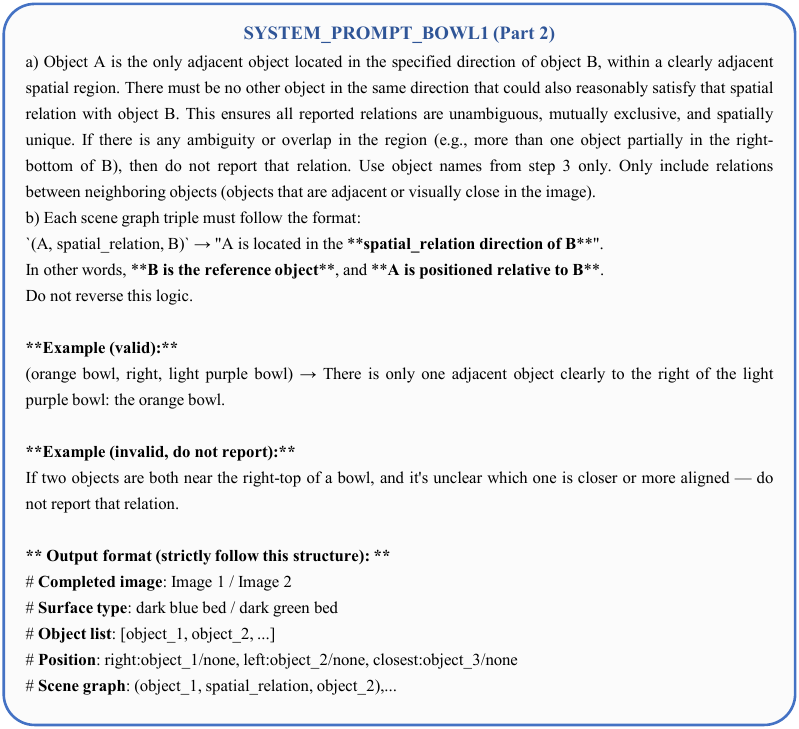}
    \caption{SYSTEM PROMPT BOWL1 (Part 2)}
    \label{fig:visual-trans-prompt2}
\end{figure*}

\renewcommand{\thefigure}{\arabic{figure}}

\begin{figure*}[htbp]
    \centering
    \includegraphics[width=1.0\linewidth]{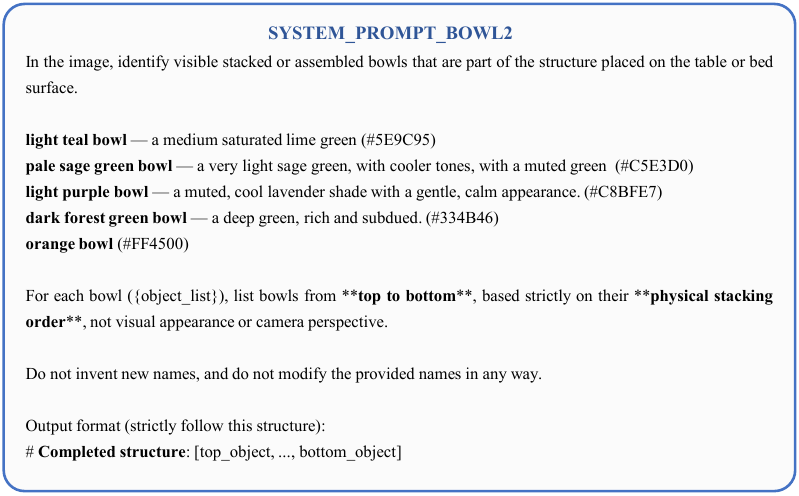}
    \caption{SYSTEM PROMPT BOWL2}
    \label{fig:visual-trans-prompt3}
\end{figure*}

\renewcommand{\thefigure}{\arabic{figure}a}

\begin{figure*}[htbp]
    \centering
    \includegraphics[width=1.0\linewidth]{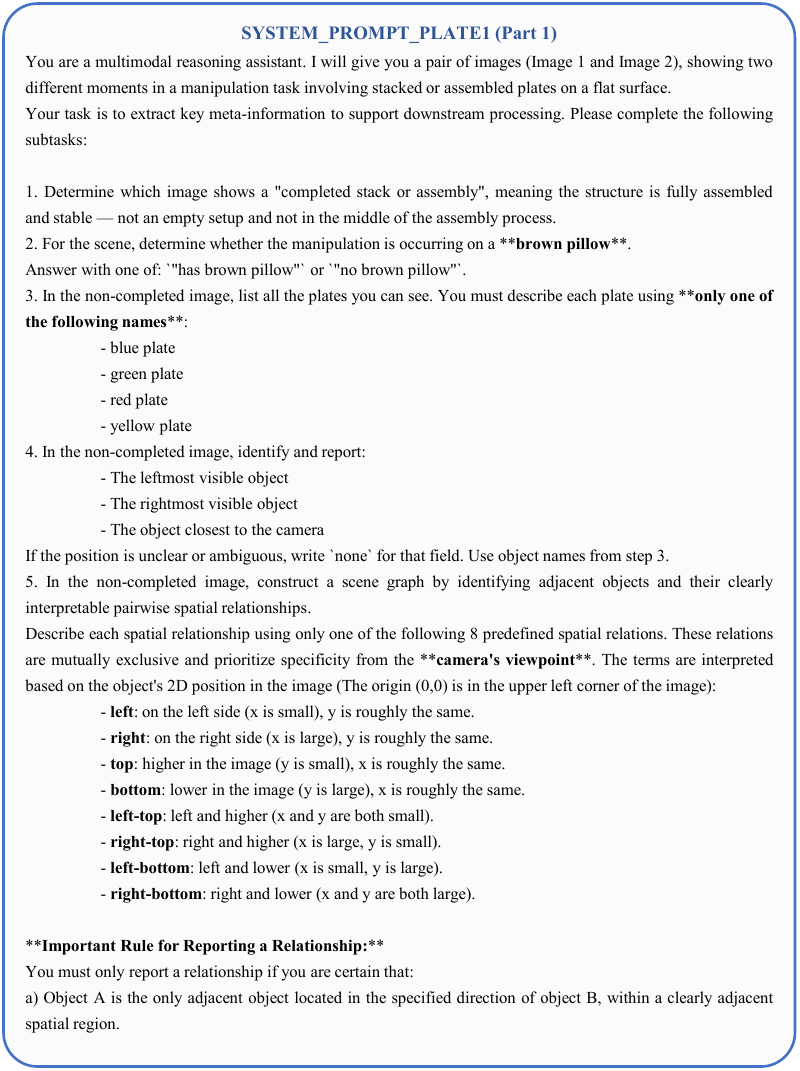}
    \caption{SYSTEM PROMPT PLATE1 (Part 1)}
    \label{fig:visual-trans-prompt4}
\end{figure*}

\renewcommand{\thefigure}{\arabic{figure}b}
\addtocounter{figure}{-1} 

\begin{figure*}[htbp]
    \centering
    \includegraphics[width=1.0\linewidth]{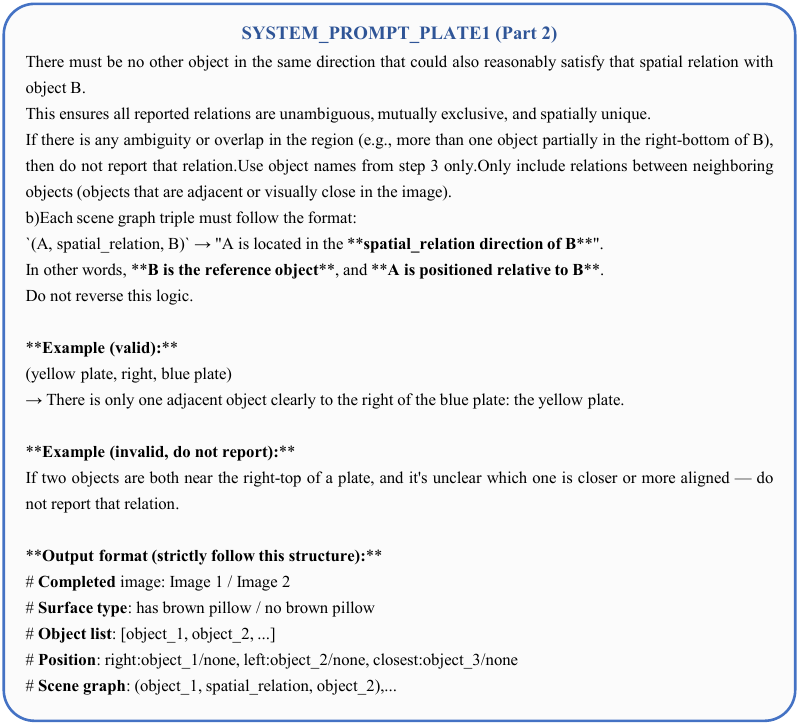}
    \caption{SYSTEM PROMPT PLATE1 (Part 2)}
    \label{fig:visual-trans-prompt5}
\end{figure*}

\renewcommand{\thefigure}{\arabic{figure}}

\begin{figure*}[htbp]
    \centering
    \includegraphics[width=1.0\linewidth]{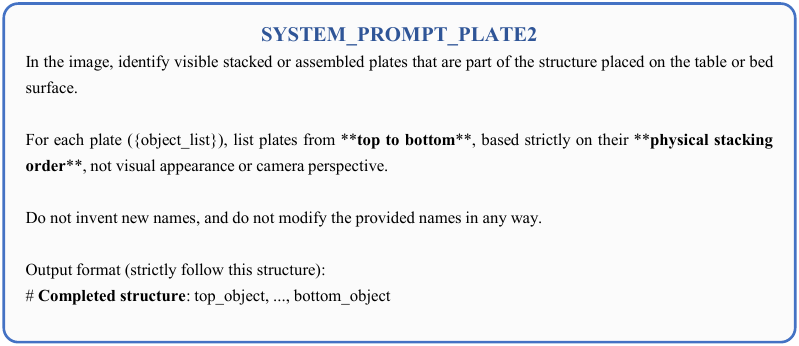}
    \caption{SYSTEM PROMPT PLATE2}
    \label{fig:visual-trans-prompt6}
\end{figure*}

\renewcommand{\thefigure}{\arabic{figure}a}

\begin{figure*}[htbp]
    \centering
    \includegraphics[width=1.0\linewidth]{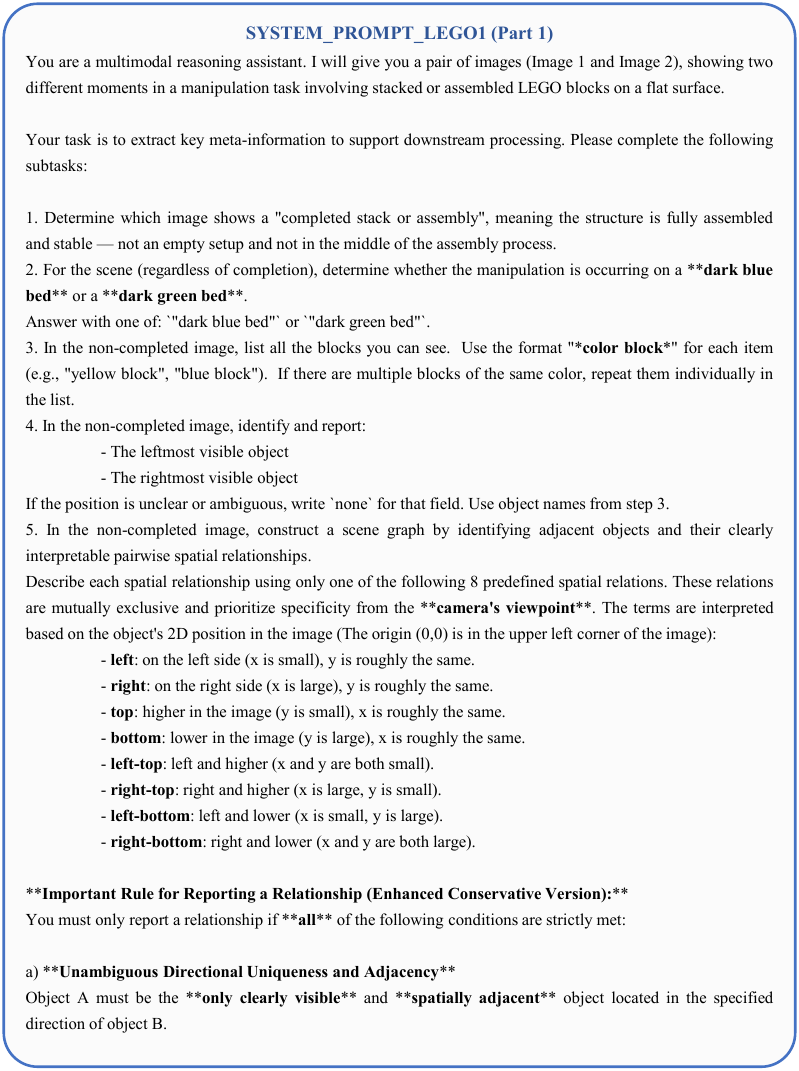}
    \caption{SYSTEM PROMPT LEGO1 (Part 1)}
    \label{fig:visual-trans-prompt7}
\end{figure*}

\renewcommand{\thefigure}{\arabic{figure}b}
\addtocounter{figure}{-1} 

\begin{figure*}[htbp]
    \centering
    \includegraphics[width=1.0\linewidth]{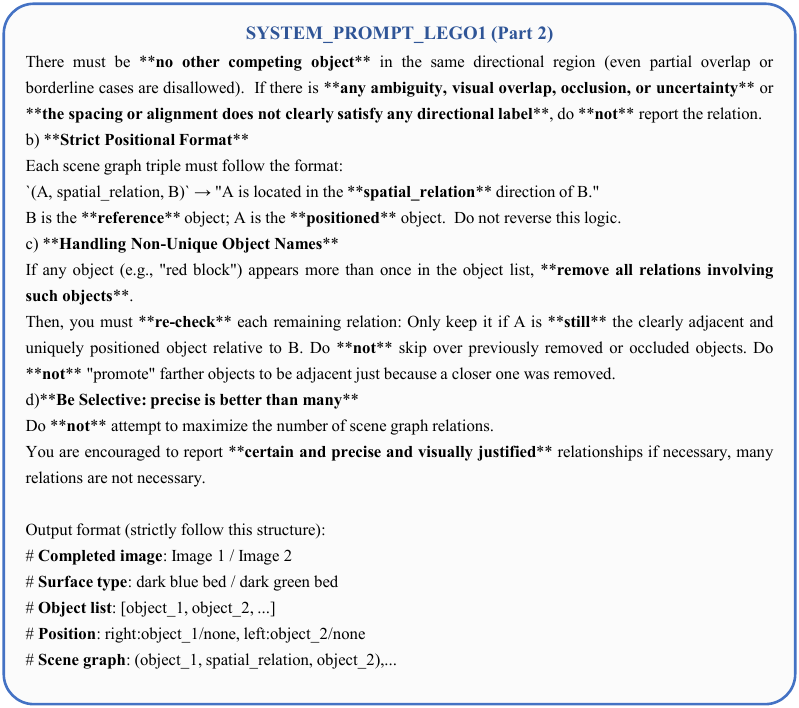}
    \caption{SYSTEM PROMPT LEGO1 (Part 2)}
    \label{fig:visual-trans-prompt8}
\end{figure*}

\renewcommand{\thefigure}{\arabic{figure}}

\begin{figure*}[htbp]
    \centering
    \includegraphics[width=1.0\linewidth]{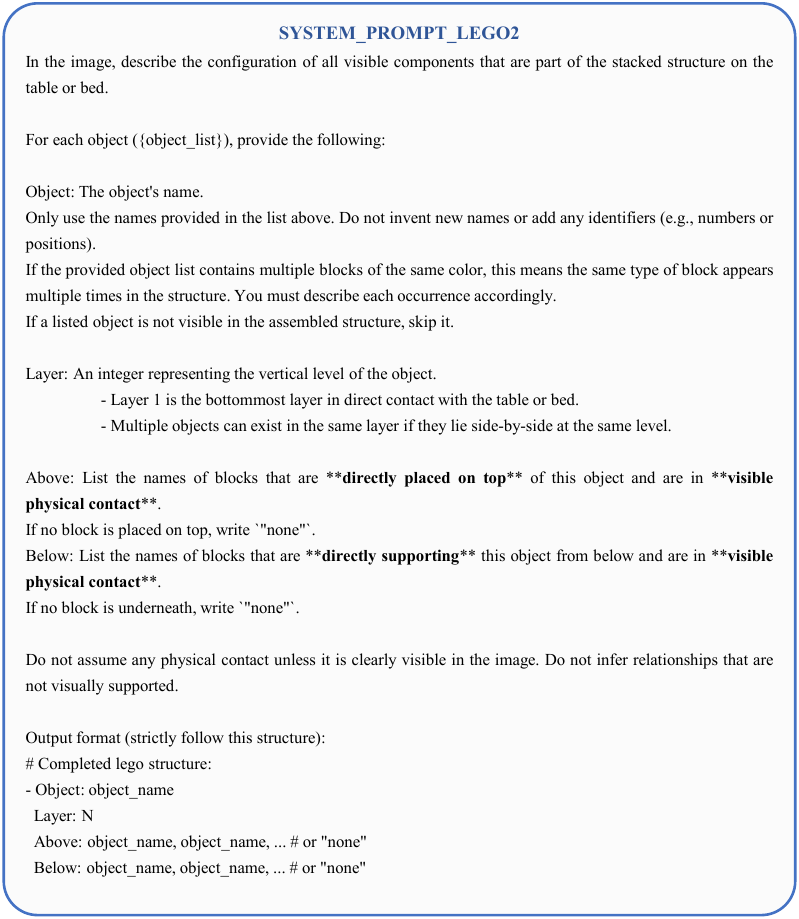}
    \caption{SYSTEM PROMPT LEGO2}
    \label{fig:visual-trans-prompt9}
\end{figure*}

\renewcommand{\thefigure}{\arabic{figure}a}

\begin{figure*}[htbp]
    \centering
    \includegraphics[width=1.0\linewidth]{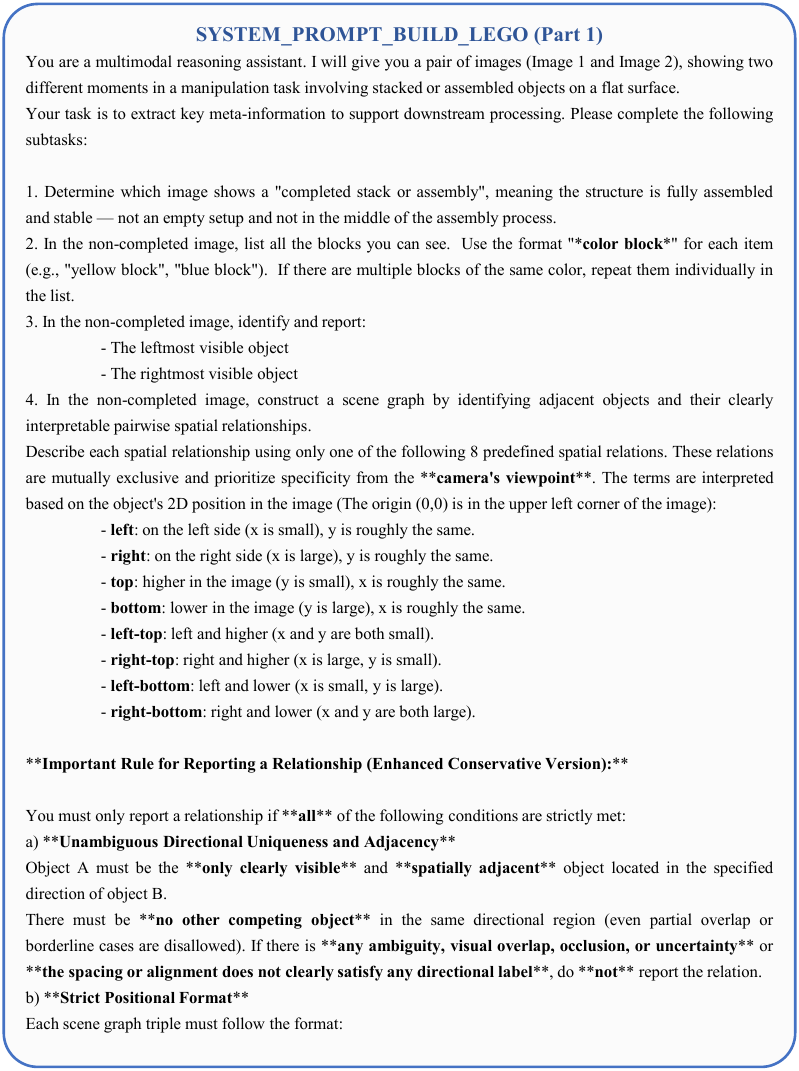}
    \caption{SYSTEM PROMPT BUILD LEGO (Part 1)}
    \label{fig:visual-trans-prompt10}
\end{figure*}

\renewcommand{\thefigure}{\arabic{figure}b}
\addtocounter{figure}{-1} 

\begin{figure*}[htbp]
    \centering
    \includegraphics[width=1.0\linewidth]{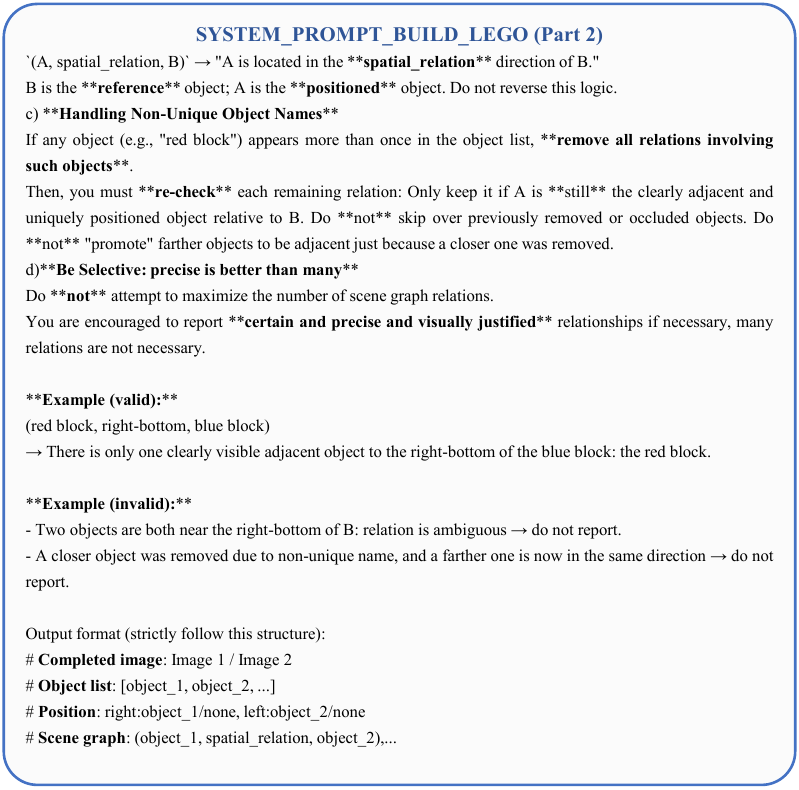}
    \caption{SYSTEM PROMPT BUILD LEGO (Part 2)}
    \label{fig:visual-trans-prompt11}
\end{figure*}

\renewcommand{\thefigure}{\arabic{figure}}
\renewcommand{\thefigure}{\arabic{figure}a}

\begin{figure*}[htbp]
    \centering
    \includegraphics[width=1.0\linewidth]{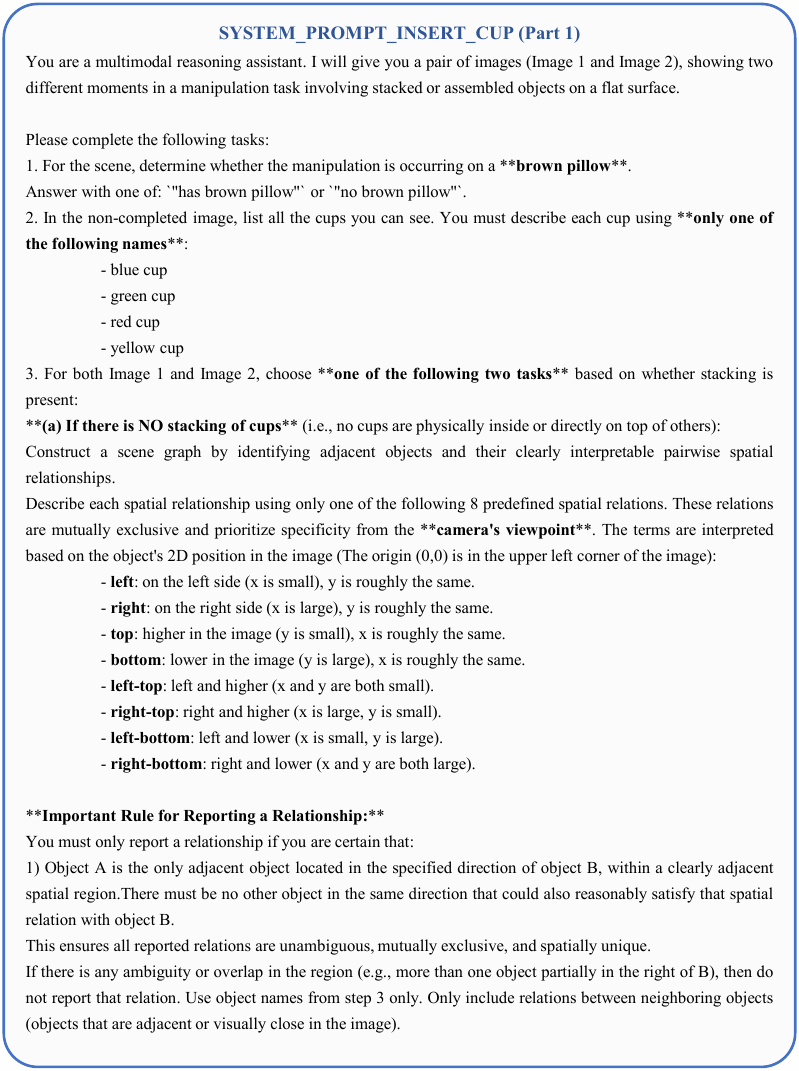}
    \caption{SYSTEM PROMPT INSERT CUP (Part 1)}
    \label{fig:visual-trans-prompt12}
\end{figure*}

\renewcommand{\thefigure}{\arabic{figure}b}
\addtocounter{figure}{-1} 

\begin{figure*}[htbp]
    \centering
    \includegraphics[width=1.0\linewidth]{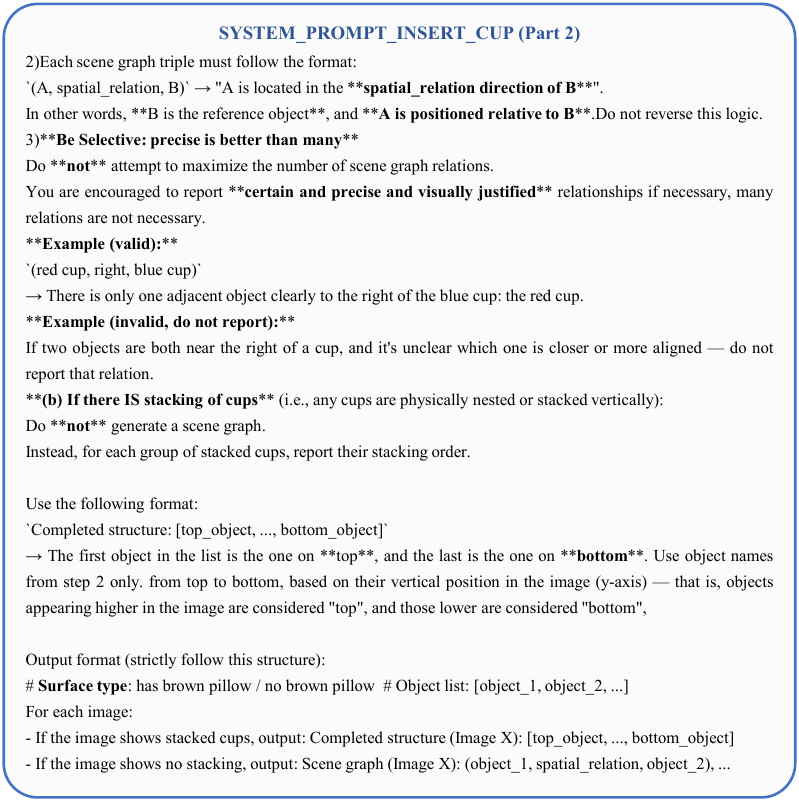}
    \caption{SYSTEM PROMPT INSERT CUP (Part 2)}
    \label{fig:visual-trans-prompt13}
\end{figure*}

\renewcommand{\thefigure}{\arabic{figure}}
\renewcommand{\thefigure}{\arabic{figure}a}

\begin{figure*}[htbp]
    \centering
    \includegraphics[width=1.0\linewidth]{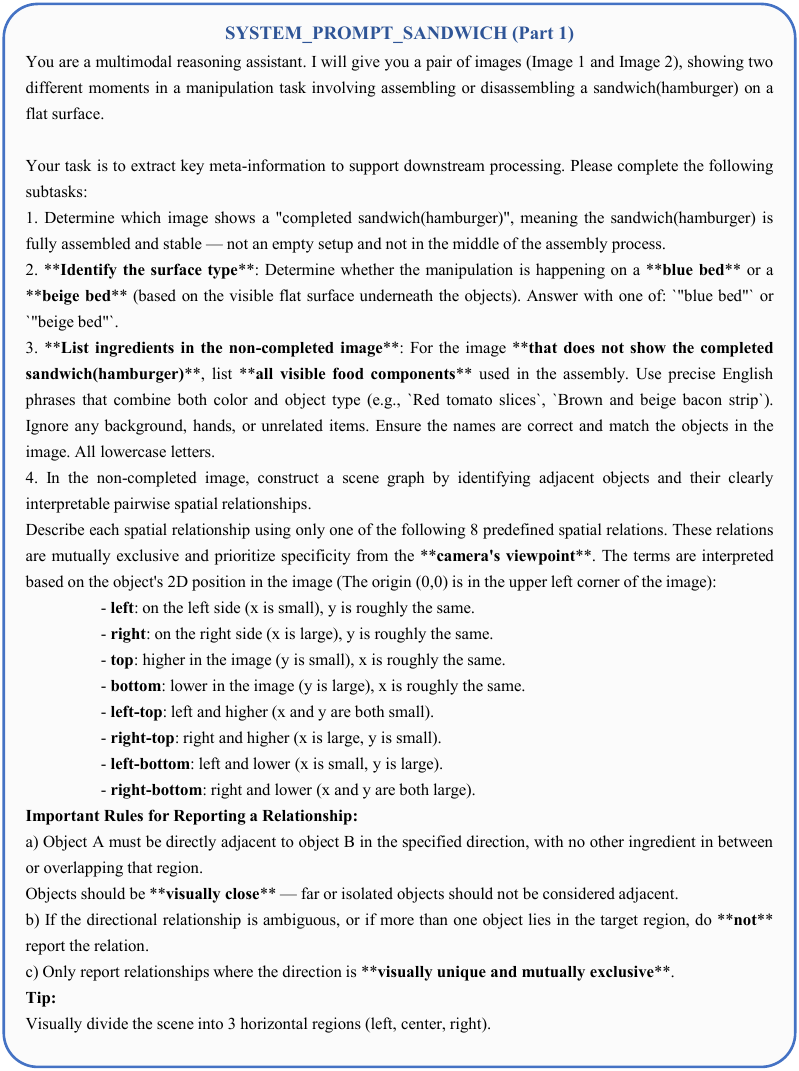}
    \caption{SYSTEM PROMPT SANDWICH (Part 1)}
    \label{fig:visual-trans-prompt14}
\end{figure*}

\renewcommand{\thefigure}{\arabic{figure}b}
\addtocounter{figure}{-1} 

\begin{figure*}[htbp]
    \centering
    \includegraphics[width=1.0\linewidth]{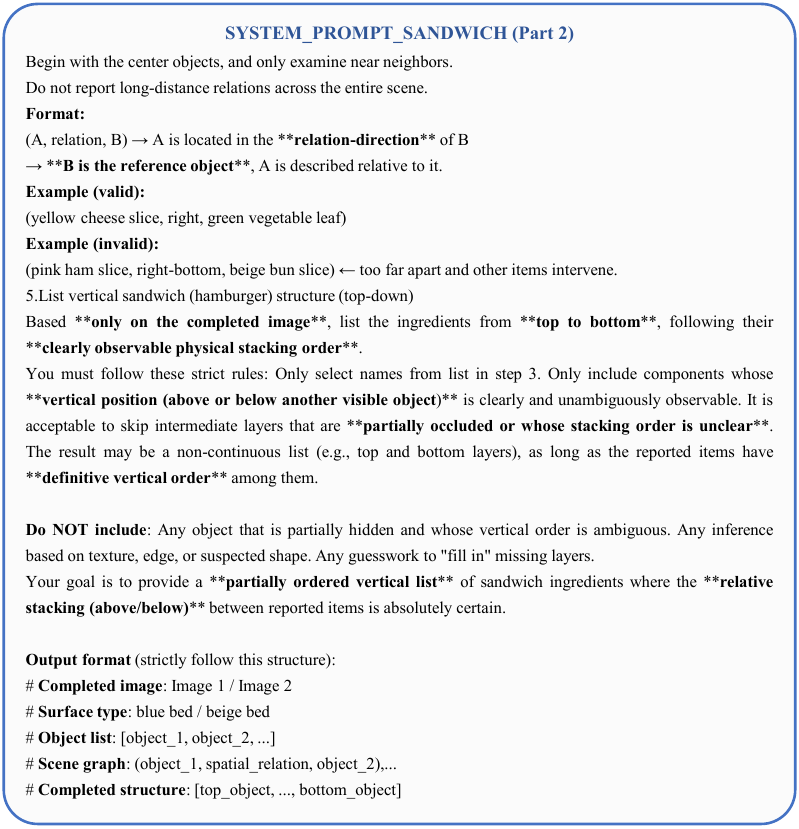}
    \caption{SYSTEM PROMPT SANDWICH (Part 2)}
    \label{fig:visual-trans-prompt15}
\end{figure*}

\renewcommand{\thefigure}{\arabic{figure}}

\begin{figure*}[htbp]
    \centering
    \includegraphics[width=1.0\linewidth]{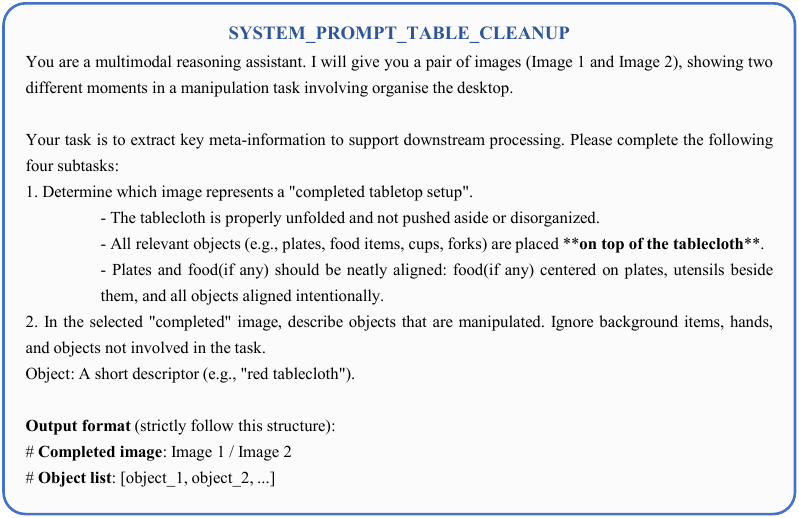}
    \caption{SYSTEM PROMPT TABLE CLEANUP}
    \label{fig:visual-trans-prompt16}
\end{figure*}

\begin{figure*}[htbp]
    \centering
    \includegraphics[width=1.0\linewidth]{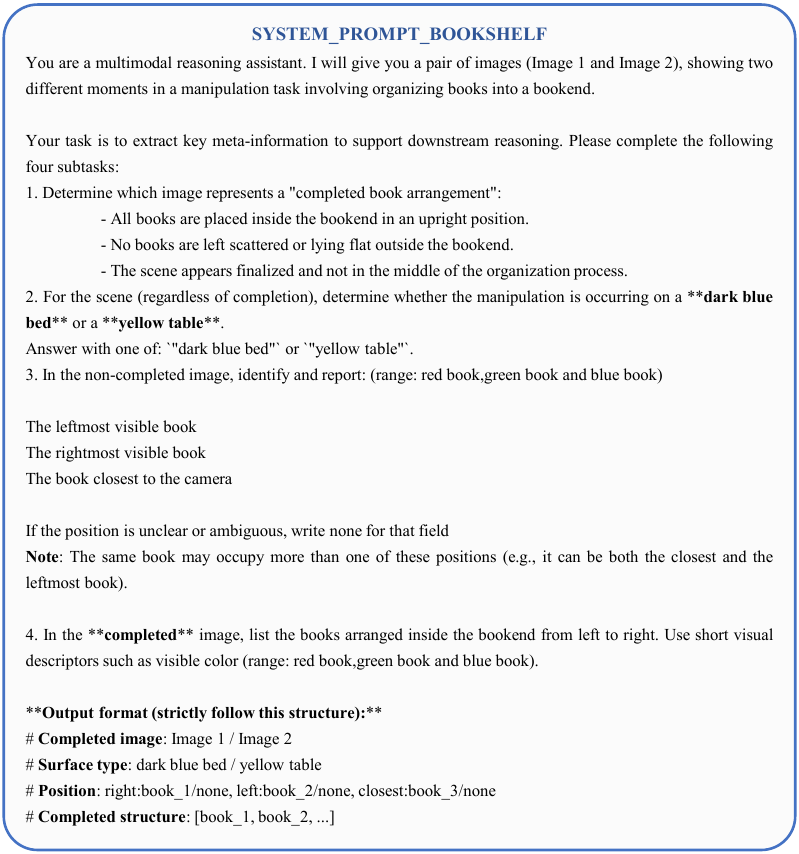}
    \caption{SYSTEM PROMPT BOOKSHELF}
    \label{fig:visual-trans-prompt17}
\end{figure*}

\begin{figure*}[htbp]
    \centering
    \includegraphics[width=1.0\linewidth]{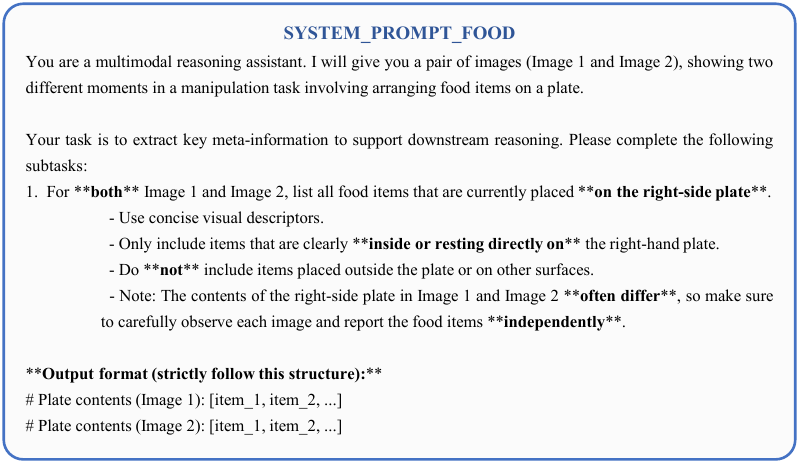}
    \caption{SYSTEM PROMPT FOOD}
    \label{fig:visual-trans-prompt18}
\end{figure*}

\begin{figure*}[htbp]
    \centering
    \includegraphics[width=1.0\linewidth]{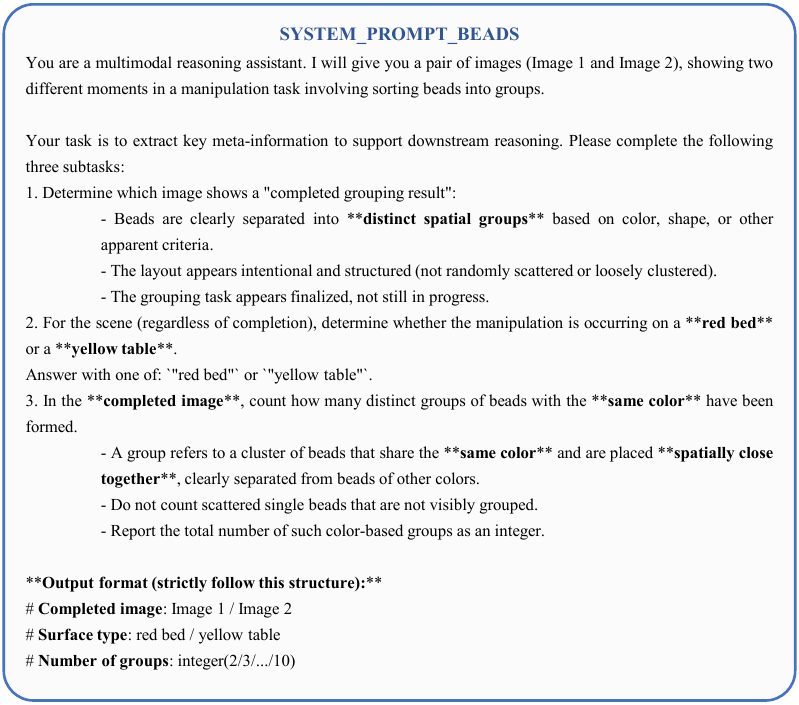}
    \caption{SYSTEM PROMPT BEADS}
    \label{fig:visual-trans-prompt19}
\end{figure*}

\begin{figure*}[htbp]
    \centering
    \includegraphics[width=1.0\linewidth]{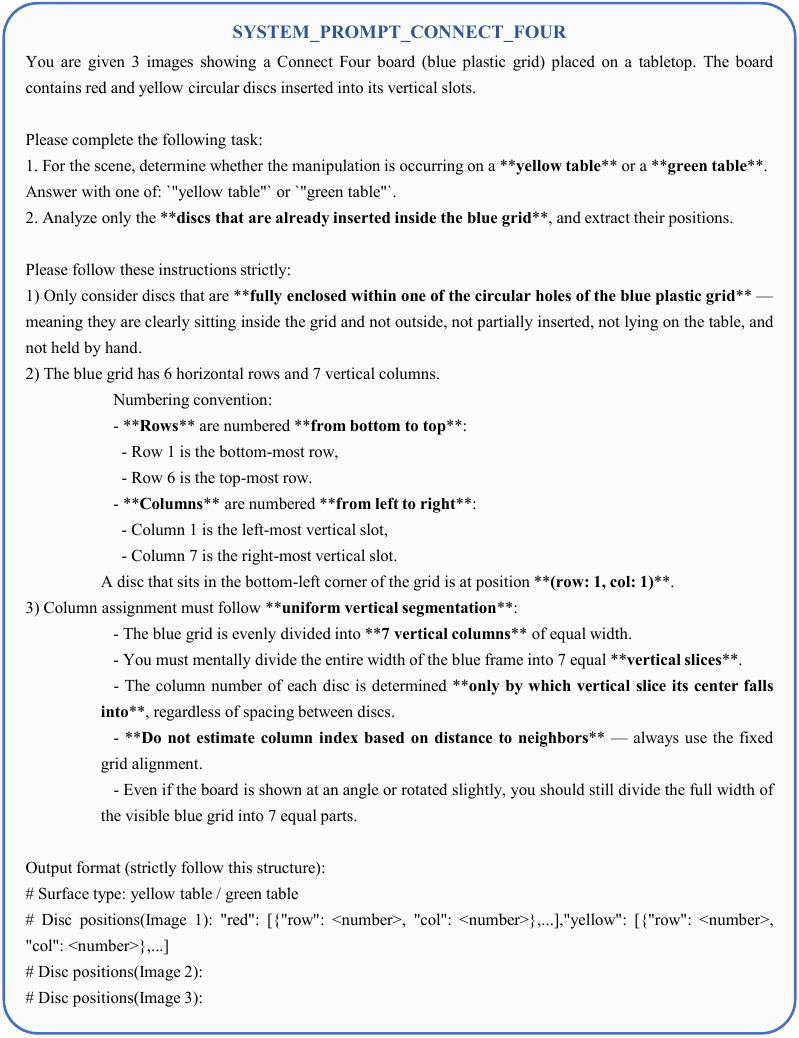}
    \caption{SYSTEM PROMPT CONNECT FOUR}
    \label{fig:visual-trans-prompt20}
\end{figure*}

\begin{figure*}[htbp]
    \centering
    \includegraphics[width=1.0\linewidth]{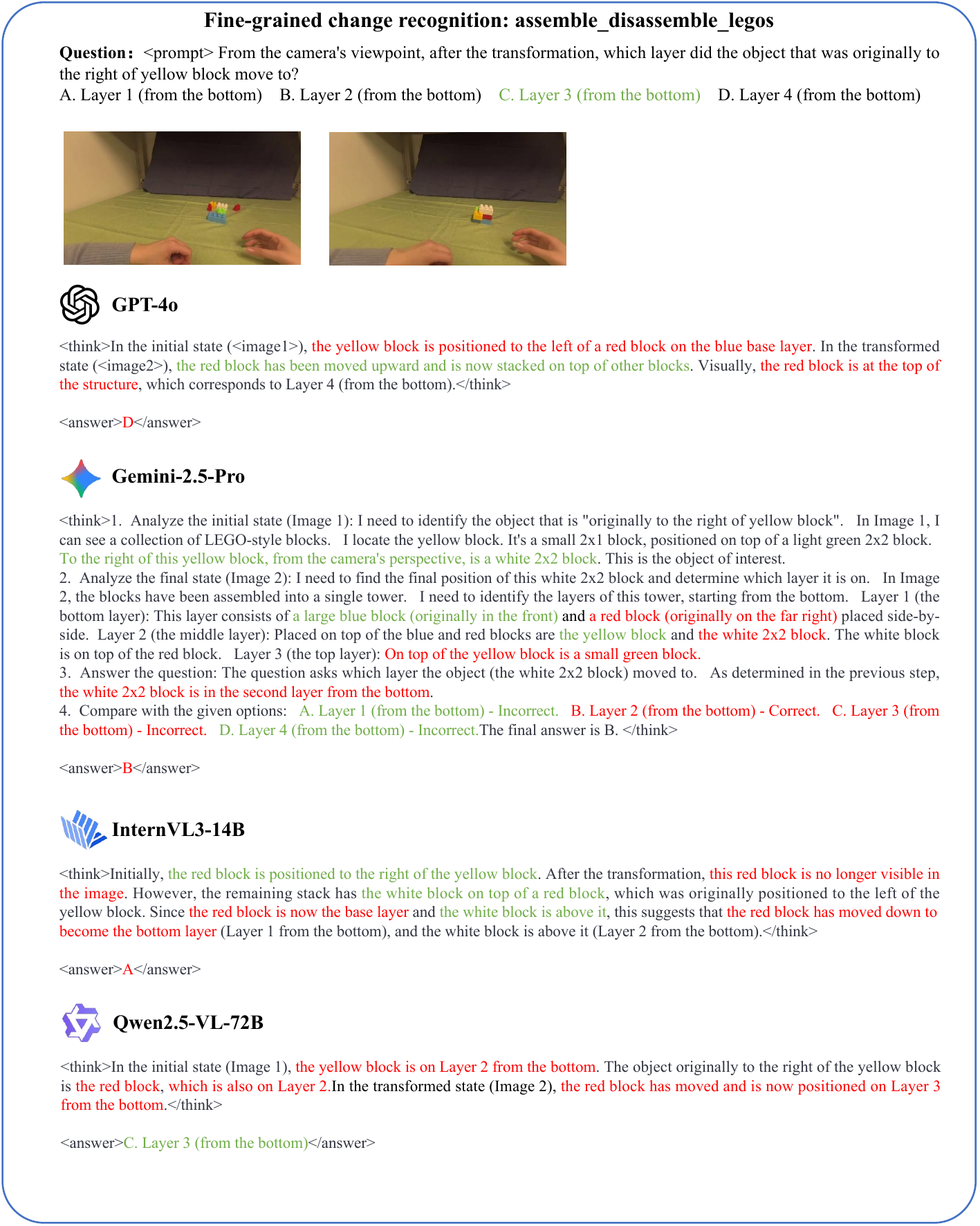}
    \caption{Fine-grained change recognition: assemble\_disassemble\_legos (Sample 156)}
    \label{fig:Visualization1}
\end{figure*}

\begin{figure*}[htbp]
    \centering
    \includegraphics[width=1.0\linewidth]{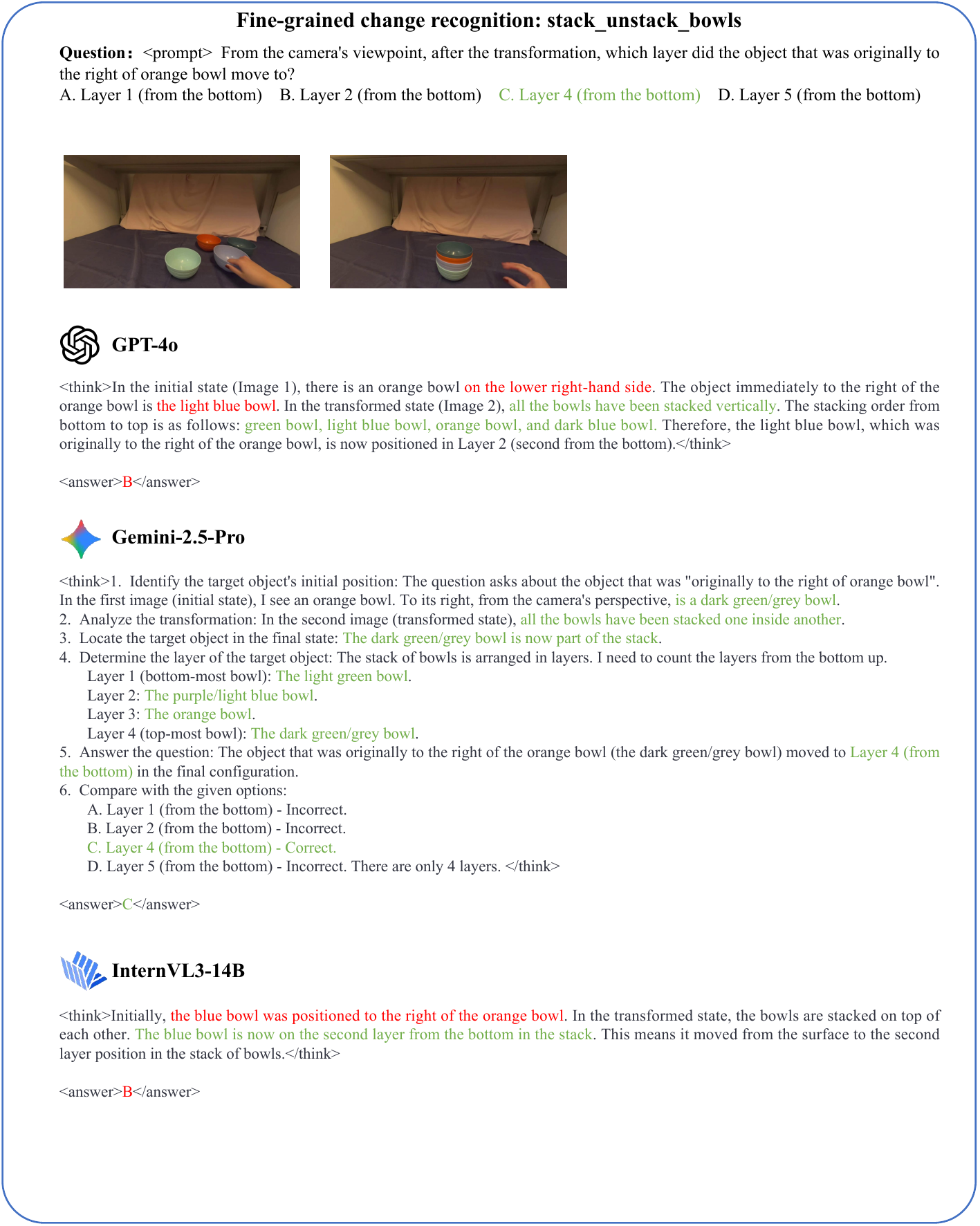}
    \caption{Fine-grained change recognition: stack\_unstack\_bowls (Sample 188)}
    \label{fig:Visualization2}
\end{figure*}

\begin{figure*}[htbp]
    \centering
    \includegraphics[width=1.0\linewidth]{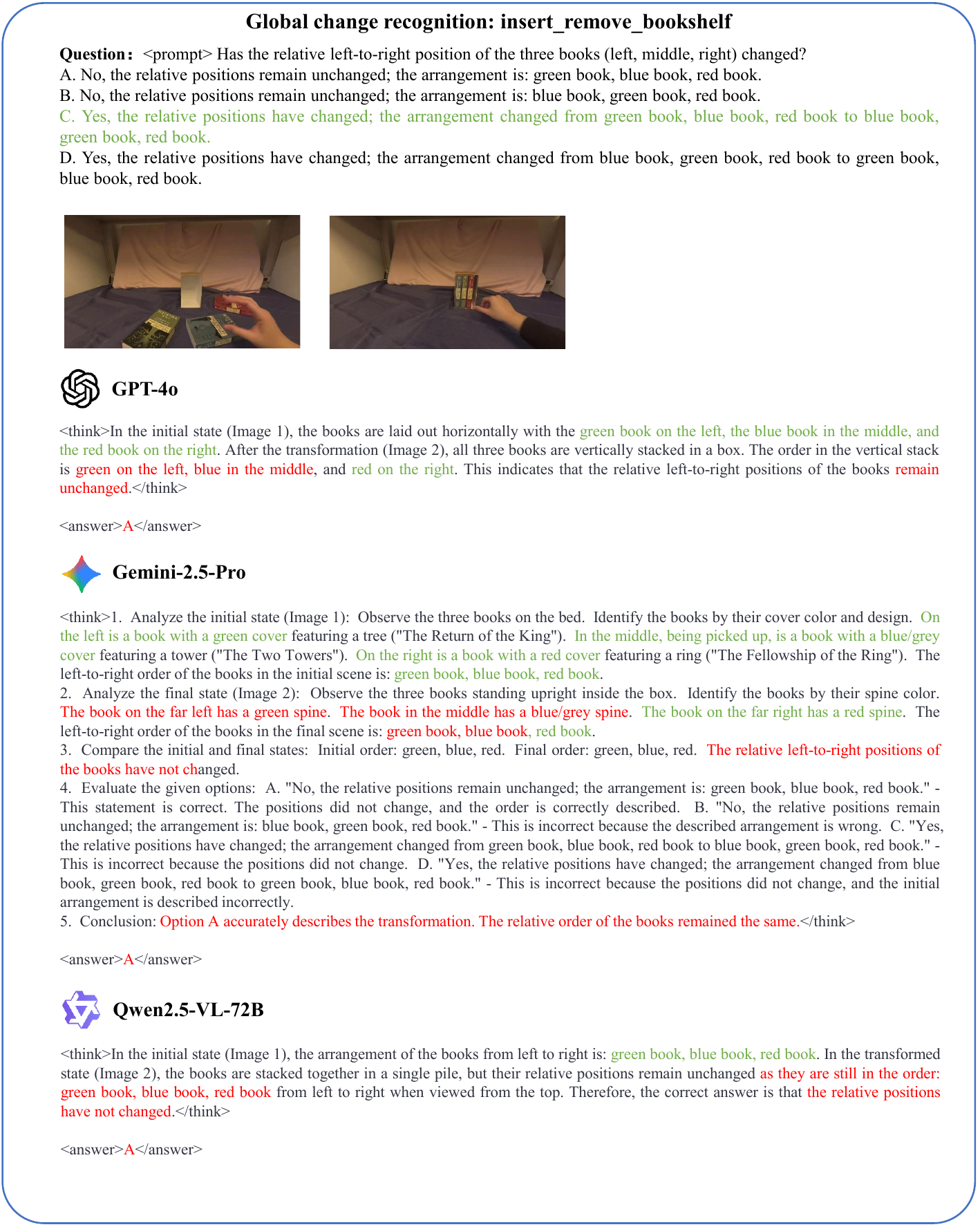}
    \caption{Global change recognition: insert\_remove\_bookshelf (Sample 458)}
    \label{fig:Visualization3}
\end{figure*}

\begin{figure*}[htbp]
    \centering
    \includegraphics[width=1.0\linewidth]{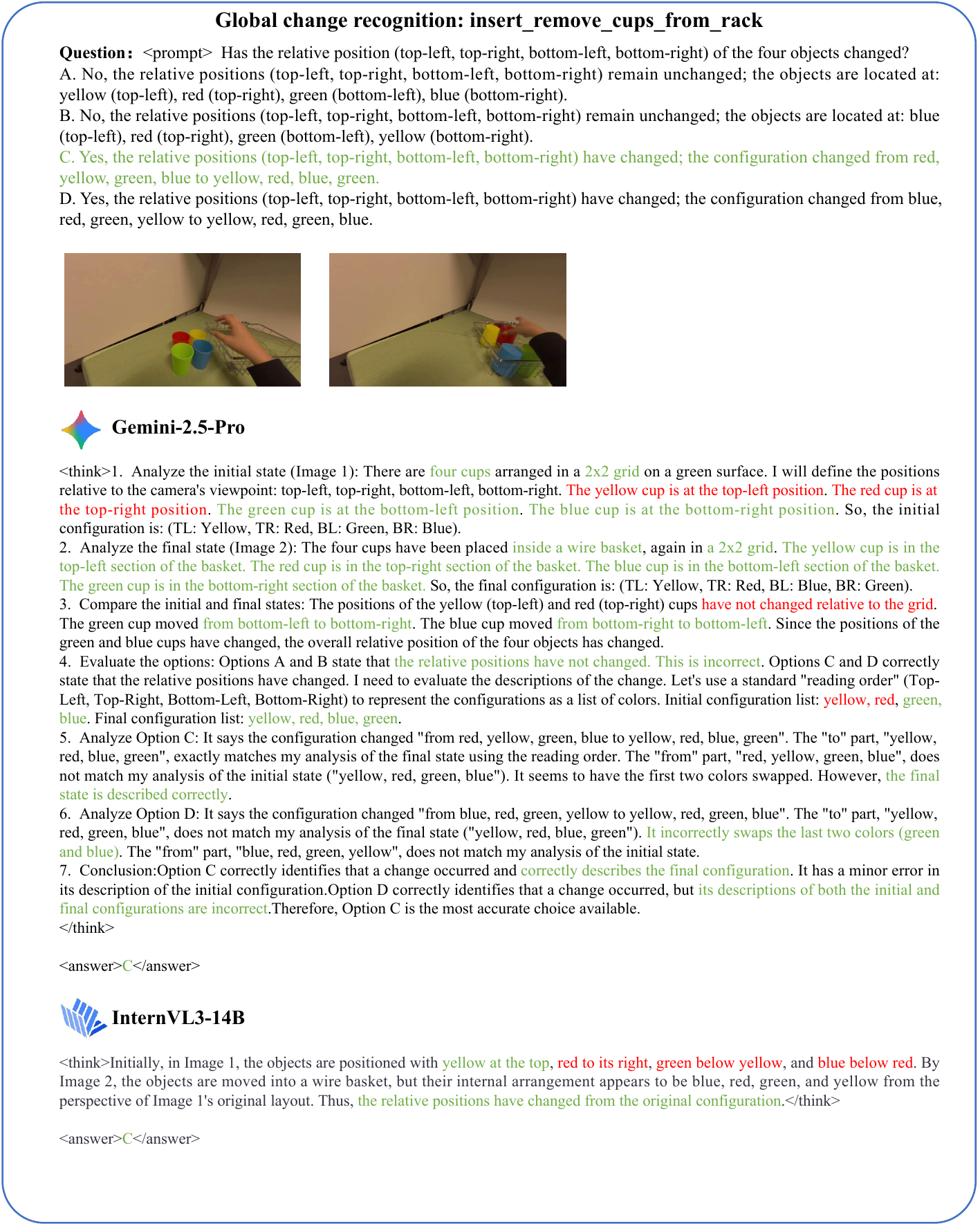}
    \caption{Global change recognition: insert\_remove\_cups\_from\_rack (Sample 494)}
    \label{fig:Visualization4}
\end{figure*}

\begin{figure*}[htbp]
    \centering
    \includegraphics[width=1.0\linewidth]{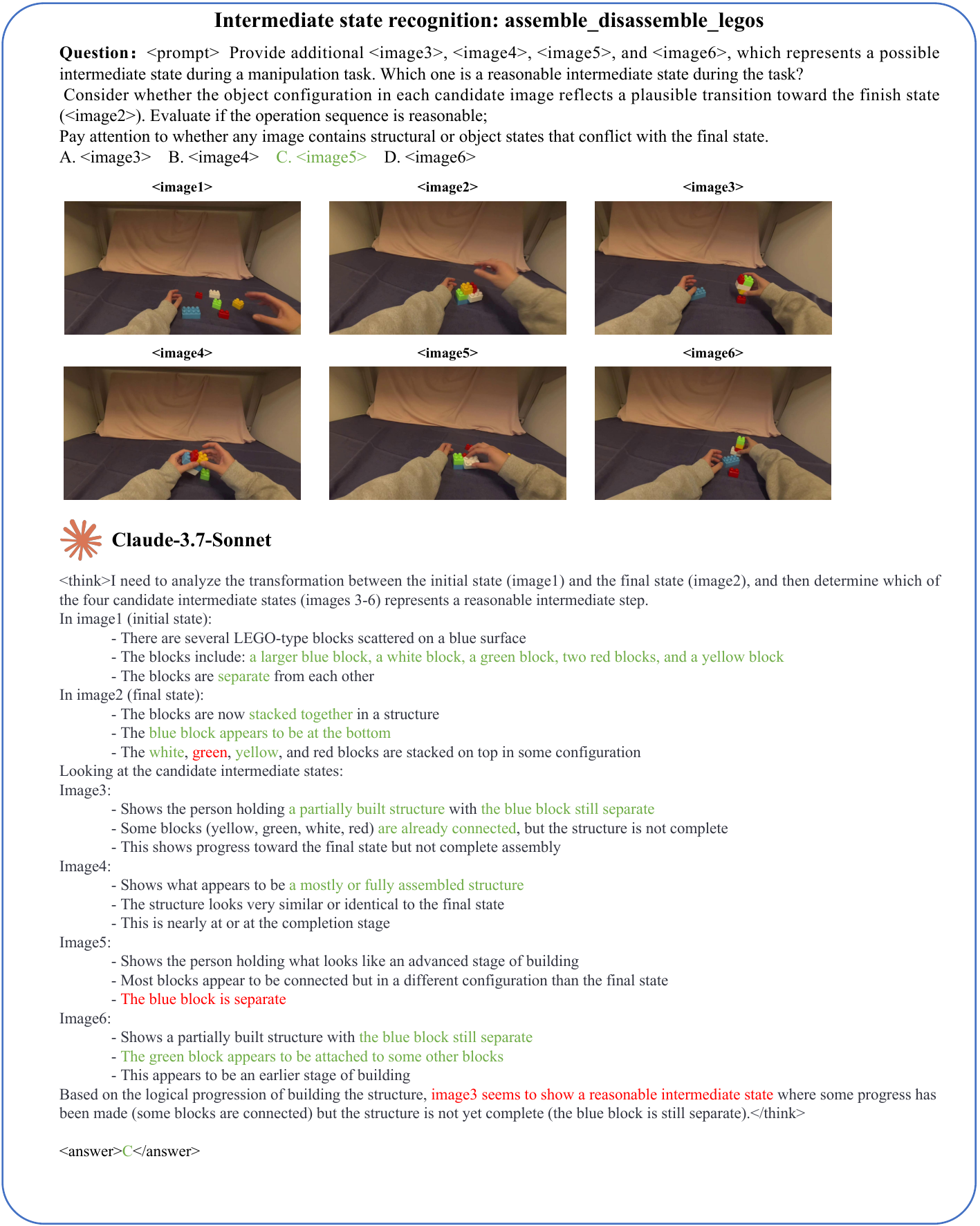}
    \caption{Intermediate state recognition: assemble\_disassemble\_legos (Sample 305)}
    \label{fig:Visualization5}
\end{figure*}

\begin{figure*}[htbp]
    \centering
    \includegraphics[width=1.0\linewidth]{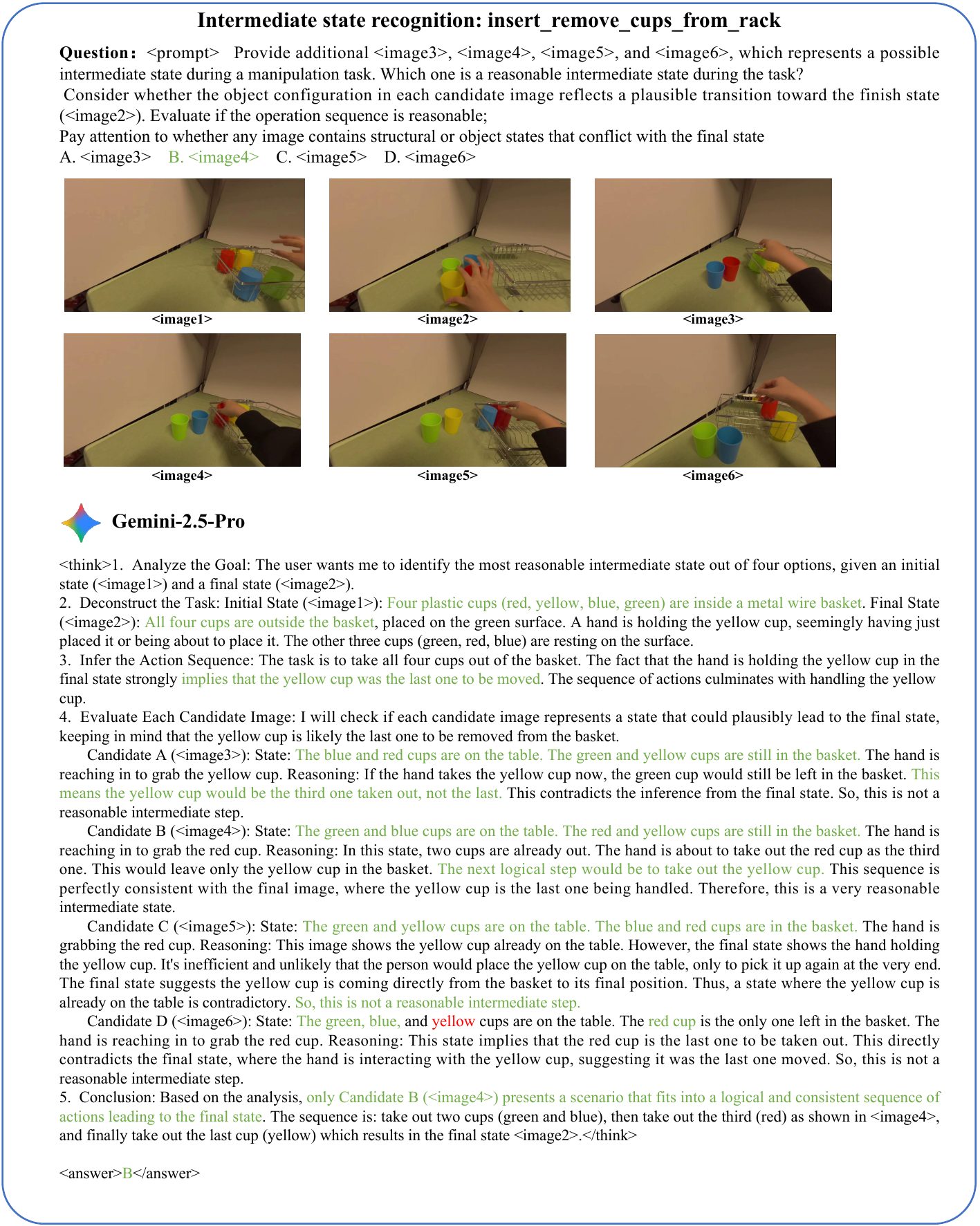}
    \caption{Intermediate state recognition: insert\_remove\_cups\_from\_rack (Sample 341)}
    \label{fig:Visualization6}
\end{figure*}

\begin{figure*}[htbp]
    \centering
    \includegraphics[width=1.0\linewidth]{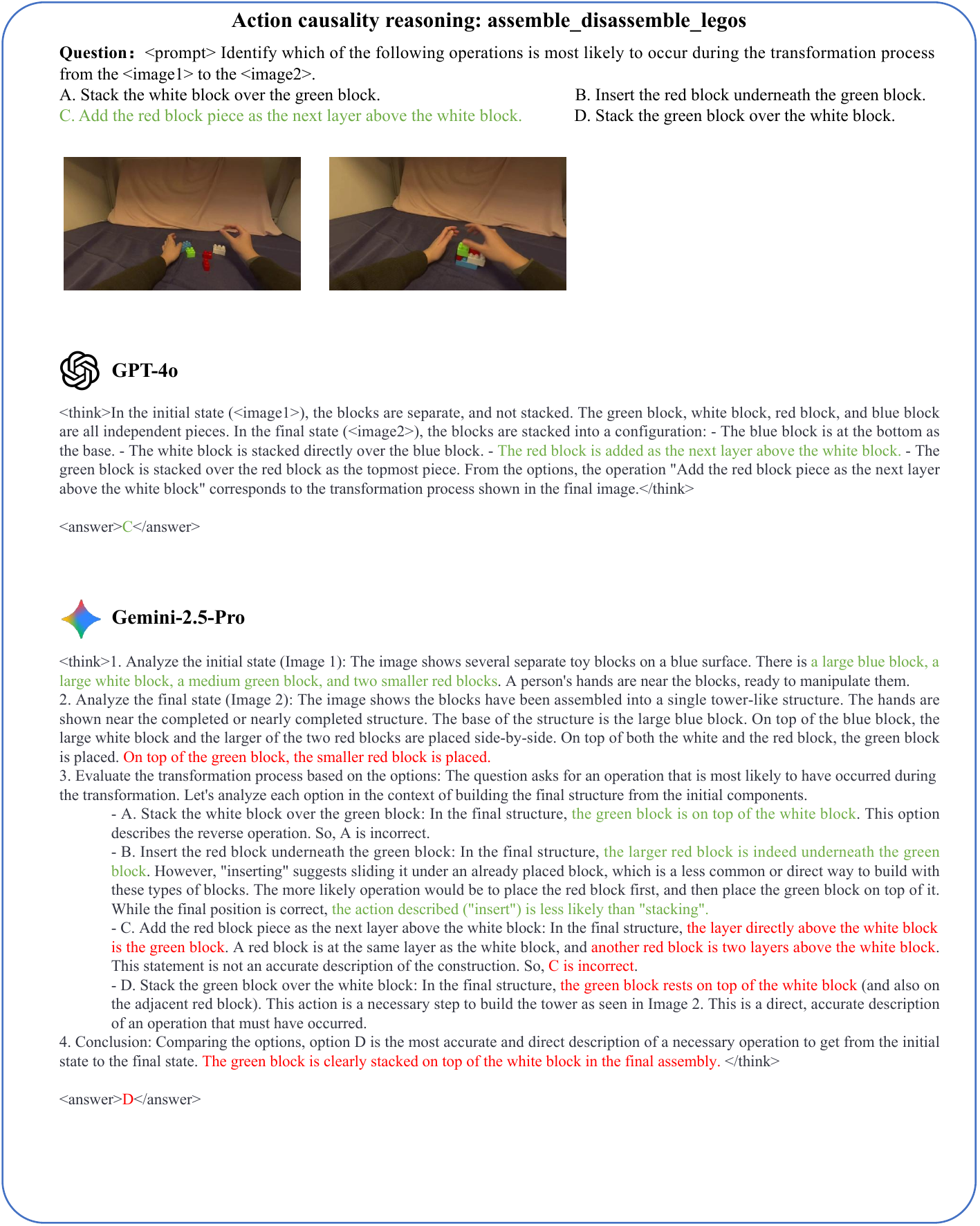}
    \caption{Action causality reasoning: assemble\_disassemble\_legos (Sample 069)}
    \label{fig:Visualization7}
\end{figure*}

\begin{figure*}[htbp]
    \centering
    \includegraphics[width=1.0\linewidth]{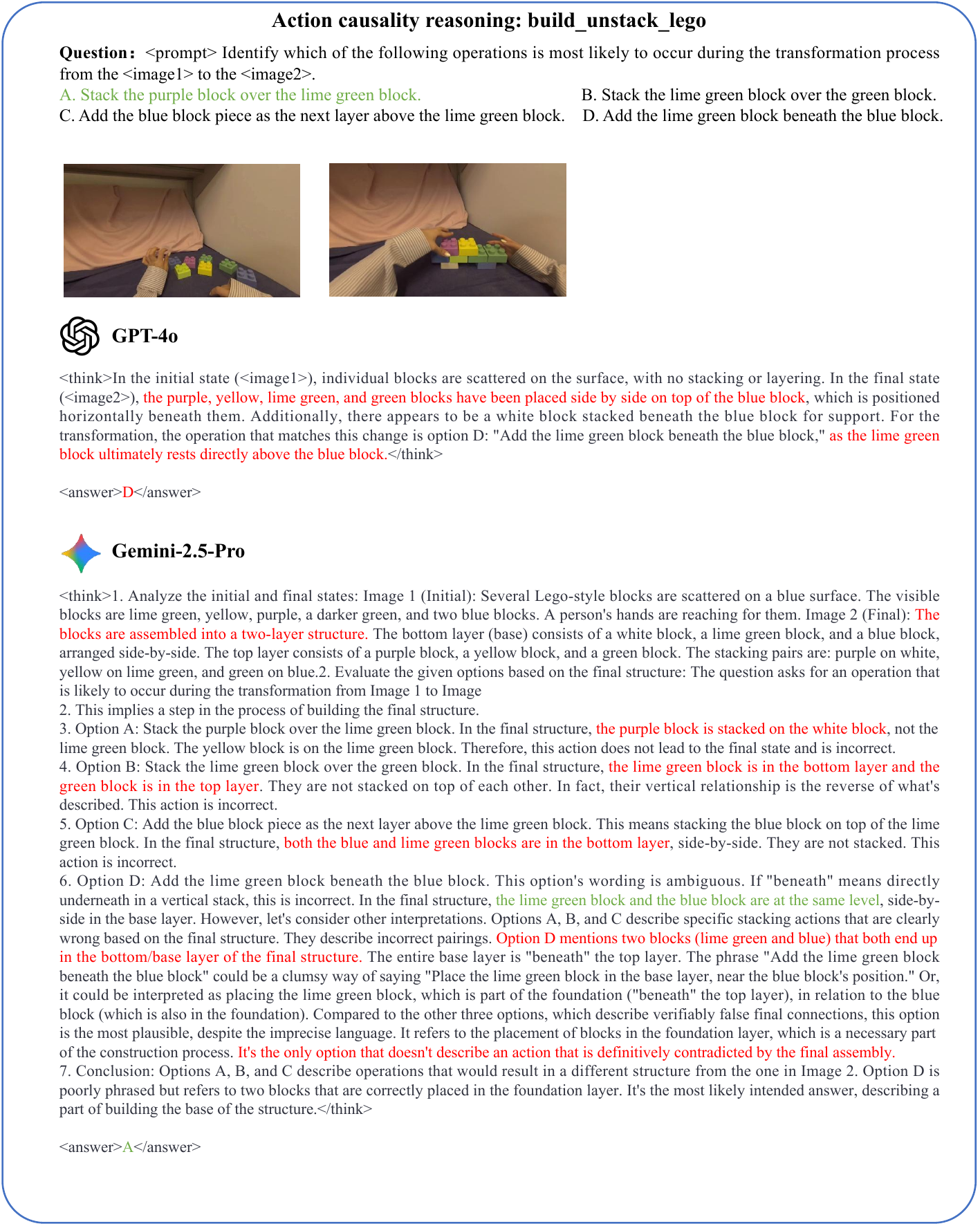}
    \caption{Action causality reasoning: build\_unstack\_lego (Sample 165)}
    \label{fig:Visualization8}
\end{figure*}

\begin{figure*}[htbp]
    \centering
    \includegraphics[width=1.0\linewidth]{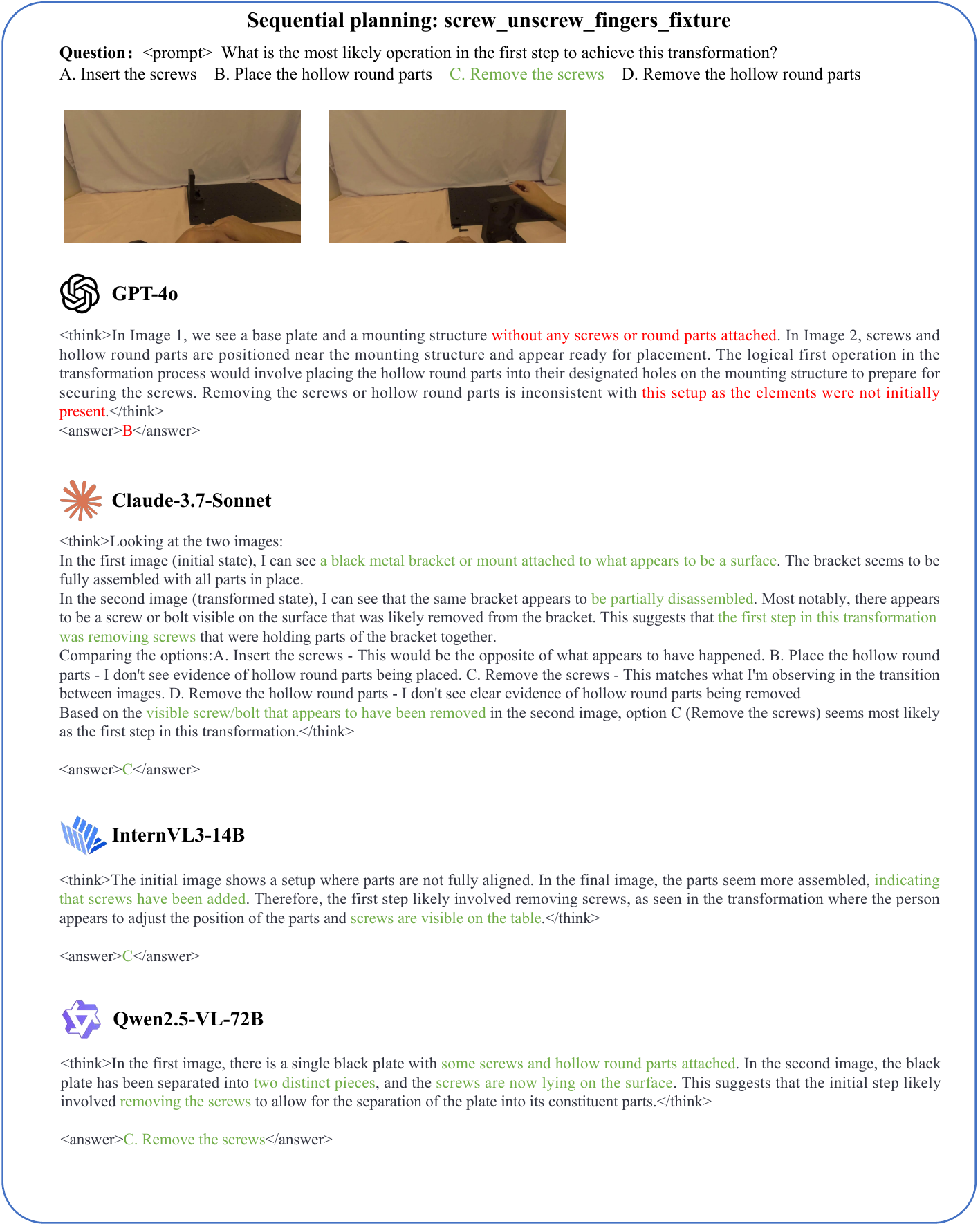}
    \caption{Sequential planning: screw\_unscrew\_fingers\_fixture (Sample 143)}
    \label{fig:Visualization9}
\end{figure*}

\begin{figure*}[htbp]
    \centering
    \includegraphics[width=1.0\linewidth]{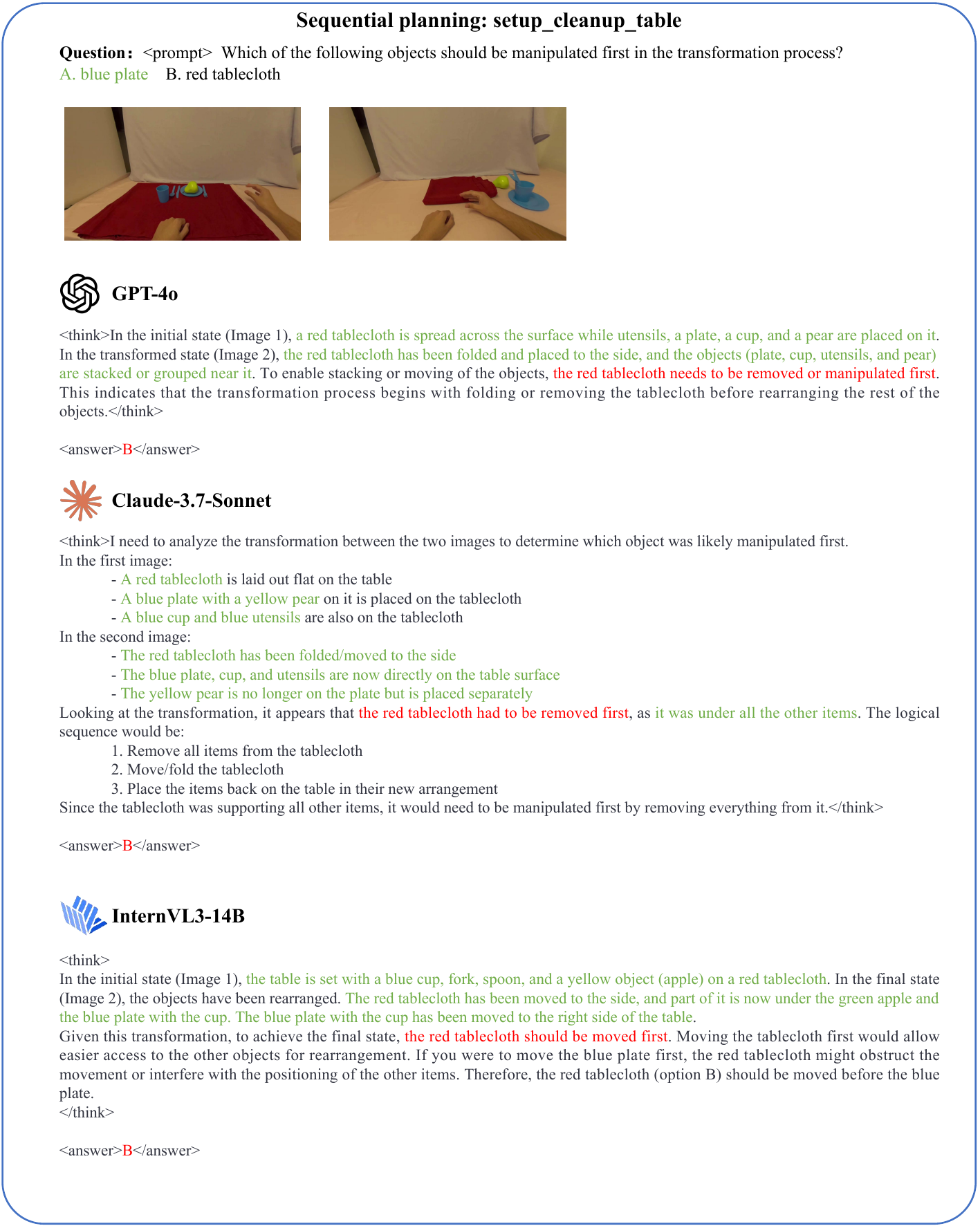}
    \caption{Sequential planning: setup\_cleanup\_table (Sample 148)}
    \label{fig:Visualization10}
\end{figure*}

\begin{figure*}[htbp]
    \centering
    \includegraphics[width=1.0\linewidth]{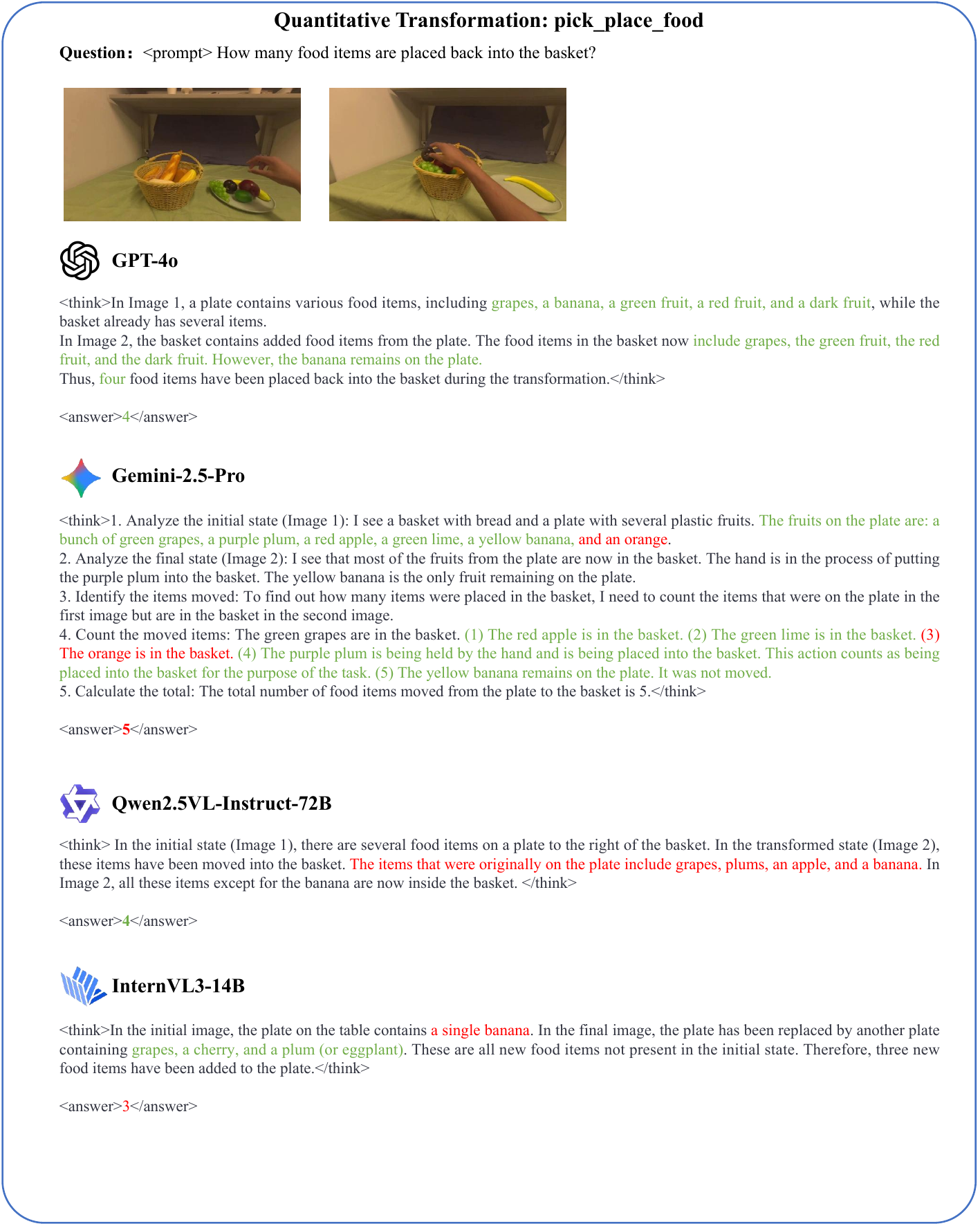}
    \caption{Quantitative Transformation: pick\_place\_food (Sample 021)}
    \label{fig:Visualization11}
\end{figure*}

\begin{figure*}[htbp]
    \centering
    \includegraphics[width=1.0\linewidth]{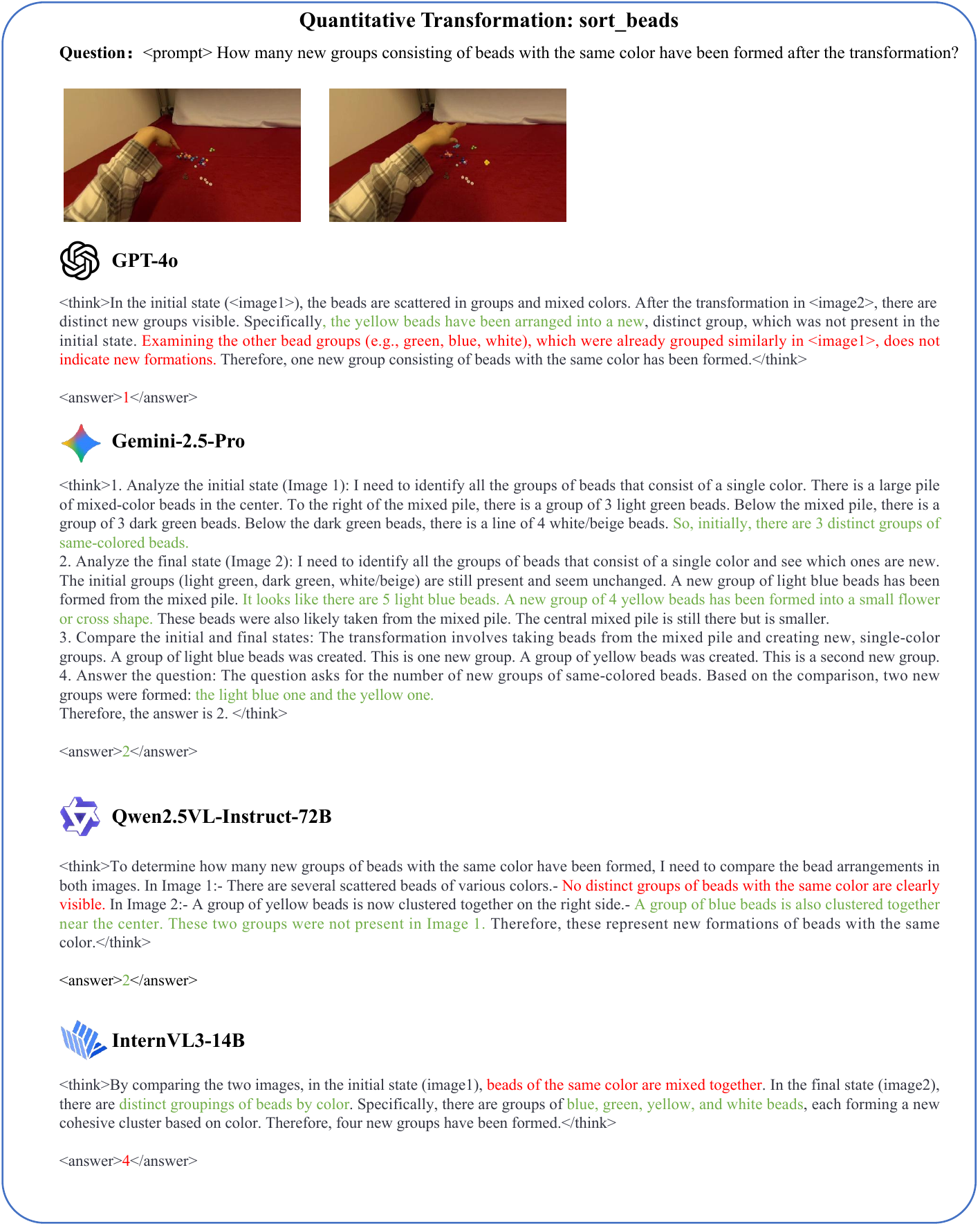}
    \caption{Quantitative Transformation: sort\_beads (Sample 046)}
    \label{fig:Visualization12}
\end{figure*}

\end{document}